\documentclass{article}
\usepackage[a4paper,left=1in,top=1in,right=1in,bottom=1in,nohead]{geometry}
\usepackage[onehalfspacing]{setspace}
\usepackage[utf8]{inputenc}
\usepackage{graphicx}
\usepackage{url}
\usepackage[export]{adjustbox}
\usepackage{amsmath}
\usepackage{natbib}

\title{Interpolation, extrapolation, and local generalization in common neural networks}

\author{Laurent Bonnasse-Gahot}
\date{Centre d'Analyse et de Math\'ematique Sociales \\
	CAMS, UMR 8557 CNRS-EHESS \\
	\'Ecole des Hautes \'Etudes en Sciences Sociales \\
	54 bd. Raspail, 75006 Paris, France\\
	\texttt{lbg@ehess.fr}\\
	$\,$
	
	July 18, 2022
}

\begin{document}
\maketitle
\noindent\rule{\textwidth}{0.8pt}
\begin{abstract}	
There has been a long history of works showing that neural networks have hard time extrapolating beyond the training set. A recent study by \citet{balestriero2021learning} challenges this view: defining interpolation as the state of belonging to the convex hull of the training set, they show that the test set, either in input or neural space, cannot lie for the most part in this convex hull, due to the high dimensionality of the data, invoking the well known curse of dimensionality. Neural networks are then assumed to necessarily work in extrapolative mode. We here study the neural activities of the last hidden layer of typical neural networks. Using an autoencoder to uncover the intrinsic space underlying the neural activities, we show that this space is actually low-dimensional, and that the better the model, the lower the dimensionality of this intrinsic space. In this space, most samples of the test set actually lie in the convex hull of the training set: under the convex hull definition, the models thus happen to work in interpolation regime. Moreover, we show that belonging to the convex hull does not seem to be the relevant criteria. Different measures of proximity to the training set are actually better related to performance accuracy. Thus, typical neural networks do seem to operate in interpolation regime. Good generalization performances are linked to the ability of a neural network to operate well in such a regime.

\end{abstract}
\noindent\rule{\textwidth}{0.8pt}

\section{Introduction}
\label{sec:introduction}

Deep learning is the modern rebranding of artificial neural networks, which were revived ten years ago and have been very successful since then in many domains including computer vision and natural language processing \citep[see][for reviews]{lecun2015deep, schmidhuber2015deep}. But in spite of this tremendous success, some authors also point out the limitations of these current techniques, described as essentially curve-fitting models \citep{chollet2021deep,marcus2018deep}. Contrary to human beings, these techniques are very ``data-hungry'', as they need to have access to a very large training set, and have limited generalization capabilities, well below human abilities \citep{marcus2018deep}. In particular, neural networks do not extrapolate well beyond their training data: there is a long history of results on this point \citep{barnard1992extrapolation,haley1992extrapolation,marcus1998rethinking}, and more recent results come to a similar conclusion \citep[see for example][]{barrett2018measuring, lake2018generalization, saxton2019analysing}. Recently, \citet{balestriero2021learning} have challenged the idea that neural networks essentially perform interpolation of the training data. The authors provide an operational definition of interpolation/extrapolation: there is interpolation whenever a new sample lie within the convex hull of the training data. The argument is then essentially based on the well-known curse of dimensionality: in high dimension, every point is far from another. In particular, in high dimension, any new point almost surely lie outside of the training set convex hull. Given the high-dimensionality nature of the data that we are typically dealing with, the authors conclude that neural networks cannot do interpolation and thus contrary to previous claims always actually operate in an extrapolation regime. Beyond looking at the input (pixel) space, \citet{balestriero2021learning} extend their analysis to the study of the embedding space constructed by a neural network, anticipating the criticism that interpolation does not happen in input space, but rather in such embedding space. In this case too, their argument remains the same: due to the high dimensionality of such space, the model still operates in extrapolation regime. A good point of this work is the will to provide a quantitative investigation of this debate. Although we agree on the facts about the curse of dimensionality, we do not agree that it ruled out the possibility that current neural networks essentially perform interpolation. Input data such as images are indeed high-dimensional at the surface level, but they actually often live in a lower dimensional space called the intrinsic manifold. This refers to the manifold hypothesis, and is at the basis of dimensionality reduction techniques such as Isomap~\citep{tenenbaum2000global}, Locally Linear Embedding~\citep{roweis2000nonlinear}---see \citet{cayton2005algorithms} for a review. Similarly, the embedding space defined by the neural space (the last layer before the categorical decision) is not the latent space itself, nor is any of its affine subspace,  but there exists a low-dimensional (latent) space that implicitly characterizes the neural space.\\

\begin{figure}
	\centering	
	\begin{minipage}[t]{0.28\linewidth}	
		\textbf{A}\\[-5pt]
		\hspace{-0.3cm}
		\includegraphics[width=.95\linewidth,valign=T]{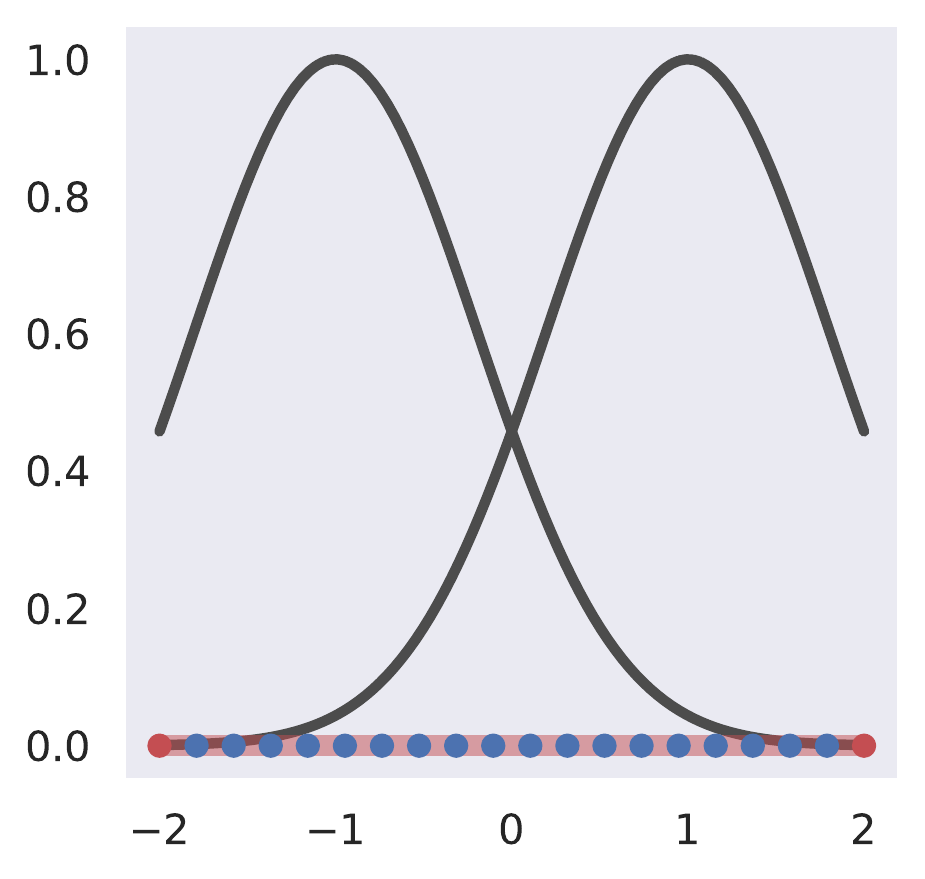}\\[10pt]
		\textbf{B}\\[-5pt]
		\hspace{-0.3cm}
		\includegraphics[width=.95\linewidth,valign=T]{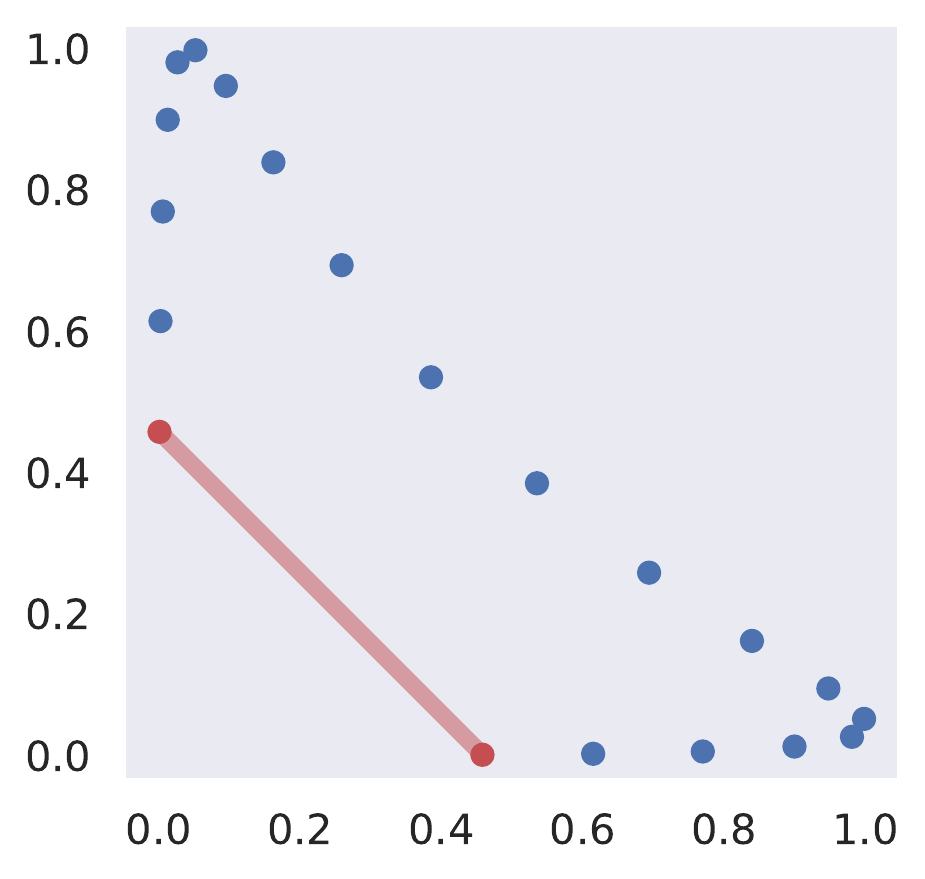} 
	\end{minipage}
	\hspace{2.0cm}
	\begin{minipage}[t]{0.28\linewidth}	
		\textbf{C}\\[-5pt]
		\hspace{-0.3cm}
		\includegraphics[width=.95\linewidth,valign=T]{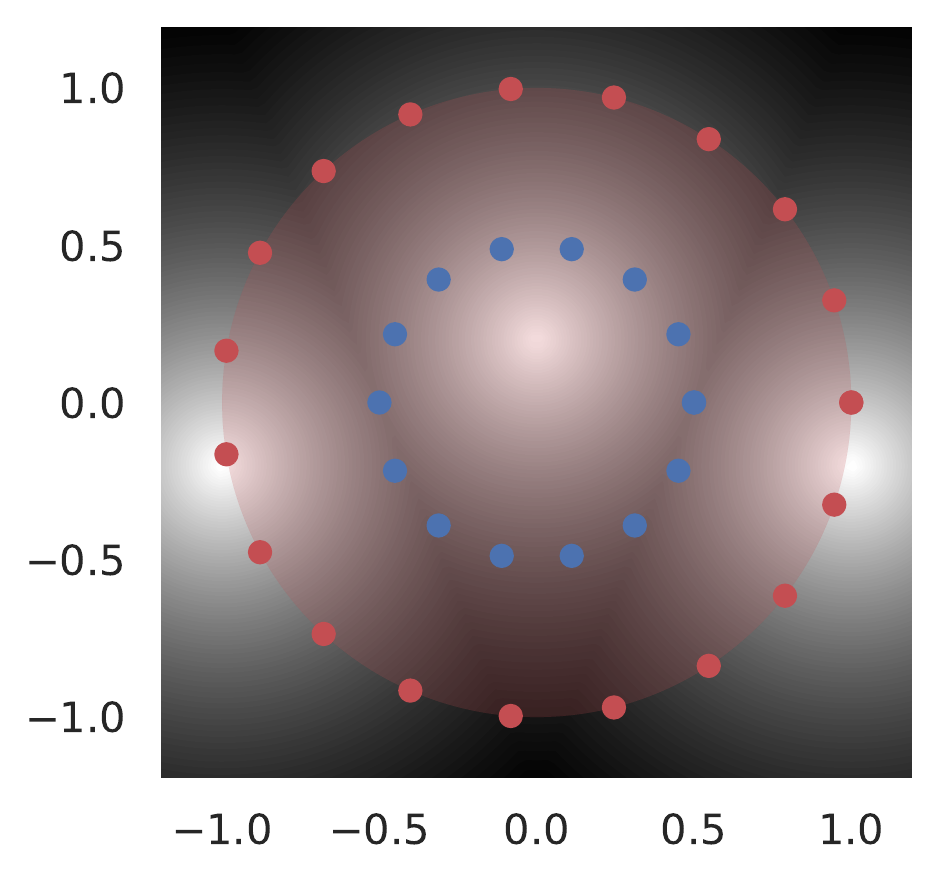}\\[10pt]
		\textbf{D}\\[-5pt]
		\hspace{-0.3cm}
		\includegraphics[width=1.\linewidth,valign=T]{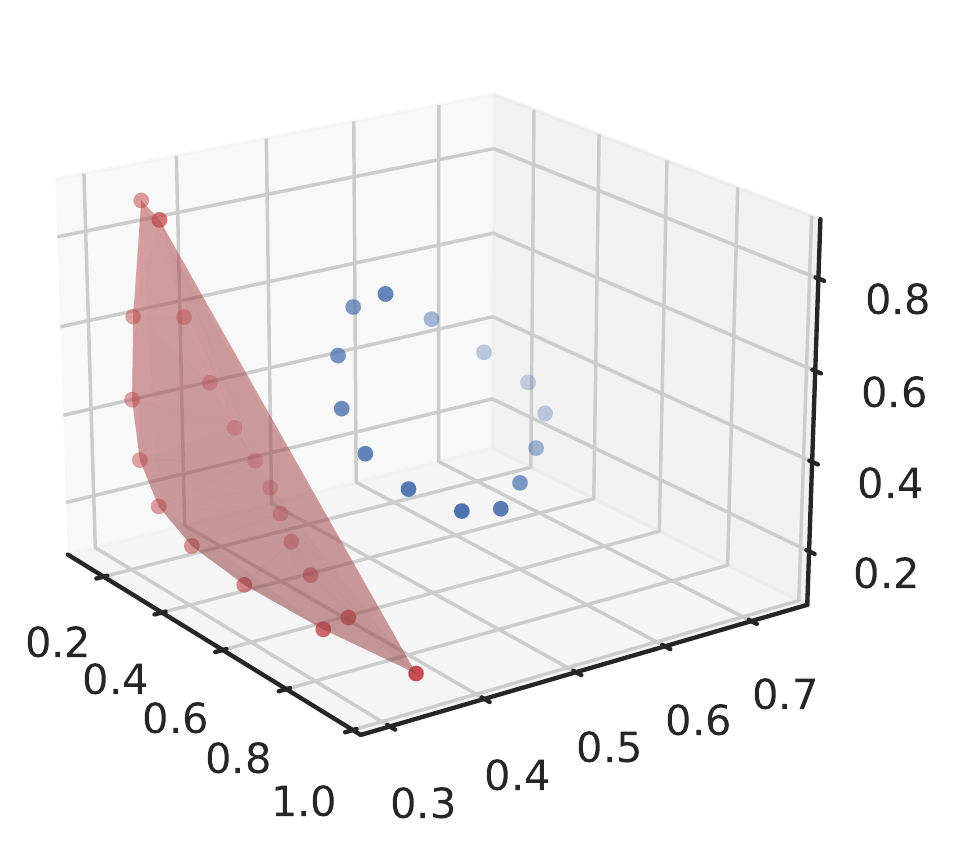} 
	\end{minipage}
	\caption{\textbf{Convex hull in intrinsic space (top) does not correspond to convex hull in embedded neural space (bottom).} (A) One dimensional intrinsic space. Response of two neurons with bell-shaped tuning curves along a one-dimensional space. If we consider the red points as the training set, the red line corresponds to the convex hull; then all the blue points, considered here as test samples, between the two red ones lie within the convex hull. But if we now move to the neural space defined by the activities of these two neurons, (B), none of these blue points actually lie within the convex hull (the red line) defined by the red samples. Similarly, (C) Two dimensional intrinsic space, with three neurons (whose responses are represented in shades of gray). In this space, all the blue points (the inner circle) lie within the convex hull (the red disk) defined by the red points (the outer circle), but turning to the neural space, (D), none of them actually lie within the new red convex hull.}
	\label{fig:convexhull}
\end{figure}

There has been a recent surge of works in neuroscience that show how (real-world biological) neural activities have to be understood as acting on a lower-dimensional intrinsic manifold~\citep{archer2014low,sadtler2014neural,cunningham2014dimensionality,gallego2017neural,chung2021neural,jazayeri2021interpreting}. In the machine learning literature too, different works have also shown that both input data and neural activities in the hidden layers of artificial neural networks actually lie in a low-dimensional space \citep{ma2018dimensionality, ansuini2019intrinsic, pope2021intrinsic}. In such a low-dimensional manifold, the models might operate in interpolation mode according to the convex hull definition, although this might not be obvious in the higher dimensional space defined by the neural activities. To clarify this point, let us take a look at Fig.~\ref{fig:convexhull} that present two very simple illustrative examples. Fig.~\ref{fig:convexhull}A describes two neurons with bell-shaped tuning curves along a one-dimensional space \citep[the tuning curve represents the response of a neuron to a range of stimulus; see for example][]{hubel1959receptive, henry1974orientation, taube1990head}. Let us imagine that the training set is made of the two red dots pictured in Fig.~\ref{fig:convexhull}A. The convex hull of this training set is the red line, and we consider as the test set all the points, in blue, between these two training samples. By definition here they all belong to the convex hull of the training set. Now, if we consider how this situation translates in the state space defined by the activities of the two neurons, the picture changes drastically (Fig.~\ref{fig:convexhull}B): all the test samples lie outside the new convex hull of the training set. A similar situation is exemplified in Fig.~\ref{fig:convexhull}C and~\ref{fig:convexhull}D. 
It might thus be misleading to directly look at the neural activities, as the high-dimensionality of this neuronal space is only apparent, and can actually be seen as lying in a lower dimensional manifold.\\

In an artificial neural network, the activity of a given layer can be seen as a population coding of a low-dimensional space paved with neurons (without necessarily a simple contiguous tuning curve). In this work we use an autoencoder (a multilayer perceptron here) in order to recover the implicit intrinsic space underlying the neural space. An autoencoder can be used to reduce the dimensionality of any high-dimensional data pertaining to a lower dimensional manifold~\citep{hinton2006reducing}. Contrary to other techniques such as Principal Component Analysis, Isomap~\citep{tenenbaum2000global} or Locally Linear Embedding~\citep{roweis2000nonlinear}, an autoencoder makes it possible not only to map the activities to positions in a low dimensional latent space, but also to estimate the neural activities from this intrinsic space. Here, we use an autoencoder to probe the representation of a given layer in a neural network after learning, and study a posteriori, without modifying the base network under study, how the classification performance changes with the number of estimated intrinsic dimension. In this latent space, we also look at the fraction of samples from the test set that lie within the convex hull defined by the training set. We find that the intrinsic dimensionality underlying the neural activity is actually quite low, and for such low values all or almost all the test set is included in the convex hull of the training set. We also experiment changing the characteristics of the networks so as to study how the intrinsic dimensionality of the neural space vary with the classification performances, while keeping the same number of ambient dimension. We find that the better the performances the lower the number of intrinsic dimensions. Following \citet{balestriero2021learning} definition of interpolation as being in the convex hull, we thus find that the better a model is, the more it is in an interpolative regime. In a second part, we will see that the notion of interpolation as belonging to the convex hull is not the most relevant, and we study properties of the network with respect to the local proximity to elements of the training set. This study reveals that as expected the closer a new test sample of the training set, the higher its probability of being correctly classified: this behavior is typical of interpolation. This phenomenon is actually even more marked the better a model is. It is not the case that the best models manage to extrapolate better: they actually better represent the data so as to take advantage of the interpolative regime in which they operate. Thus, contrary to the claims of \citet{balestriero2021learning}, classic deep learning techniques does seem to perform in interpolative regime, as described for instance by \citet{marcus2018deep} or \citet{chollet2021deep}. 
Generalization performances are actually tightly related to the notion of interpolation, at least with the simple models that we consider here.

\section{Materials and Methods}
\label{sec:methods}

\subsection{Probing the neural representation}

Our method to probe the neural representation is illustrated in Figure~\ref{fig:method}. A neural network (we consider a multilayer perceptron or a convolutional neural network in the Results section) is trained to learn some classification task. In the Results section we will consider the recognition of handwritten digits with the MNIST database \citep{lecun1998gradient} and the classification of natural images with the CIFAR-10 database \citep{krizhevsky2009learning}. The first layer corresponds to the input space. For instance, in the case of MNIST, this input space (pixel space) is of dimension $28\times28=786$, and in the case of CIFAR-10 $32\times32\times3=3072$, which correspond in all cases to high-dimensional data. One or several hidden layer(s) follow(s) the input layer, depending on the architecture. Although we could probe any layer in the network, we will focus in this work on probing the last layer before the categorical decision, where the categories are organized into linearly separable clusters of neural activities. We call this space the neural space. In all the numerical experiments of the section Results, this neural space has the same ambient dimension, 128. In order to probe it, we use an autoencoder to recover the latent space underlying the neural activities of the layer under scrutiny (Fig.~\ref{fig:method}B). This is done by training the autoencoder to reconstruct the neural activities, using the training set (the loss is the mean square error here). It is important to note that this analysis is done a posteriori, after the network under investigation has learned its task. Finally, we substitute the neural activities by the ones approximated by the autoencoder, and we look at the classification accuracy of this hybrid network (Fig.~\ref{fig:method}C). We repeat this procedure for different numbers of dimension of the estimated intrinsic space (the number of neurons in the bottleneck of the autoencoder). We assume that for a given dimension, if the autoencoder is able to reconstruct the neural activities well enough, this hybrid network should have similar classification performance (on the test set). If the number of dimensions is too low, the classification performances should be lower than the original network. Starting from the true intrinsic dimension, the performances should be equal to the ones of the original network, and plateau for larger values of the estimated intrinsic dimension. Note again that during this analysis all the parts from the original network are left untouched, completely frozen, and that the autoencoder is not trained on the classification task, but only on the task of reconstructing the neural activities.\\

\subsection{Toy example with ten Gaussian categories}
\begin{figure}	
	\centering
	\begin{minipage}{0.3\linewidth}		
		\textbf{A}
		\includegraphics[width=0.8\linewidth,valign=t]{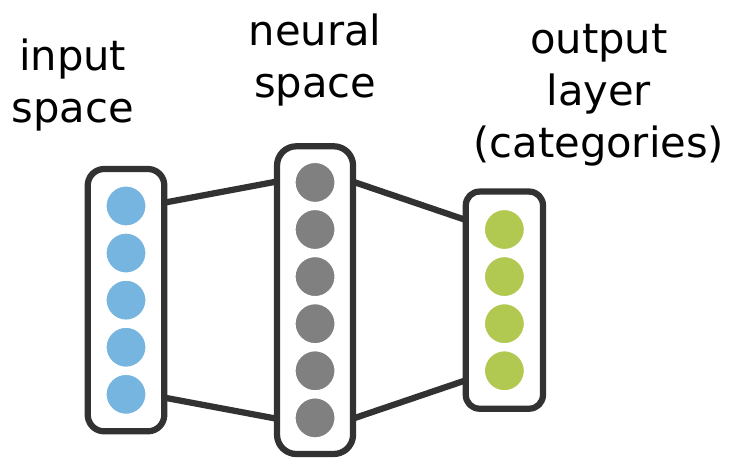}\\
		
		\vspace{0.2cm}
		\textbf{B}
		\includegraphics[width=0.8\linewidth,valign=t]{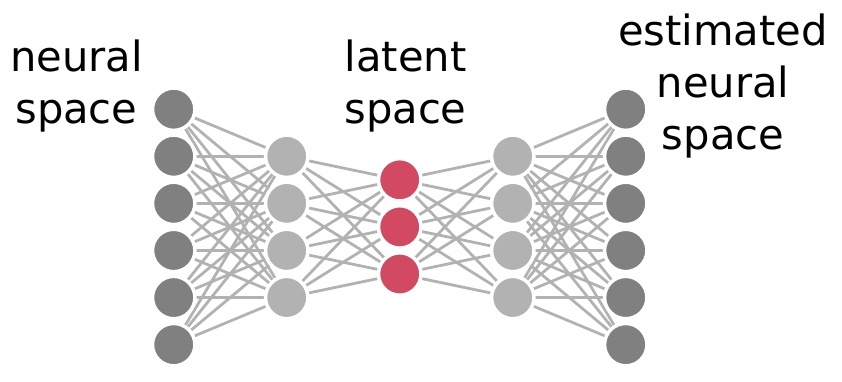}\\
				
		\vspace{0.2cm}
		\textbf{C}
		\includegraphics[width=0.8\linewidth,valign=t]{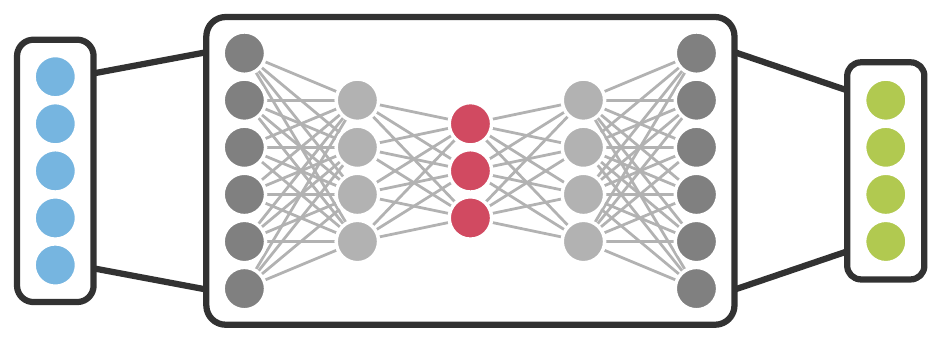}\\
	\end{minipage}
	\hspace{0.8cm}
	\begin{minipage}{0.6\linewidth}	
		\textbf{D}
		\includegraphics[width=0.8\linewidth,valign=t]{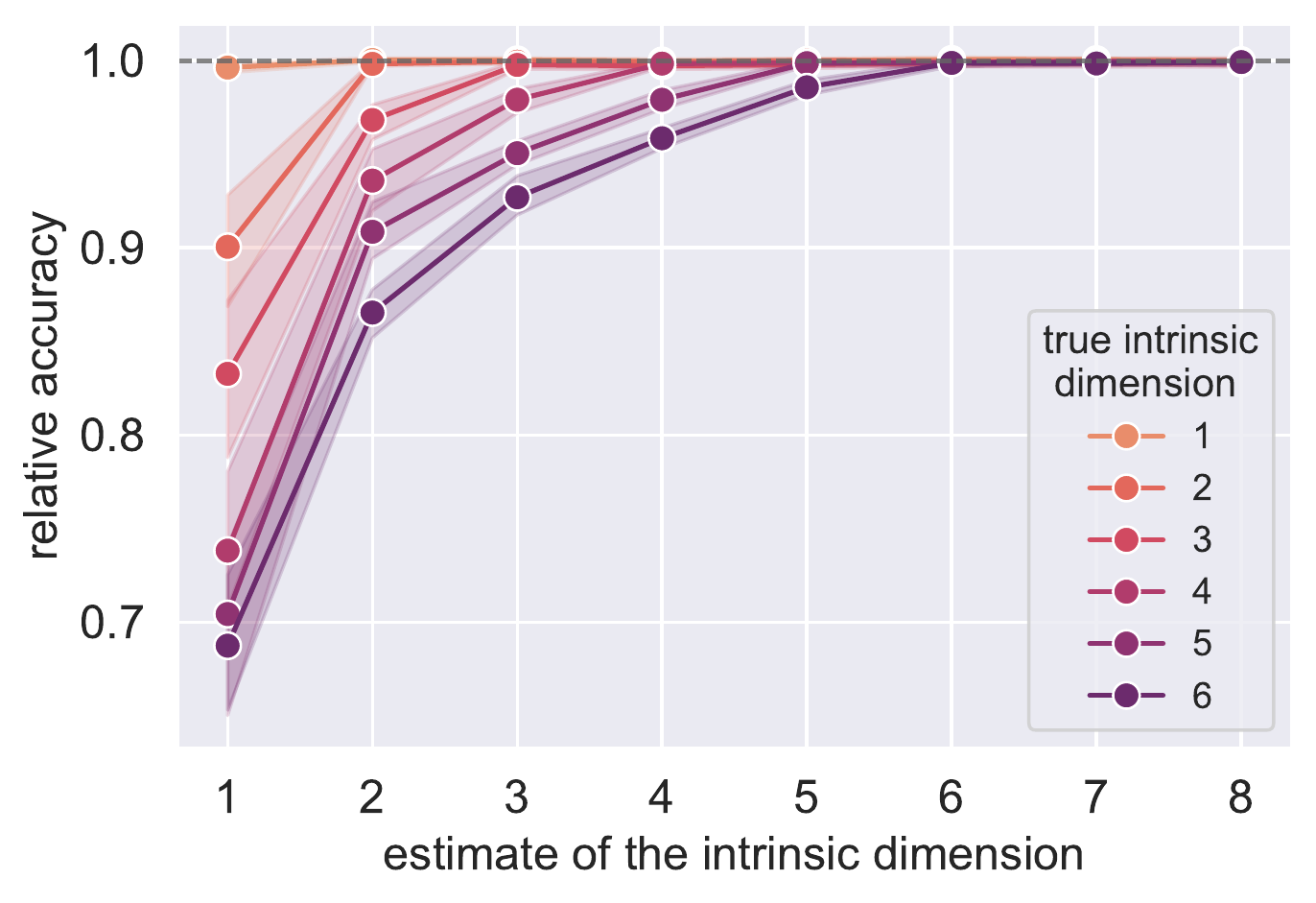}
	\end{minipage}
	\caption{\textbf{Description of the method use to probe the representation of a neural space.} A neural network, (A), trained on a given classification task is then analyzed by using an autoencoder, (B), which is trained to reconstruct the neural activities under consideration. Using an hybrid network, (C), made of the original neural network and the activities predicted by the autoencoder, we then compute the classification accuracy on the test set (without any learning/finetuning). Varying the number of neurons in the bottleneck of the autoencoder serves as an estimate of the number of dimensions of the intrinsic space underlying the neural activities. (D) A simple example involving 10 Gaussian categories living in a controlled space of varying intrinsic dimension but same input dimension: when then number of dimensions in the reconstructed latent space reaches the true intrinsic dimension, the classification accuracy of the hybrid network (that forces the neural activities to be estimated through a small dimensional bottleneck) is the same as the one of the original network. The relative accuracy refers to accuracy of the hybrid network divided by the one of the original network (mean absolute accuracy is 71\%). For each value of the true intrinsic dimension, the process is repeated for 10 trials, and error bars indicate 95\% bootstrap confidence intervals.}
	\label{fig:method}
\end{figure}

In order to validate this general framework, we first go through a simple fully controlled example involving classification of Gaussian categories. In this example we consider 10 categories (as in the MNIST and CIFAR-10 examples that will be used in the Results section). In all the following cases, the stimuli drawn from these categories live in same input space, with $n_{\text{input}}=32$ dimensions, but depending on the condition the intrinsic space has different dimensions. In order to control for the number of dimensions of the intrinsic space,  for a given desired value notated $n_{\text{id}}$, we generate the centers of the categories randomly in a space of dimension $n_{\text{id}}$. These centers are then plunged into a $n_{\text{input}}$-dimensional space by applying a random rotation/reflection matrix. This transformation is generated thanks to a QR decomposition of a random matrix $A = QR$ drawn from a normal distribution with mean 0 and variance 1. This factorization yields $Q$, which is an orthogonal matrix, and corresponds to a rotation in input space (possibly combined with a reflection). Finally, we generate 5000 stimuli per class for the training set and 1000 stimuli per class for the test set by drawing random realizations of a normal distribution centered in each of these centers and with an isotropic variance scaled so as to yield an average classification performance around 70$\%$.\\
A multilayer perceptron with one hidden layer of 32 neurons is then trained to classify the categories. It is important to note that in all the different cases, the neural space has then the exact same dimension $n_{\text{neural}}=32$. After learning, we freeze and analyze this network by probing the hidden layer with the method described above, using different number of neurons in the autoencoder bottleneck as an estimate of the intrinsic dimension. Finally, in each case, using the test set, we compute the classification accuracy of the hybrid network that use the frozen base neural network and the neural activations predicted from the autoencoder. Figure~\ref{fig:method}D shows that this simple setting indeed enables to recover the true intrinsic dimension. As expected, when the probe has a lower dimension than the true intrinsic dimension, the accuracy of the model is lower than the base model; then, when it reaches the true intrinsic dimension (known here by construction), the model has the same classification accuracy, which then plateaus for greater inner dimension.

\subsection{Technical details}
\begin{figure}	
	\centering
	\textbf{A}
	\includegraphics[width=0.15\linewidth,valign=T]{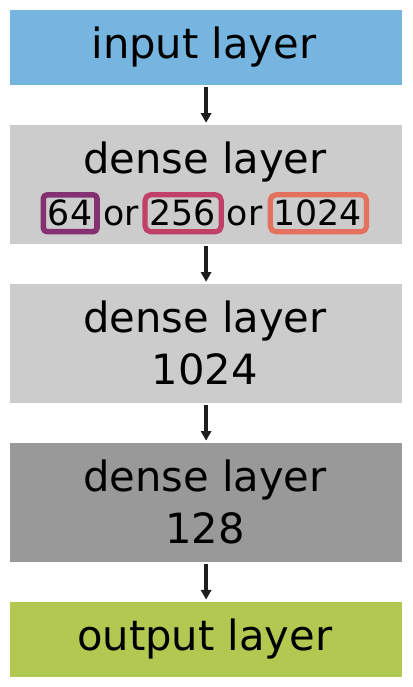} 
	\hspace{1.0cm}
	\textbf{B}
	\includegraphics[width=0.25\linewidth,valign=T]{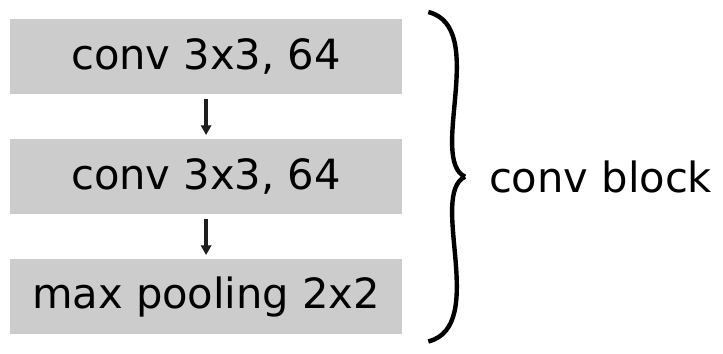} 
	\hspace{0.2cm}
	\includegraphics[width=0.16\linewidth,valign=T]{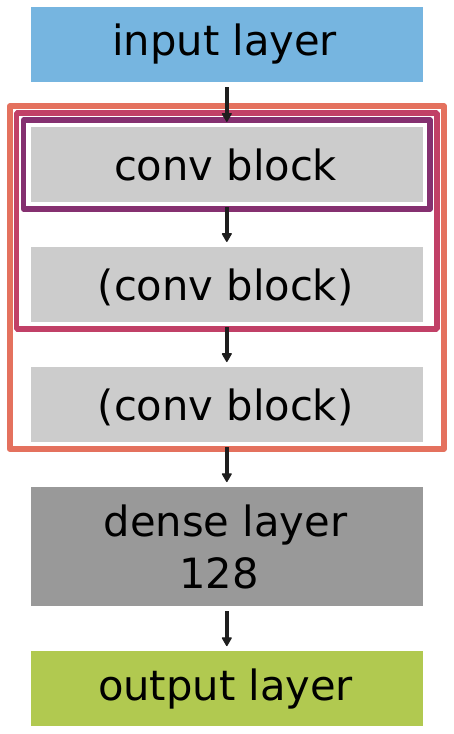} 
	\caption{\textbf{Artificial neural networks used in the present study.} (A) Multilayer perceptrons with three hidden layers. The number of neurons in the first hidden layer vary so as to induce variability in classification performances, in order to study its relationship with intrinsic dimensionality. (B) Convolutional neural networks. A convolutional block is defined as two conv layers followed by a max pooling layer. We consider three different cases, that use either 1, 2, or 3 such conv blocks. In all cases, the neural space under consideration is the dark gray layer, the hidden dense layer right before the output layer that gives the class probabilities. In all the different cases, this neural space has the same ambient dimension, 128 here.}
	\label{fig:ann}
\end{figure}

We consider two different databases, namely the MNIST database \citep{lecun1998gradient} and the CIFAR-10 database \citep{krizhevsky2009learning}. Figure~\ref{fig:ann} presents the two neural architectures (and their variants) that we consider. We vary the width of the first layer in the case of the multilayer perceptron, and the depth of the convolutional neural network, by repeating a certain number of times the convolutional block described in Fig.~\ref{fig:ann}B. The goal here is to explore different variants of a same architecture, with the same final neural space, but different performances due to the width or depth of the network. 
For the multilayer perceptron, each dense layer is followed by a dropout layer \citep{srivastava2014dropout}, with probability of dropping a unit $p=0.2$ for the first hidden layer, $p=0.4$ for the second one, and $p=0.5$ for the last one. For the convolutional neural network, all the conv layers and the dense layer are followed by a batch normalization layer, and each conv block is followed by a dropout layer, with rate $p=0.2$ for the first one, $p=0.3$ for the second one, and $p=0.4$ for the last one. Before the softmax layer, we also apply dropout with rate $p=0.5$. All networks are trained through gradient descent using the Adam optimizer \citep{kingma2015adam}.\\
The autoencoder used to probe the representation of a given layer is a multilayer perceptron with 256 cells before and after the bottleneck. We consider 2, 4, 8 or 16 neurons in the bottleneck, used as an estimate of the intrinsic dimensionality of the neural space. All neurons use the ReLU activation function, except in the bottleneck layer, that makes use of the linear activation function. From the neural activities of a given layer (the layer before the output layer here), computed from the training set, the autoencoder is trained to reconstruct these neural activities, using the mean square error as the loss function. \\
For each case, all the results are averaged over 10 trials, with different random initializations, error bars indicating 95\% bootstrap confidence intervals.

\paragraph*{Computer code.} 
The custom Python 3 code written for the present project makes use of the following libraries: \texttt{tensorflow v2.8.0} \citep{tensorflow2015-whitepaper} (using \texttt{tf.keras} API, \citealp{chollet2015keras}), \texttt{numpy v1.21.4} \citep{harris2020array}, \texttt{scipy v1.8.0} \citep{scipy}, \texttt{pandas v1.3.4} \citep{mckinney2010data}, \texttt{matplotlib v3.4.3} \citep{hunter2007matplotlib}, \texttt{seaborn v0.11.2} \citep{Waskom2021} and \texttt{statsmodels v0.13.2} \citep{seabold2010statsmodels}. The code will be made available on GitHub.

\section{Results}
\label{sec:results}

\subsection{Convex hull in latent space}
\label{sec:results_ch}
\begin{figure}
	\centering
	\textbf{A}
	\hspace{-0.3cm}
	\includegraphics[width=0.22\linewidth,valign=T]{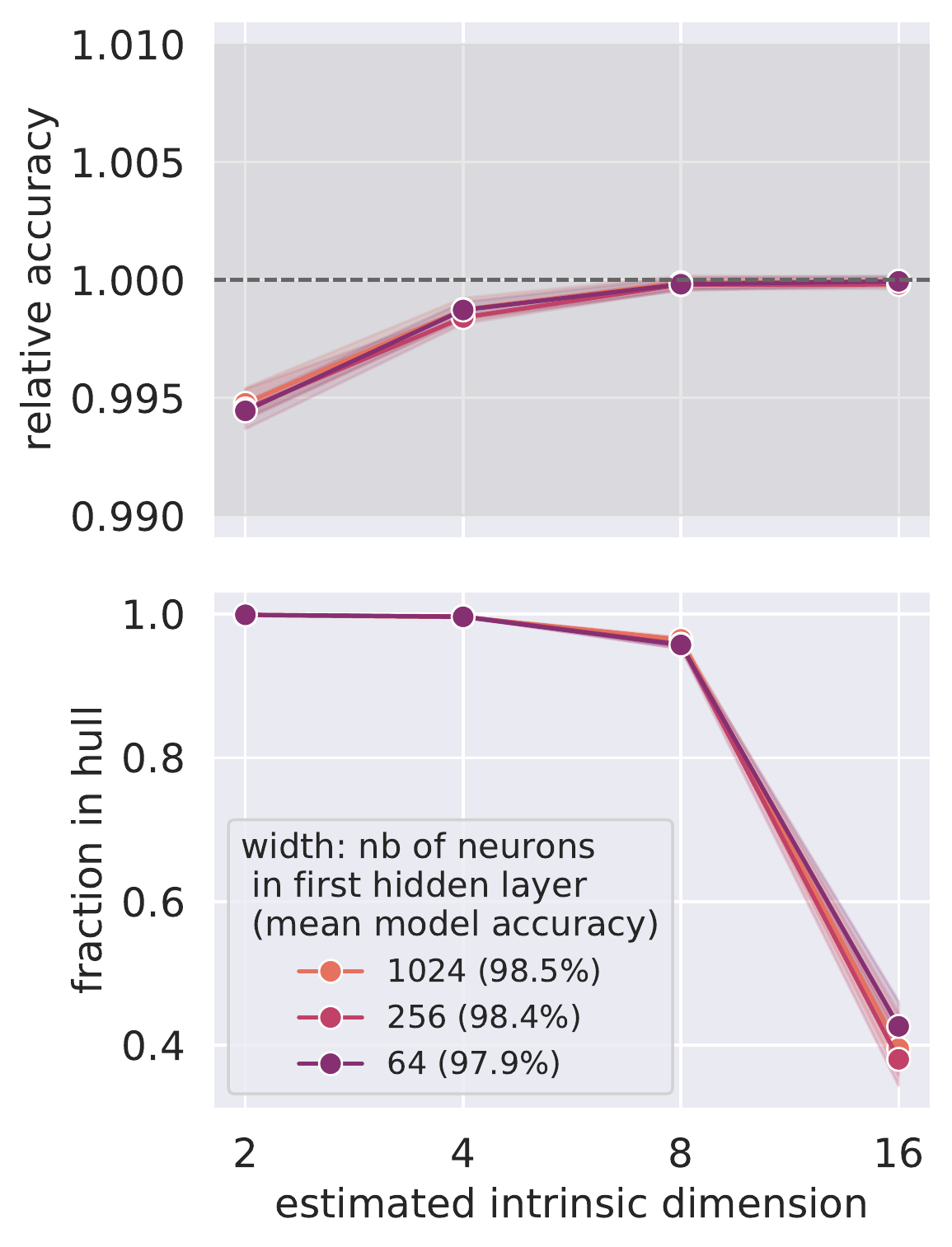} 
	\hfill
	\textbf{B}
	\hspace{-0.3cm}
	\includegraphics[width=0.22\linewidth,valign=T]{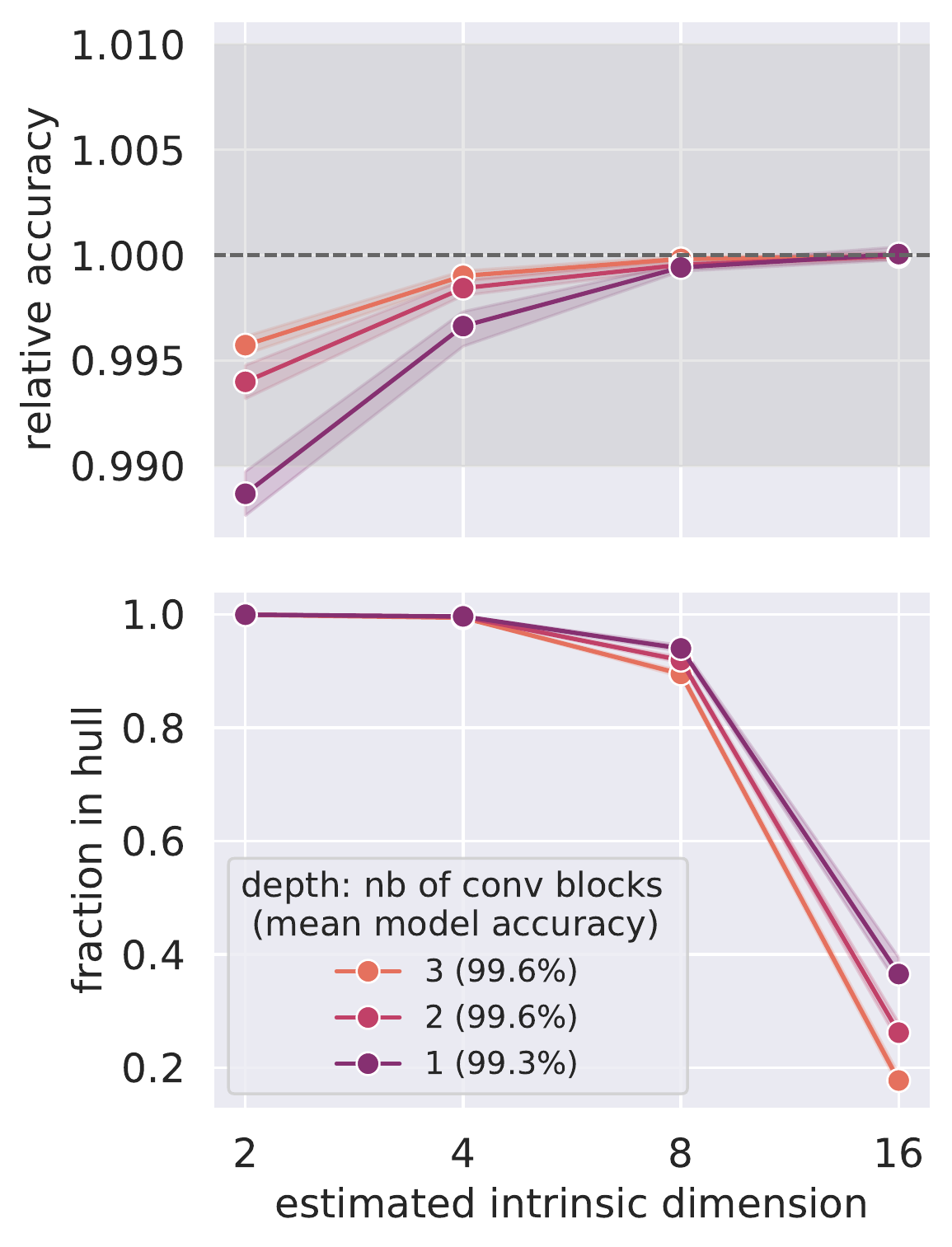}  
	\hfill
	\textbf{C}
	\hspace{-0.3cm}
	\includegraphics[width=0.22\linewidth,valign=T]{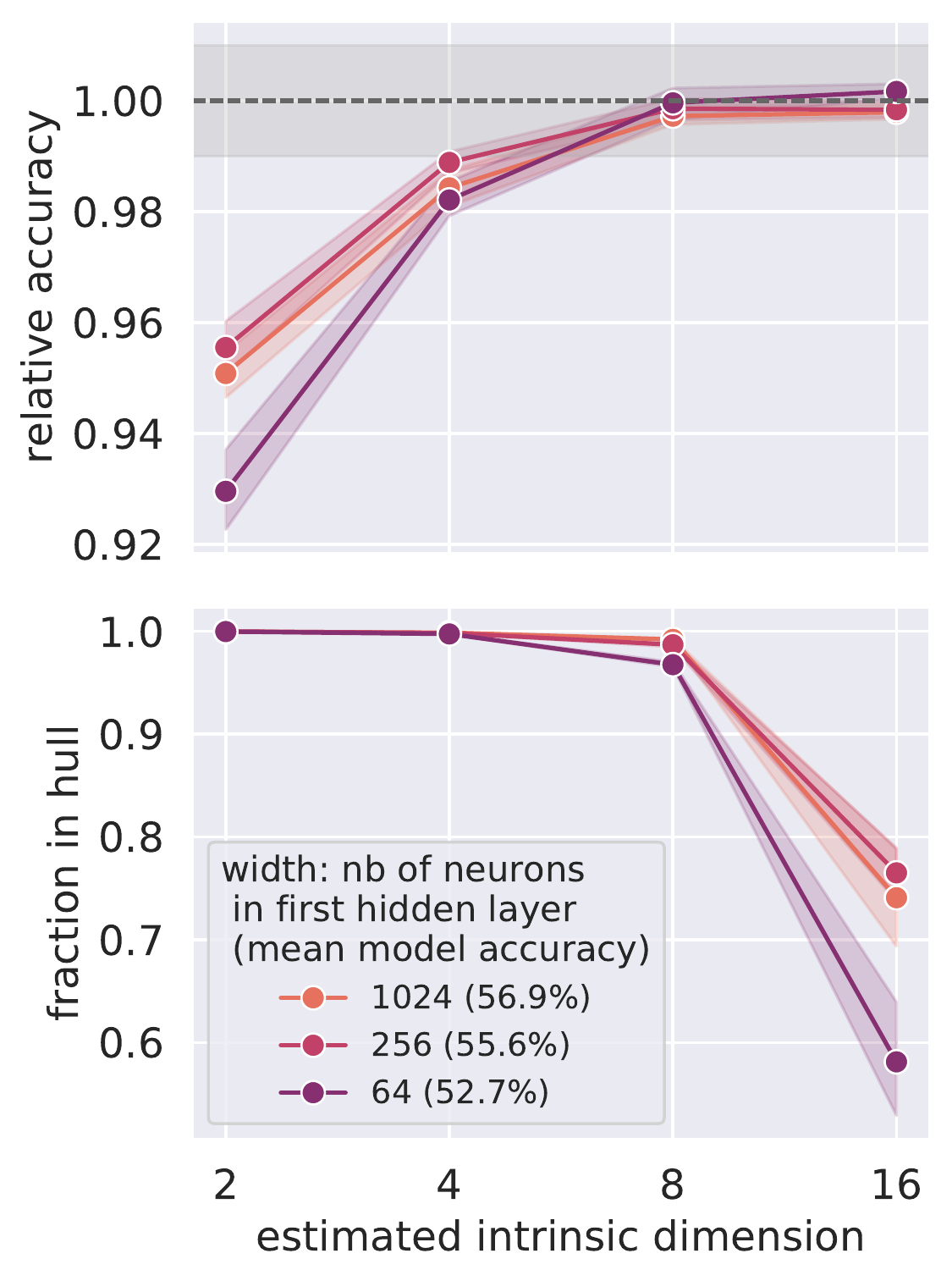}  
	\hfill
	\textbf{D}
	\hspace{-0.3cm}
	\includegraphics[width=0.22\linewidth,valign=T]{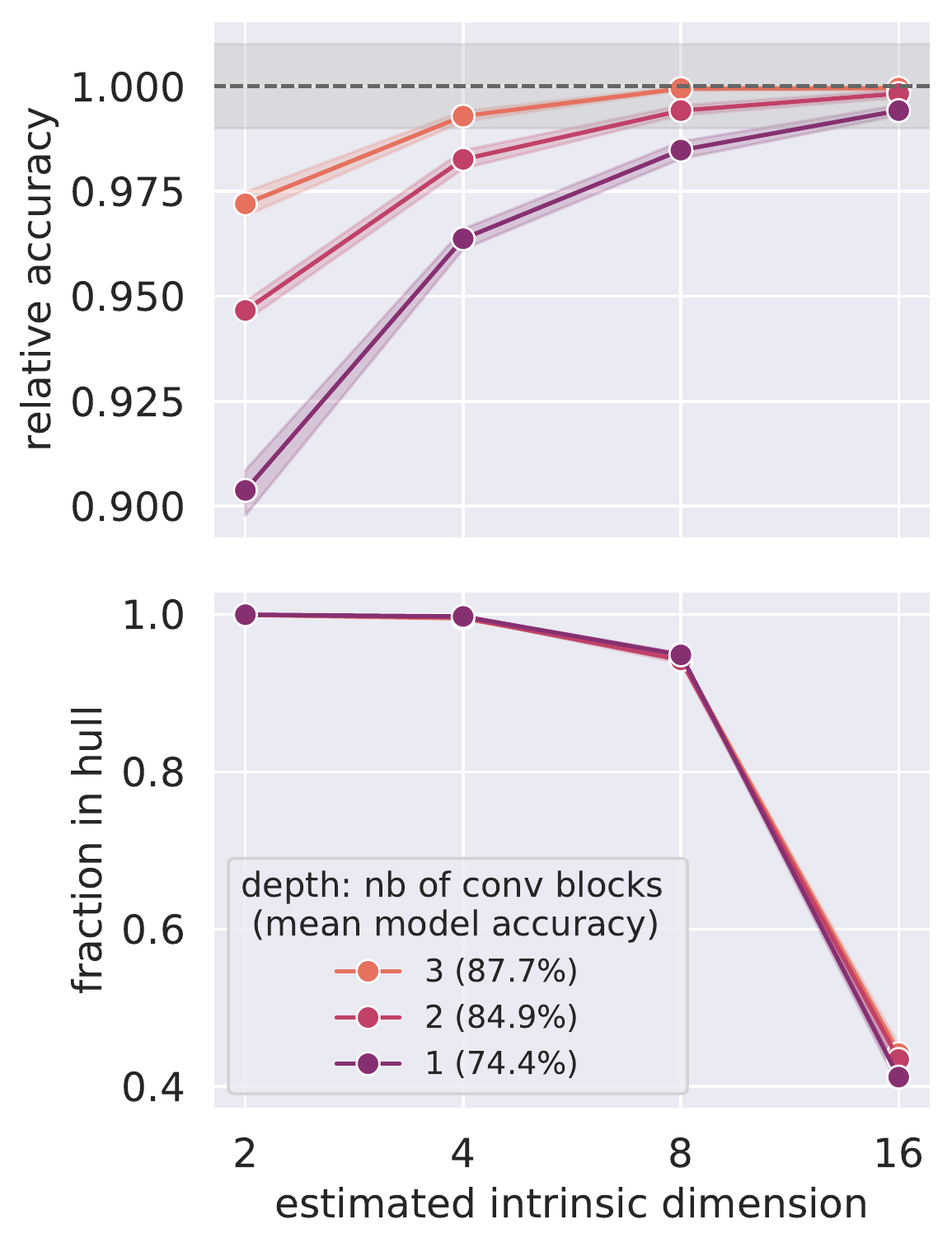} 
	\caption{\textbf{Estimate of the intrinsic dimension of the neural space along with the percentage of test samples in the convex hull of training set.} Experiment with different data sets, MNIST (A, B) and CIFAR-10 (C, D), and different architectures, multilayer perceptrons (A, C) and convolutional neural network (B, D). The color code goes from violet to yellow with increasing performance. The gray filled area represents a 1\% deviation from the model mean accuracy. The x-axis is in log2 scale.}
	\label{fig:id_ch}
\end{figure}

Figure~\ref{fig:id_ch} presents the results of the investigation on the latent space using the autoencoder method described in the previous section. All the results converge to the same findings. First and foremost, the latent space underlying the neural activities is of low dimension, with 8 or even 4 dimensions being enough to capture these activities sufficiently well so as to yield the same classification performance. These figures are actually in agreement with the estimations of intrinsic dimensions found for these two data sets by \citet{ma2018dimensionality} and \citet{ansuini2019intrinsic}, using different techniques for this estimation. Moreover, the better the model the lower the intrinsic dimension of its neural space, which echoes the result found by \citet{ansuini2019intrinsic}. If we compare the results yield by the two data sets, it seems that the intrinsic dimension of the neural space is larger for CIFAR-10 than for MNIST, in agreement with the intuition one can have with respect to the complexity of natural images vs. handwritten digits.\\
Second, the proportion of the test set within the convex hull of the training set decreases significantly with the number of dimensions, as expected by the famous curse of dimensionality \citep{balestriero2021learning}. But interpolation is not ``doomed'': given the low intrinsic dimension of the neural space, the vast majority of the test samples actually lie within the convex hull of the training set. Hence, adopting the definition of interpolation proposed by \citet{balestriero2021learning}, the model actually operate in interpolation regime, although one has to look at the latent space that underlies the neural activities.\\

\subsection{Local proximity to the training set}
\label{sec:results_dist}
\begin{figure}
	\centering
	\begin{minipage}[t]{0.28\linewidth}	
	\textbf{A}
	\hspace{-0.3cm}
	\includegraphics[width=.95\linewidth,valign=T]{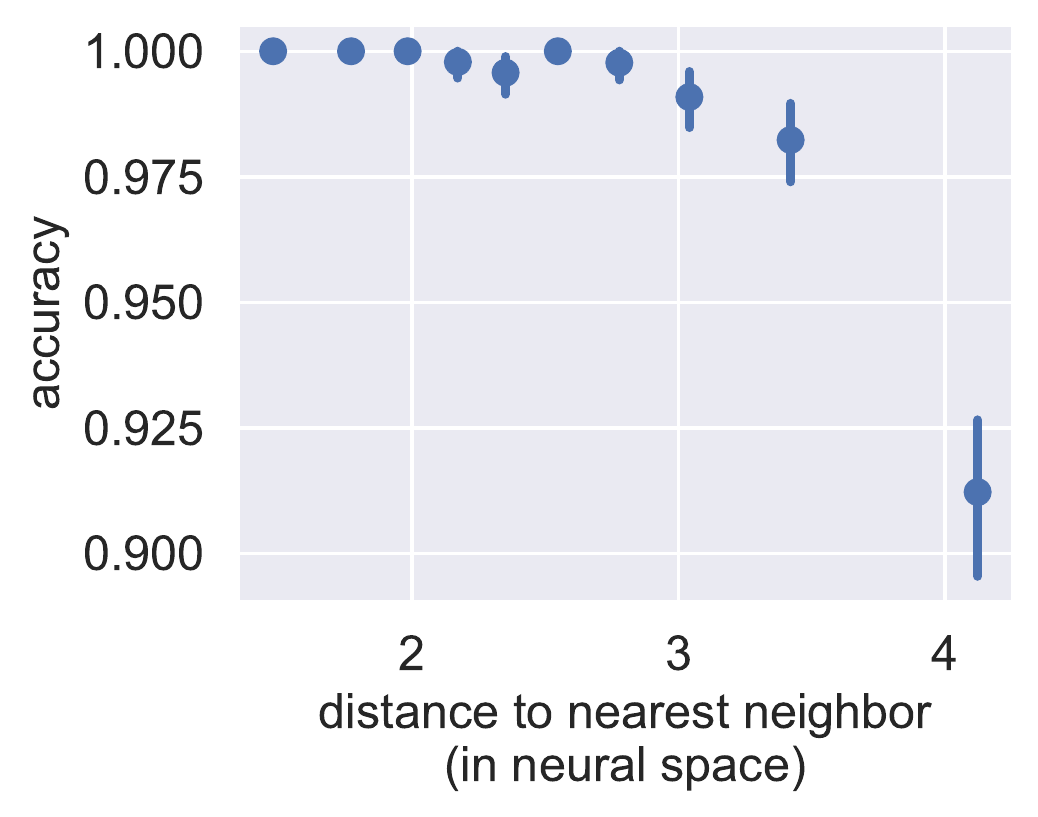}\\[20pt]
	\textbf{B}
	\hspace{-0.3cm}
	\includegraphics[width=.95\linewidth,valign=T]{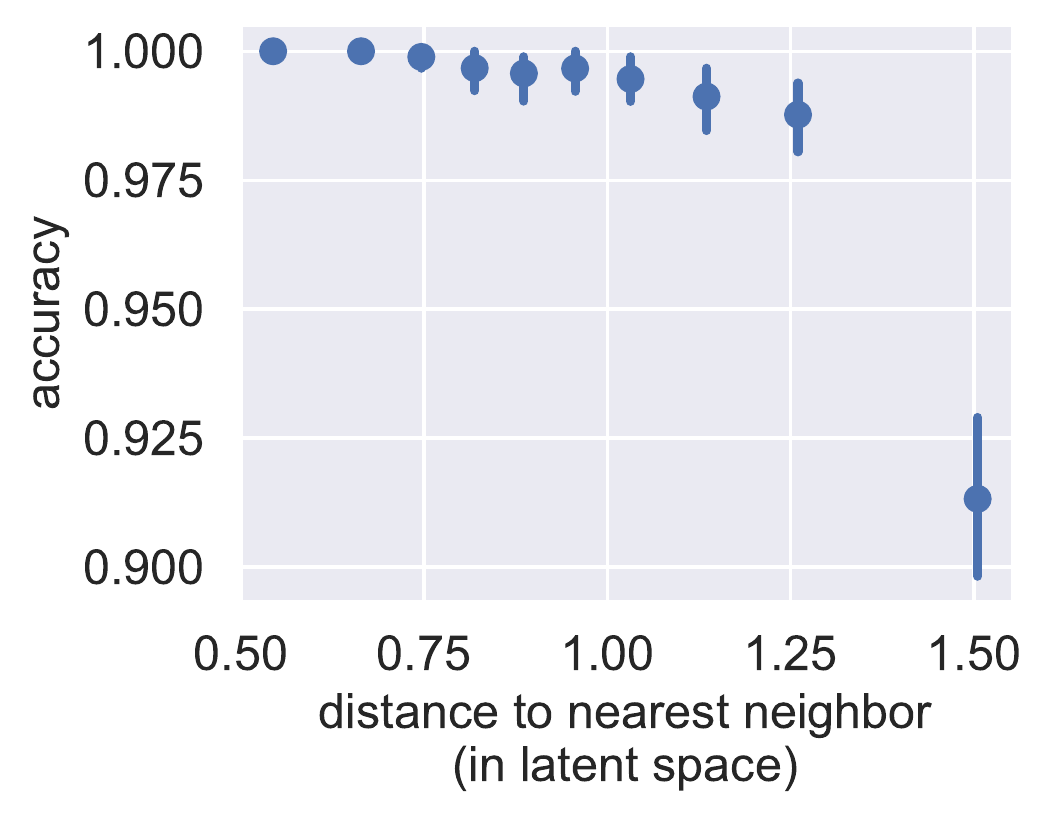} 
	\end{minipage}
	\hfill
	\begin{minipage}[t]{0.28\linewidth}	
	\textbf{C}
	\hspace{-0.5cm}\\	
	\includegraphics[width=0.9\linewidth,valign=T]{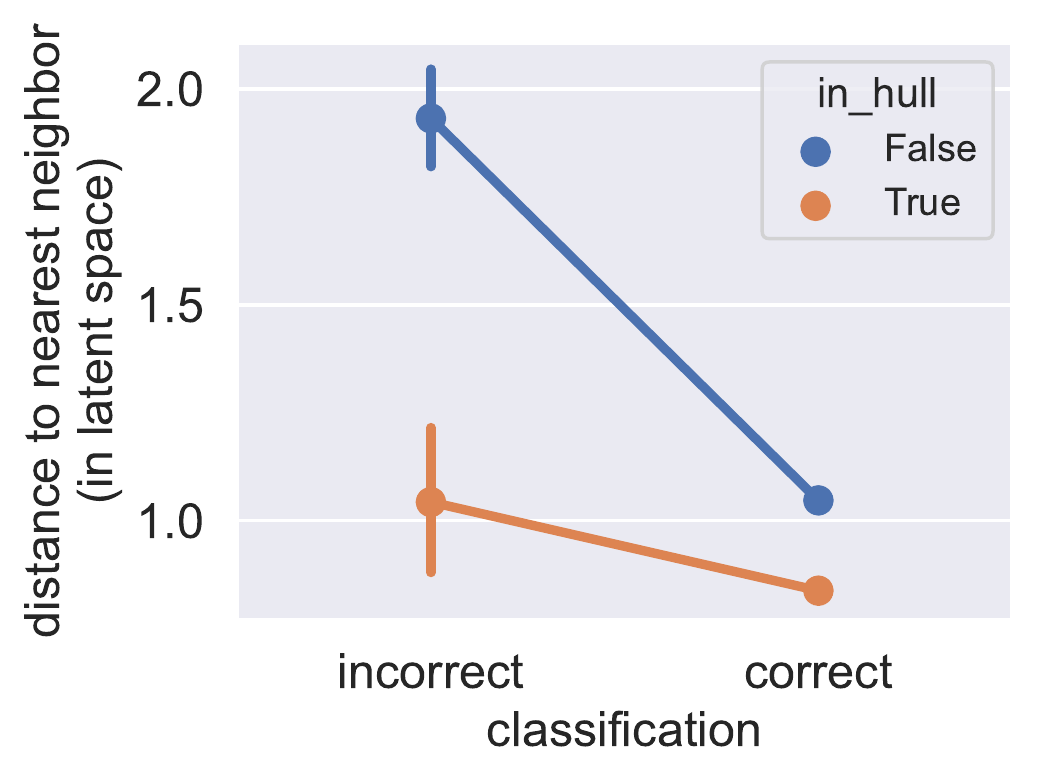}\\
	\textbf{D} 
	\hspace{-0.5cm}
	\includegraphics[width=0.9\linewidth,valign=T]{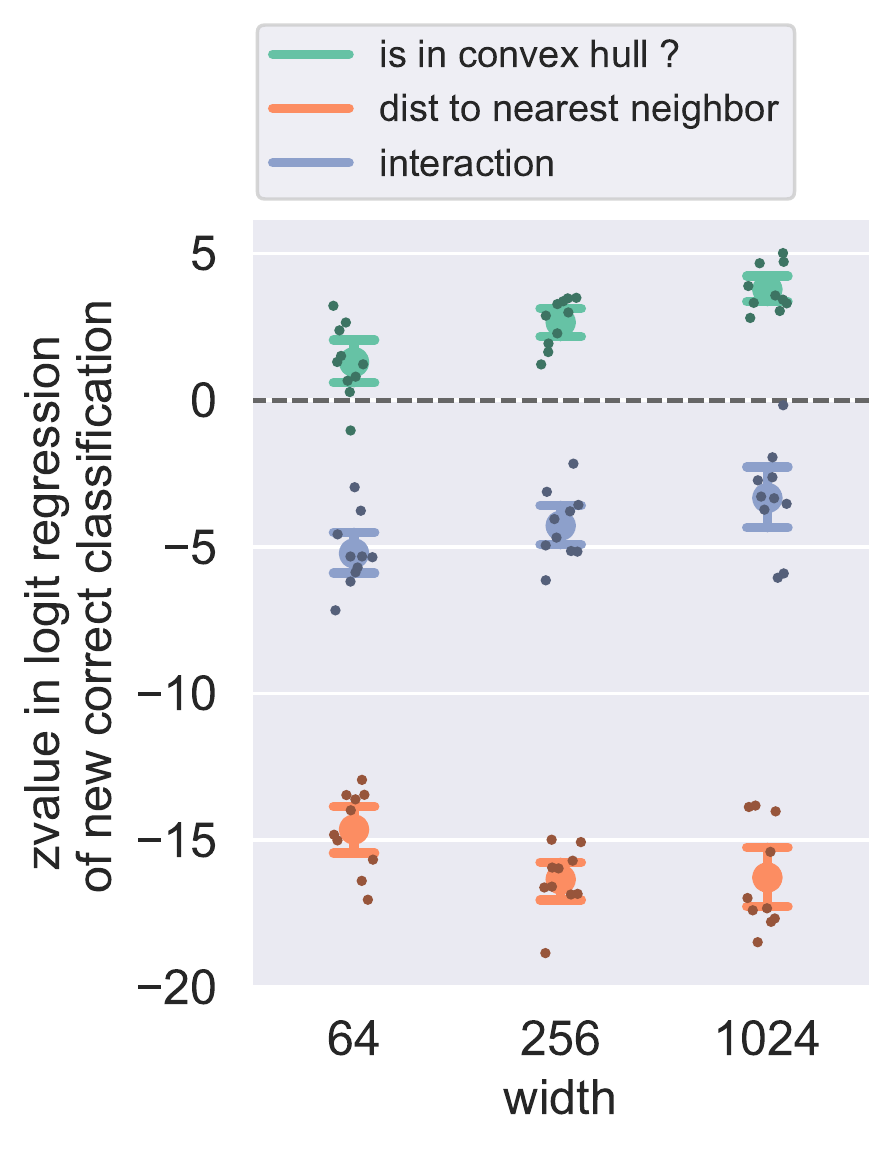} 
	\end{minipage}
	\hfill	
	\textbf{E}
	\hspace{-0.3cm}
	\includegraphics[width=0.32\linewidth,valign=T]{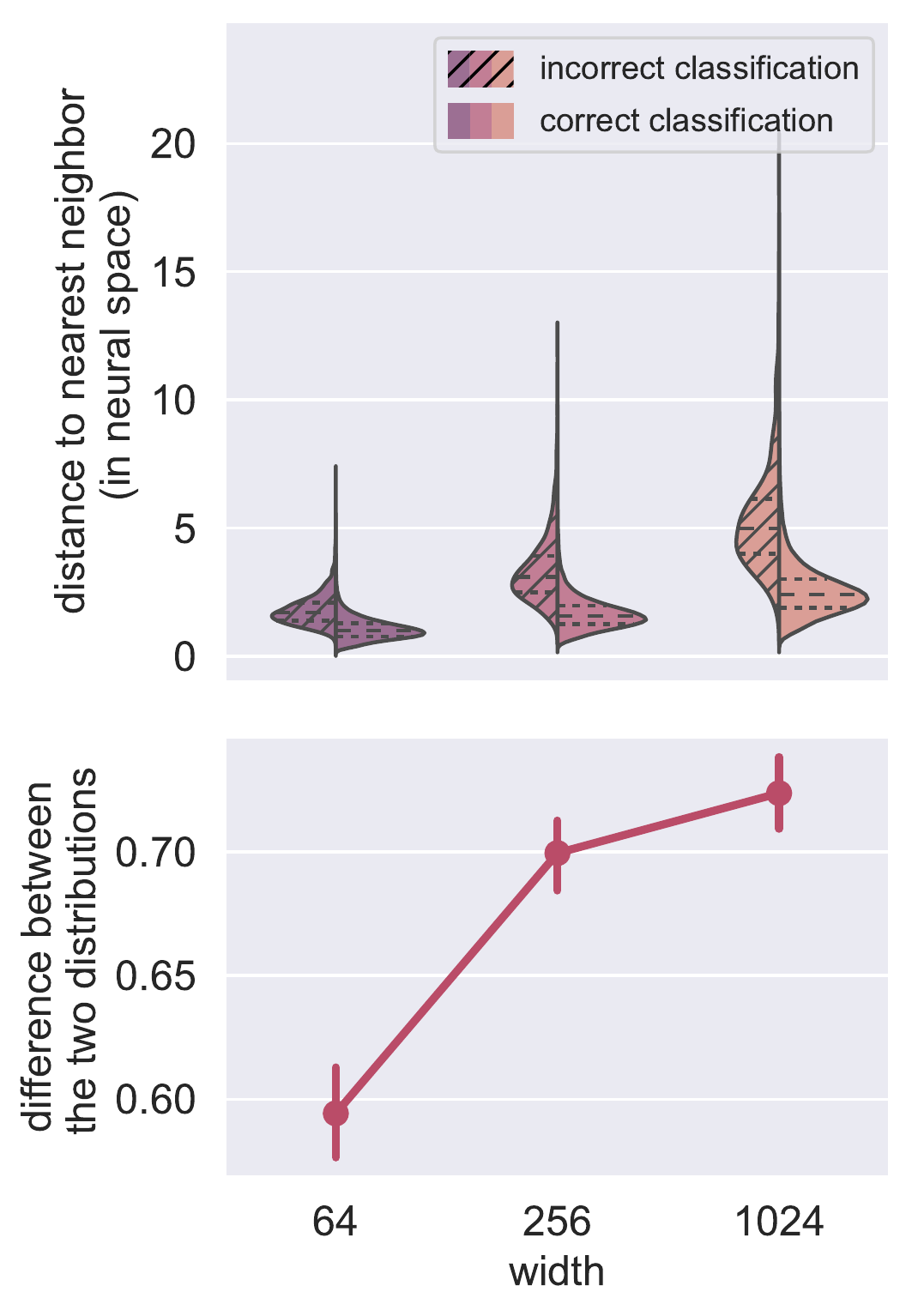} 

	\caption{\textbf{Distance to training set is indicative of generalization performance.} Results for the MNIST data set and the three multilayer perceptrons considered (see Fig.~\ref{fig:ann}A and Fig.~\ref{fig:id_ch}A). See Supplementary Figures for other data sets, architectures, and definitions of distance. The distance used here corresponds to the Euclidean distance of a new test sample to the nearest neighbor in the training set. (A) Accuracy as a function of the distance to the training set, in neural space, for the best network (the one with 1024 units in the first layer), for the first trial. Each bin represents 10\% of the test data, while error bars indicate 95\% bootstrap confidence intervals. (B) Same, in latent space. The latent space is 16 dimensional here (for the other lower values, almost all the samples lie in the convex hull). (C) Average distance to training set as a function of whether a new test samples is correctly classified or not (x-axis), and depending on whether it lies inside ou outside the convex hull of the training set (orange vs. blue). (D) Results of the logistic regression \texttt{accuracy $\sim$ distance + in\_hull + distance:in\_hull}. zvalues (which provides the significance of the corresponding coefficient, a zvalue of 1.96 for instance corresponding to a p-value of 0.05) of the coefficient for each of the two factors and their interaction. Each dot corresponds to a given trial, error bars indicate 95\% bootstrap confidence intervals over ten trials. (E) In neural space, (top) distribution of the distance of a test sample to the training set, depending whether the sample is correctly or incorrectly (dashed) classified, for the three multilayer perceptrons considered, and (bottom) for each case the corresponding difference between these two distributions, computed thanks to the Kolmogorov-Smirnov statistic. Error bars indicate 95\% bootstrap confidence intervals over ten trials.}
	\label{fig:dist}
\end{figure}

In Figure~\ref{fig:id_ch}, for a given estimated intrinsic dimension, no obvious relationship emerges from looking at the association between generalization performance and percentage of test set in training convex hull. One can ask whether the definition of interpolation as belonging to the convex hull is the most appropriate. \citet{chollet2021deep} rather talks about \textit{local generalization} (see also the related blog page at \url{https://blog.keras.io/the-limitations-of-deep-learning.html}): performances do not deteriorate abruptly as soon as you get outside the convex hull, and conversely, a new sample might lie within the convex hull but be relatively far from any training point and then more prone to error in classification, if the model is essentially in interpolative mode. Although with the 2, 4 and 8 dimensional cases almost all the test samples fall within the convex hull of the training set, in the case of a 16 dimensional latent space, about half of the test samples fall within the convex hull. This gives the opportunity to study the relationship between the fact of falling within the convex hull and the probability of being correctly classified. All the following examples consider the 16 dimensional latent space only. We also look at different measures of proximity to the training set. Figure~\ref{fig:dist} presents the results of this investigation for the case of multilayer perceptrons trained on the MNIST data set (corresponding to Fig.~\ref{fig:id_ch}A), and with the use of the Euclidean distance. Supplementary figures present similar results for the other architectures and data sets presented in Fig.~\ref{fig:id_ch}, as well as a variety of distances, namely (nearest neighbor) Euclidean distance, cosine distance, and class conditional Euclidean distance (distance to nearest neighbor of the right class). All in all, the results can be summarized as the following (interestingly, for some distance measures the CIFAR-10/MLP case diverges from that, but it actually goes well with the main point as in this case these models perform quite poorly).\\

First, be it in neural space (Fig.~\ref{fig:dist}A) or in latent space (Fig.~\ref{fig:dist}B), the closer a new sample to the training set, the greater the chance of a correct classification. Points in the convex hull are closer to the training set than points outside, but in both cases, correctly classified new samples are closer to the training set than incorrectly classified ones (Fig.~\ref{fig:dist}C). If we try to predict the probability of being correctly classified by taking into account both the fact of being in the convex hull and the distance to the training set (and the interaction between these two factors), fitting a logistic regression show that the main contribution comes from the distance factor (Fig.~\ref{fig:dist}D), in the direction as expected: the closer a sample the more its chance of being correctly classified (note: in this case, Fig.~\ref{fig:dist}D, being in the convex hull increases the chance of being correctly classified, as might be expected, but the full range of cases presented in supplementary material show that this is not very reliable, contrary to the distance to training set). Finally, if we look directly in the neural space, we observe the same finding that the distance to the training set is smaller when the test samples are correctly classified (Fig.~\ref{fig:dist}E, top), and this effect is actually stronger the better the model (as quantified by the distance between the distributions of distances for the incorrectly vs. correctly classified new items). Thus, this behavior is not surprising and is very typical of pure interpolation regime. In the end, these results fit well with the ``local generalization'' picture given by \citet{chollet2021deep}.

\section{Conclusion}

We saw that the neural activities of the feature space constructed by a neural network actually live in a much smaller dimensional space, called the intrinsic latent space. In this latent space, the vast majority of new samples from the test set lie within the convex hull defined by the training set, although this is not the case in the high dimensional neural state space, as discussed in the Introduction and Figure~\ref{fig:convexhull}. If we first go with the convex hull definition of interpolation proposed by \citet{balestriero2021learning}, this means that these neural networks actually operate in interpolation regime, contrary to the claims made by these authors. Moreover, considering cases with higher dimensional latent space, where all test samples do not necessarily fall within the convex hull, we found that the probability of misclassifying a new sample is actually not well predicted by whether it lies within the convex hull or not. Even within the convex hull, if one test sample is far from the training set, its probability of being misclassified increases with this distance. Conversely, a point outside the hull but in close proximity to the training set is likely to be well classified. To be in the convex hull or not is not important: what seems to matter is the distance to the training set. As soon as one new sample gets far from the training data, be it within the convex hull or not, the performance declines---an example of the \textit{local generalization} described by \citet{chollet2021deep}. This is typical of an interpolative regime, and is in stark contrast with the generalization abilities of human subjects \citep{lake2017building}. Hence, in spite of the high-dimensionality of the data, the notion of distance to the training set remains an important aspect that governs the generalization performance of an artificial neural network (at least for common architectures like multilayer perceptrons or convolutional neural networks). These artificial neural networks face difficulties extrapolating beyond a local area close to the training set. The models that generalize best are the ones that manage to place themselves in interpolative mode, with a representation where new samples lie close to the training set. 

\bibliographystyle{apalike}
\bibliography{refs_nn_interpolation}

\begin{thebibliography}{}

\bibitem[Abadi et~al., 2015]{tensorflow2015-whitepaper}
Abadi, M., Agarwal, A., Barham, P., Brevdo, E., Chen, Z., Citro, C., Corrado,
  G.~S., Davis, A., Dean, J., Devin, M., Ghemawat, S., Goodfellow, I., Harp,
  A., Irving, G., Isard, M., Jia, Y., Jozefowicz, R., Kaiser, L., Kudlur, M.,
  Levenberg, J., Man\'{e}, D., Monga, R., Moore, S., Murray, D., Olah, C.,
  Schuster, M., Shlens, J., Steiner, B., Sutskever, I., Talwar, K., Tucker, P.,
  Vanhoucke, V., Vasudevan, V., Vi\'{e}gas, F., Vinyals, O., Warden, P.,
  Wattenberg, M., Wicke, M., Yu, Y., and Zheng, X. (2015).
\newblock {TensorFlow}: Large-scale machine learning on heterogeneous systems.
\newblock Software available from tensorflow.org.

\bibitem[Ansuini et~al., 2019]{ansuini2019intrinsic}
Ansuini, A., Laio, A., Macke, J.~H., and Zoccolan, D. (2019).
\newblock Intrinsic dimension of data representations in deep neural networks.
\newblock In Wallach, H., Larochelle, H., Beygelzimer, A., d\textquotesingle
  Alch\'{e}-Buc, F., Fox, E., and Garnett, R., editors, {\em Advances in Neural
  Information Processing Systems}, volume~32. Curran Associates, Inc.

\bibitem[Archer et~al., 2014]{archer2014low}
Archer, E.~W., Koster, U., Pillow, J.~W., and Macke, J.~H. (2014).
\newblock Low-dimensional models of neural population activity in sensory
  cortical circuits.
\newblock {\em Advances in neural information processing systems}, 27.

\bibitem[Balestriero et~al., 2021]{balestriero2021learning}
Balestriero, R., Pesenti, J., and LeCun, Y. (2021).
\newblock Learning in high dimension always amounts to extrapolation.
\newblock {\em arXiv preprint arXiv:2110.09485}.

\bibitem[Barnard and Wessels, 1992]{barnard1992extrapolation}
Barnard, E. and Wessels, L. (1992).
\newblock Extrapolation and interpolation in neural network classifiers.
\newblock {\em IEEE Control Systems Magazine}, 12(5):50--53.

\bibitem[Barrett et~al., 2018]{barrett2018measuring}
Barrett, D., Hill, F., Santoro, A., Morcos, A., and Lillicrap, T. (2018).
\newblock Measuring abstract reasoning in neural networks.
\newblock In {\em International conference on machine learning}, pages
  511--520. PMLR.

\bibitem[Cayton, 2005]{cayton2005algorithms}
Cayton, L. (2005).
\newblock Algorithms for manifold learning.
\newblock {\em Univ. of California at San Diego Tech. Rep}, 12(1-17):1.

\bibitem[Chollet, 2021]{chollet2021deep}
Chollet, F. (2021).
\newblock {\em Deep Learning with Python, Second Edition}.
\newblock Manning.

\bibitem[Chollet et~al., 2015]{chollet2015keras}
Chollet, F. et~al. (2015).
\newblock Keras.
\newblock \url{https://keras.io}.

\bibitem[Chung and Abbott, 2021]{chung2021neural}
Chung, S. and Abbott, L. (2021).
\newblock Neural population geometry: An approach for understanding biological
  and artificial neural networks.
\newblock {\em Current opinion in neurobiology}, 70:137--144.

\bibitem[Cunningham and Byron, 2014]{cunningham2014dimensionality}
Cunningham, J.~P. and Byron, M.~Y. (2014).
\newblock Dimensionality reduction for large-scale neural recordings.
\newblock {\em Nature neuroscience}, 17(11):1500--1509.

\bibitem[Gallego et~al., 2017]{gallego2017neural}
Gallego, J.~A., Perich, M.~G., Miller, L.~E., and Solla, S.~A. (2017).
\newblock Neural manifolds for the control of movement.
\newblock {\em Neuron}, 94(5):978--984.

\bibitem[Haley and Soloway, 1992]{haley1992extrapolation}
Haley, P.~J. and Soloway, D. (1992).
\newblock Extrapolation limitations of multilayer feedforward neural networks.
\newblock In {\em [Proceedings 1992] IJCNN International Joint Conference on
  Neural Networks}, volume~4, pages 25--30. IEEE.

\bibitem[Harris et~al., 2020]{harris2020array}
Harris, C.~R., Millman, K.~J., van~der Walt, S.~J., Gommers, R., Virtanen, P.,
  Cournapeau, D., Wieser, E., Taylor, J., Berg, S., Smith, N.~J., et~al.
  (2020).
\newblock Array programming with numpy.
\newblock {\em Nature}, 585(7825):357--362.

\bibitem[Henry et~al., 1974]{henry1974orientation}
Henry, G., Dreher, B., and Bishop, P. (1974).
\newblock Orientation specificity of cells in cat striate cortex.
\newblock {\em Journal of Neurophysiology}, 37:1394--1409.

\bibitem[Hinton and Salakhutdinov, 2006]{hinton2006reducing}
Hinton, G.~E. and Salakhutdinov, R.~R. (2006).
\newblock Reducing the dimensionality of data with neural networks.
\newblock {\em science}, 313(5786):504--507.

\bibitem[Hubel and Wiesel, 1959]{hubel1959receptive}
Hubel, D.~H. and Wiesel, T.~N. (1959).
\newblock Receptive fields of single neurones in the cat's striate cortex.
\newblock {\em The Journal of physiology}, 148(3):574--591.

\bibitem[Hunter, 2007]{hunter2007matplotlib}
Hunter, J.~D. (2007).
\newblock Matplotlib: A 2d graphics environment.
\newblock {\em Computing in Science \& Engineering}, 9(3):90--95.

\bibitem[Jazayeri and Ostojic, 2021]{jazayeri2021interpreting}
Jazayeri, M. and Ostojic, S. (2021).
\newblock Interpreting neural computations by examining intrinsic and embedding
  dimensionality of neural activity.
\newblock {\em Current opinion in neurobiology}, 70:113--120.

\bibitem[Kingma and Ba, 2015]{kingma2015adam}
Kingma, D.~P. and Ba, J. (2015).
\newblock Adam: A method for stochastic optimization.
\newblock In {\em Proceedings of the 3rd International Conference on Learning
  Representations (ICLR)}.

\bibitem[Krizhevsky, 2009]{krizhevsky2009learning}
Krizhevsky, A. (2009).
\newblock Learning multiple layers of features from tiny images.

\bibitem[Lake and Baroni, 2018]{lake2018generalization}
Lake, B. and Baroni, M. (2018).
\newblock Generalization without systematicity: On the compositional skills of
  sequence-to-sequence recurrent networks.
\newblock In {\em International conference on machine learning}, pages
  2873--2882. PMLR.

\bibitem[Lake et~al., 2017]{lake2017building}
Lake, B.~M., Ullman, T.~D., Tenenbaum, J.~B., and Gershman, S.~J. (2017).
\newblock Building machines that learn and think like people.
\newblock {\em Behavioral and brain sciences}, 40.

\bibitem[LeCun et~al., 2015]{lecun2015deep}
LeCun, Y., Bengio, Y., and Hinton, G. (2015).
\newblock Deep learning.
\newblock {\em Nature}, 521(7553):436--444.

\bibitem[LeCun et~al., 1998]{lecun1998gradient}
LeCun, Y., Bottou, L., Bengio, Y., and Haffner, P. (1998).
\newblock Gradient-based learning applied to document recognition.
\newblock {\em Proceedings of the IEEE}, 86(11):2278--2324.

\bibitem[Ma et~al., 2018]{ma2018dimensionality}
Ma, X., Wang, Y., Houle, M.~E., Zhou, S., Erfani, S., Xia, S., Wijewickrema,
  S., and Bailey, J. (2018).
\newblock Dimensionality-driven learning with noisy labels.
\newblock In {\em International Conference on Machine Learning}, pages
  3355--3364. PMLR.

\bibitem[Marcus, 2018]{marcus2018deep}
Marcus, G. (2018).
\newblock Deep learning: A critical appraisal.
\newblock {\em arXiv preprint arXiv:1801.00631}.

\bibitem[Marcus, 1998]{marcus1998rethinking}
Marcus, G.~F. (1998).
\newblock Rethinking eliminative connectionism.
\newblock {\em Cognitive psychology}, 37(3):243--282.

\bibitem[McKinney et~al., 2010]{mckinney2010data}
McKinney, W. et~al. (2010).
\newblock Data structures for statistical computing in python.
\newblock In {\em Proceedings of the 9th Python in Science Conference}, volume
  445, pages 51--56. Austin, TX.

\bibitem[Pope et~al., 2021]{pope2021intrinsic}
Pope, P., Zhu, C., Abdelkader, A., Goldblum, M., and Goldstein, T. (2021).
\newblock The intrinsic dimension of images and its impact on learning.
\newblock {\em arXiv preprint arXiv:2104.08894}.

\bibitem[Roweis and Saul, 2000]{roweis2000nonlinear}
Roweis, S.~T. and Saul, L.~K. (2000).
\newblock Nonlinear dimensionality reduction by locally linear embedding.
\newblock {\em science}, 290(5500):2323--2326.

\bibitem[Sadtler et~al., 2014]{sadtler2014neural}
Sadtler, P.~T., Quick, K.~M., Golub, M.~D., Chase, S.~M., Ryu, S.~I.,
  Tyler-Kabara, E.~C., Byron, M.~Y., and Batista, A.~P. (2014).
\newblock Neural constraints on learning.
\newblock {\em Nature}, 512(7515):423--426.

\bibitem[Saxton et~al., 2019]{saxton2019analysing}
Saxton, D., Grefenstette, E., Hill, F., and Kohli, P. (2019).
\newblock Analysing mathematical reasoning abilities of neural models.
\newblock {\em arXiv preprint arXiv:1904.01557}.

\bibitem[Schmidhuber, 2015]{schmidhuber2015deep}
Schmidhuber, J. (2015).
\newblock Deep learning in neural networks: An overview.
\newblock {\em Neural networks}, 61:85--117.

\bibitem[Seabold and Perktold, 2010]{seabold2010statsmodels}
Seabold, S. and Perktold, J. (2010).
\newblock statsmodels: Econometric and statistical modeling with python.
\newblock In {\em 9th Python in Science Conference}.

\bibitem[Srivastava et~al., 2014]{srivastava2014dropout}
Srivastava, N., Hinton, G., Krizhevsky, A., Sutskever, I., and Salakhutdinov,
  R. (2014).
\newblock Dropout: a simple way to prevent neural networks from overfitting.
\newblock {\em The journal of machine learning research}, 15(1):1929--1958.

\bibitem[Taube et~al., 1990]{taube1990head}
Taube, J.~S., Muller, R.~U., and Ranck, J.~B. (1990).
\newblock Head-direction cells recorded from the postsubiculum in freely moving
  rats. i. description and quantitative analysis.
\newblock {\em Journal of Neuroscience}, 10(2):420--435.

\bibitem[Tenenbaum et~al., 2000]{tenenbaum2000global}
Tenenbaum, J.~B., De~Silva, V., and Langford, J.~C. (2000).
\newblock A global geometric framework for nonlinear dimensionality reduction.
\newblock {\em science}, 290(5500):2319--2323.

\bibitem[{Virtanen} et~al., 2020]{scipy}
{Virtanen}, P., {Gommers}, R., {Oliphant}, T.~E., {Haberland}, M., {Reddy}, T.,
  {Cournapeau}, D., {Burovski}, E., {Peterson}, P., {Weckesser}, W., {Bright},
  J., {van der Walt}, S.~J., {Brett}, M., {Wilson}, J., {Jarrod Millman}, K.,
  {Mayorov}, N., {Nelson}, A. R.~J., {Jones}, E., {Kern}, R., {Larson}, E.,
  {Carey}, C., {Polat}, {\.I}., {Feng}, Y., {Moore}, E.~W., {Vand erPlas}, J.,
  {Laxalde}, D., {Perktold}, J., {Cimrman}, R., {Henriksen}, I., {Quintero},
  E.~A., {Harris}, C.~R., {Archibald}, A.~M., {Ribeiro}, A.~H., {Pedregosa},
  F., {van Mulbregt}, P., and {Contributors}, S. .~. (2020).
\newblock {SciPy 1.0: Fundamental Algorithms for Scientific Computing in
  Python}.
\newblock {\em Nature Methods}, 17:261--272.

\bibitem[Waskom, 2021]{Waskom2021}
Waskom, M.~L. (2021).
\newblock seaborn: statistical data visualization.
\newblock {\em Journal of Open Source Software}, 6(60):3021.

\end{thebibliography}

\end{document}


\maketitle

\clearpage

\renewcommand{\thefigure}{S\arabic{figure}} 
\setcounter{figure}{0} 
\begin{figure}[h]
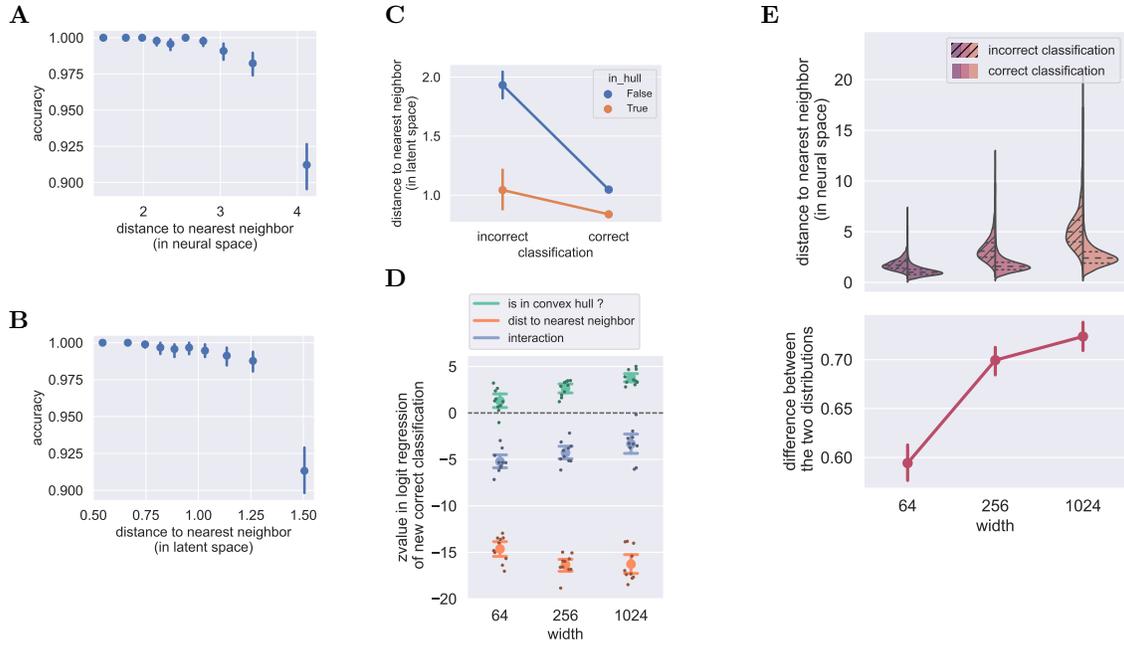

	\centering
	\begin{minipage}[t]{0.28\linewidth}	
		\textbf{A}
		\hspace{-0.3cm}
		\includegraphics[width=.95\linewidth,valign=T]{fig/mnist_mlp_width_acc_dist_r_euc}\\[20pt]
		\textbf{B}
		\hspace{-0.3cm}
		\includegraphics[width=.95\linewidth,valign=T]{fig/mnist_mlp_width_acc_dist_z16_euc} 
	\end{minipage}
	\hfill
	\begin{minipage}[t]{0.28\linewidth}	
		\textbf{C}
		\hspace{-0.5cm}\\	
		\includegraphics[width=0.9\linewidth,valign=T]{fig/mnist_mlp_width_acc_dist_hull_z16_euc}\\
		\textbf{D} 
		\hspace{-0.5cm}
		\includegraphics[width=0.9\linewidth,valign=T]{fig/mnist_mlp_width_zvalues_id16_euc} 
	\end{minipage}
	\hfill	
	\textbf{E}
	\hspace{-0.3cm}
	\includegraphics[width=0.32\linewidth,valign=T]{fig/mnist_mlp_width_dist_r_euc} 
	
	\caption{\textbf{Distance to training set is indicative of generalization performance.} Results for the MNIST dataset and the three multilayer perceptrons considered (see Fig.~3A and Fig.~4A). The distance used here corresponds to the Euclidean distance of a new test sample to the nearest neighbor in the training set. Exact reproduction of Figure~5.}
\end{figure}

\begin{figure}[!h]
	\centering
	\begin{minipage}[t]{0.28\linewidth}	
		\textbf{A}
		\hspace{-0.3cm}
		\includegraphics[width=.95\linewidth,valign=T]{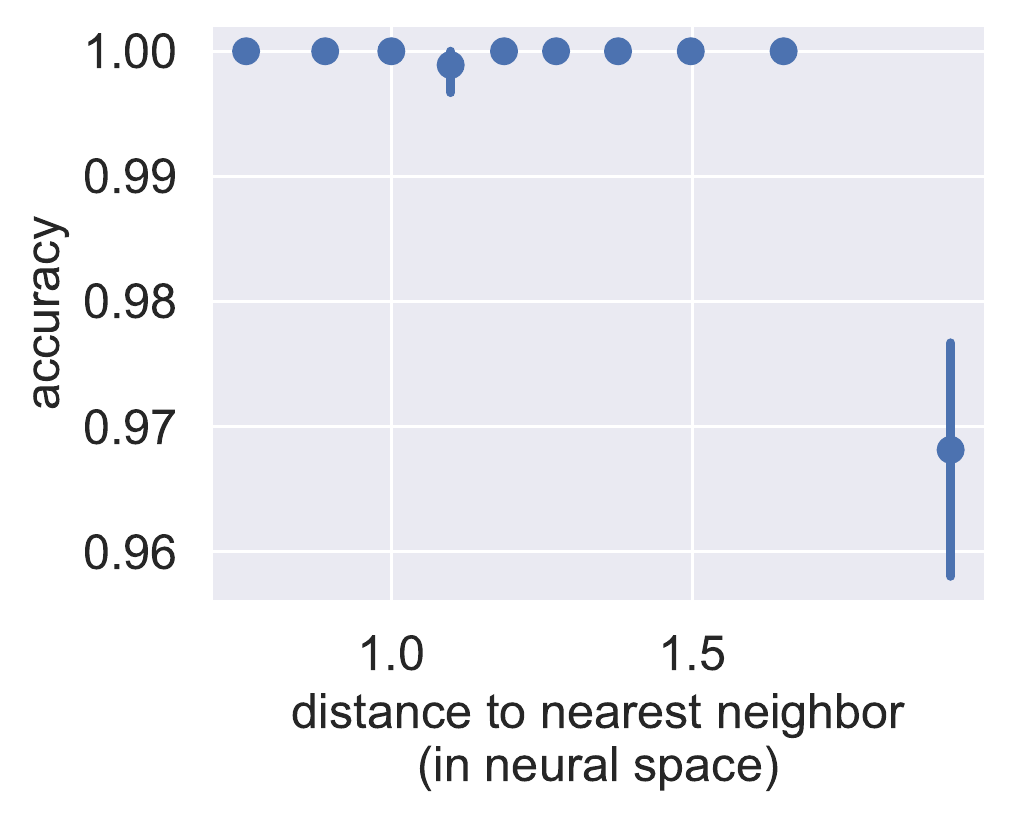}\\[20pt]
		\textbf{B}
		\hspace{-0.3cm}
		\includegraphics[width=.95\linewidth,valign=T]{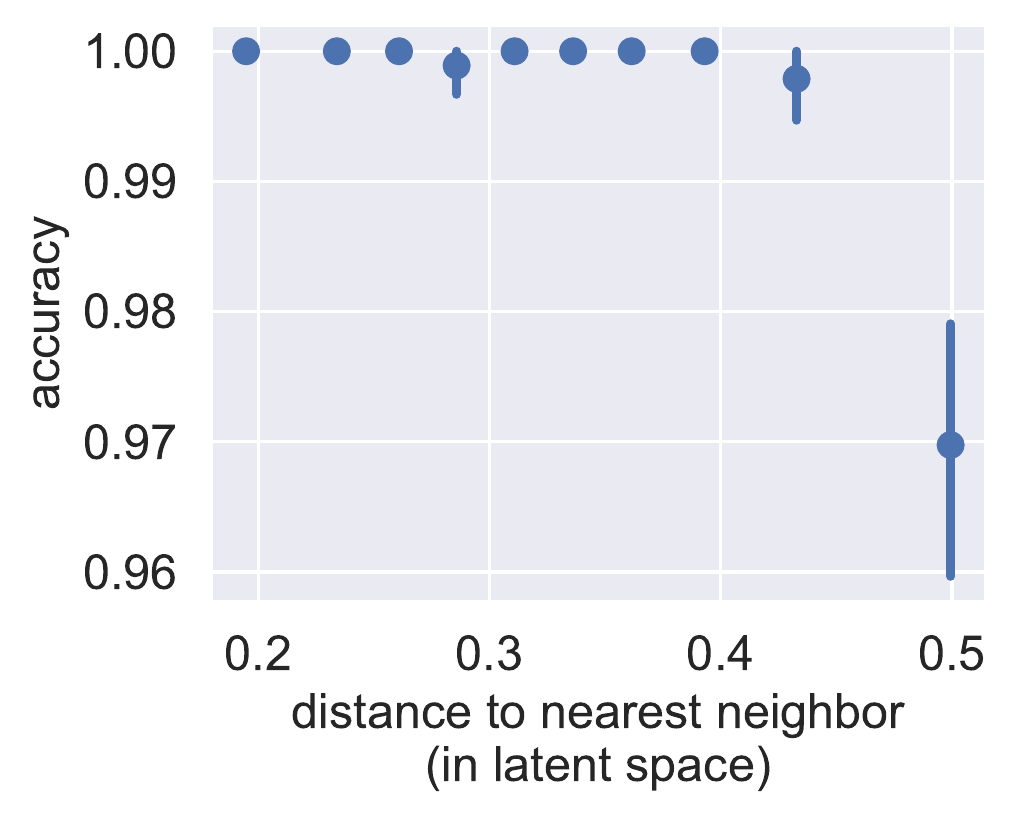} 
	\end{minipage}
	\hfill
	\begin{minipage}[t]{0.28\linewidth}	
		\textbf{C}
		\hspace{-0.5cm}\\	
		\includegraphics[width=0.9\linewidth,valign=T]{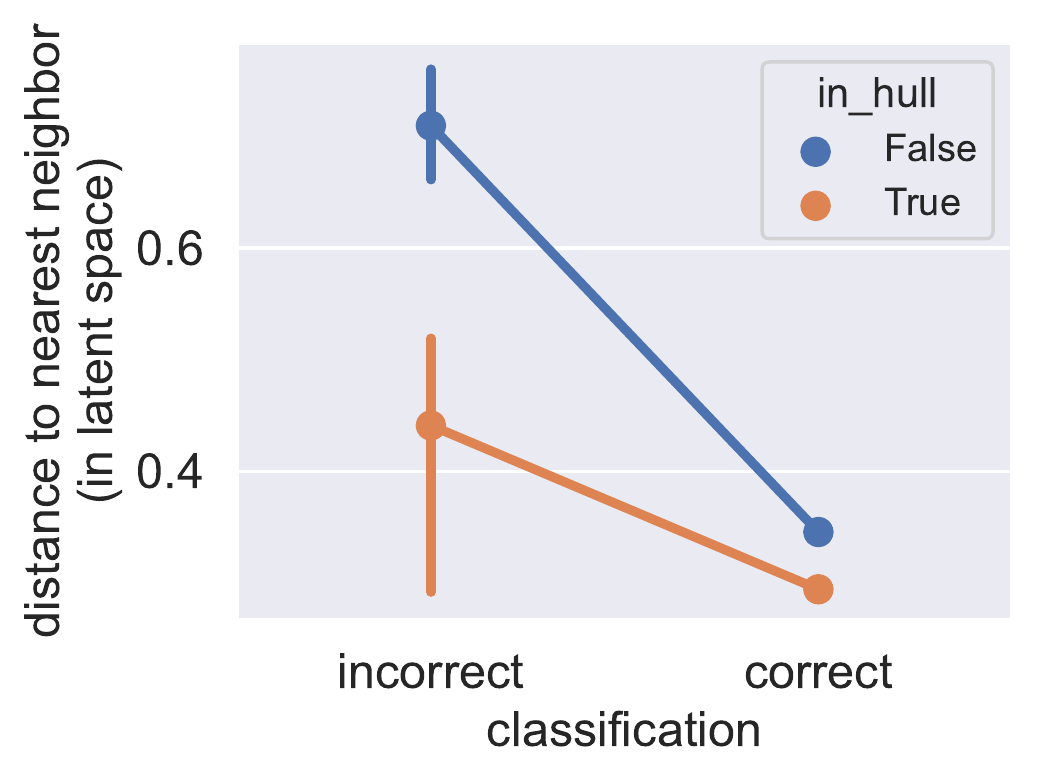}\\
		\textbf{D} 
		\hspace{-0.5cm}
		\includegraphics[width=0.9\linewidth,valign=T]{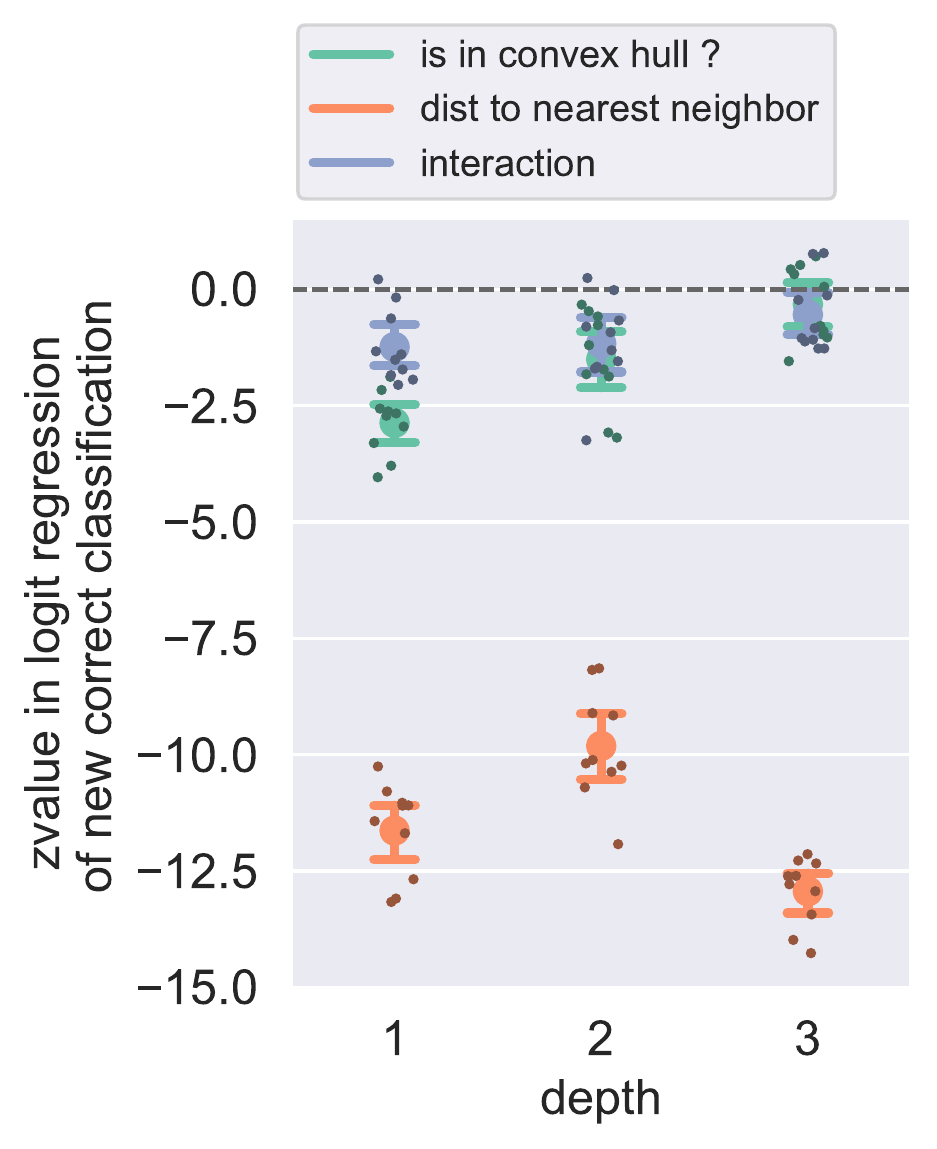} 
	\end{minipage}
	\hfill	
	\textbf{E}
	\hspace{-0.3cm}
	\includegraphics[width=0.32\linewidth,valign=T]{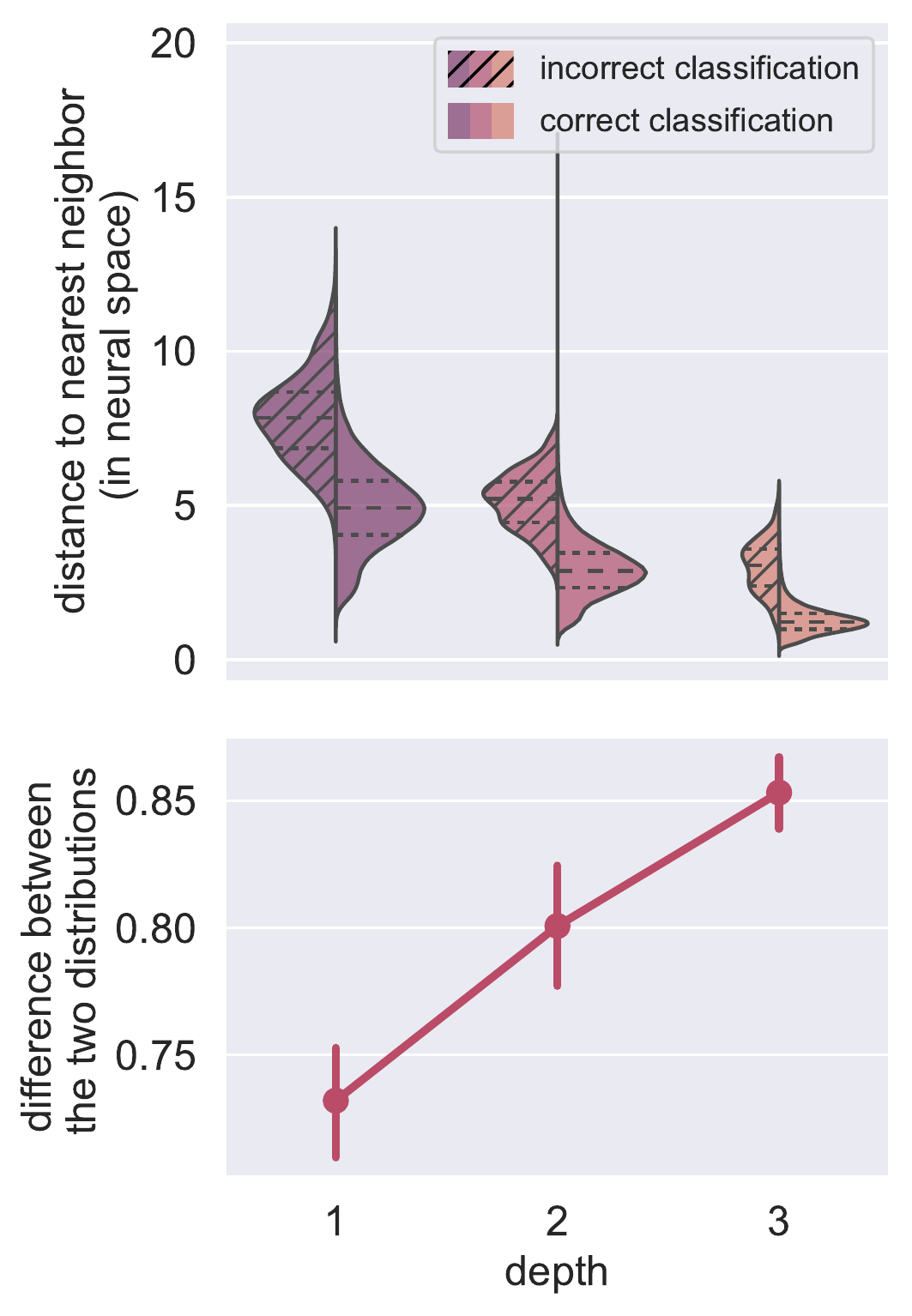} 
	
	\caption{\textbf{Distance to training set is indicative of generalization performance.} Results for the MNIST dataset and the three convolutional neural networks considered (see Fig.~3B and Fig.~4B). The distance used here corresponds to the Euclidean distance of a new test sample to the nearest neighbor in the training set. Otherwise, same legend as in Figure~5.}
\end{figure}

\begin{figure}
	\centering
	\begin{minipage}[t]{0.28\linewidth}	
		\textbf{A}
		\hspace{-0.3cm}
		\includegraphics[width=.95\linewidth,valign=T]{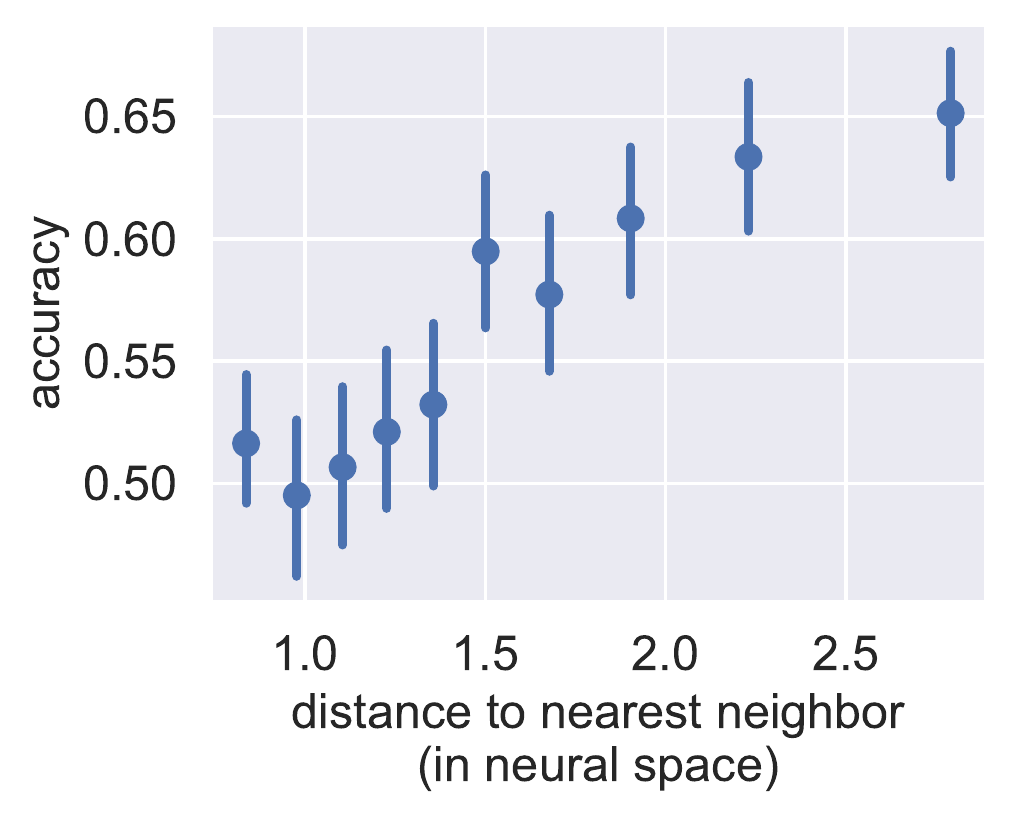}\\[20pt]
		\textbf{B}
		\hspace{-0.3cm}
		\includegraphics[width=.95\linewidth,valign=T]{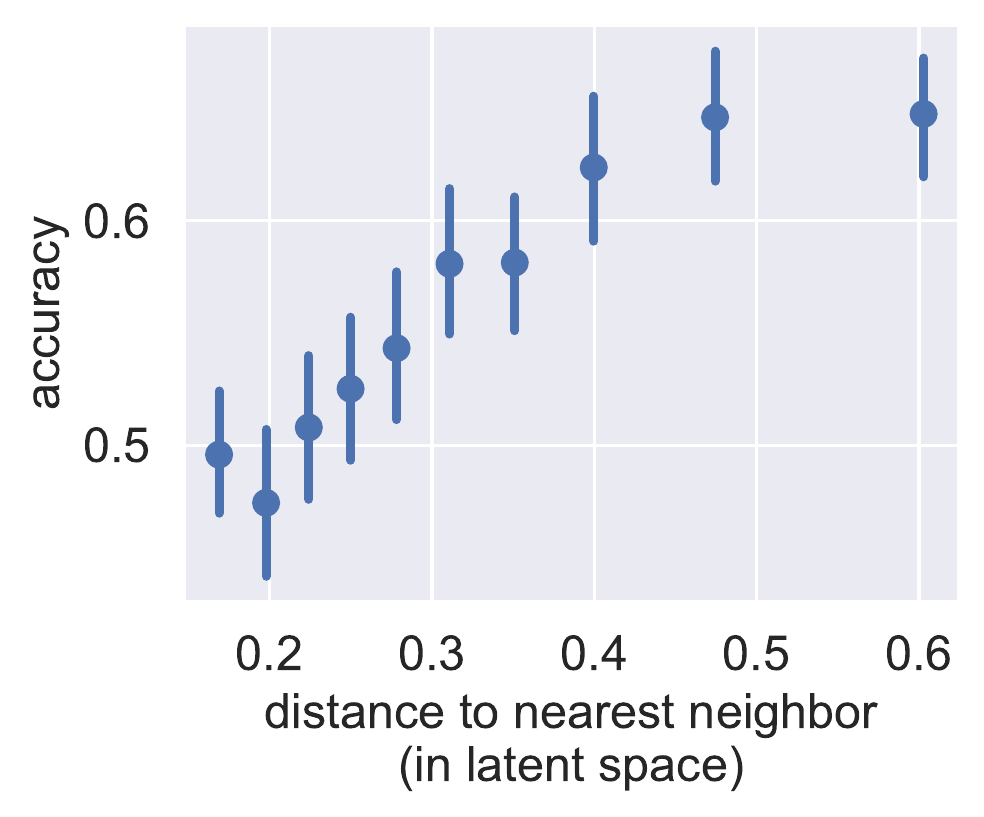} 
	\end{minipage}
	\hfill
	\begin{minipage}[t]{0.28\linewidth}	
		\textbf{C}
		\hspace{-0.5cm}\\	
		\includegraphics[width=0.9\linewidth,valign=T]{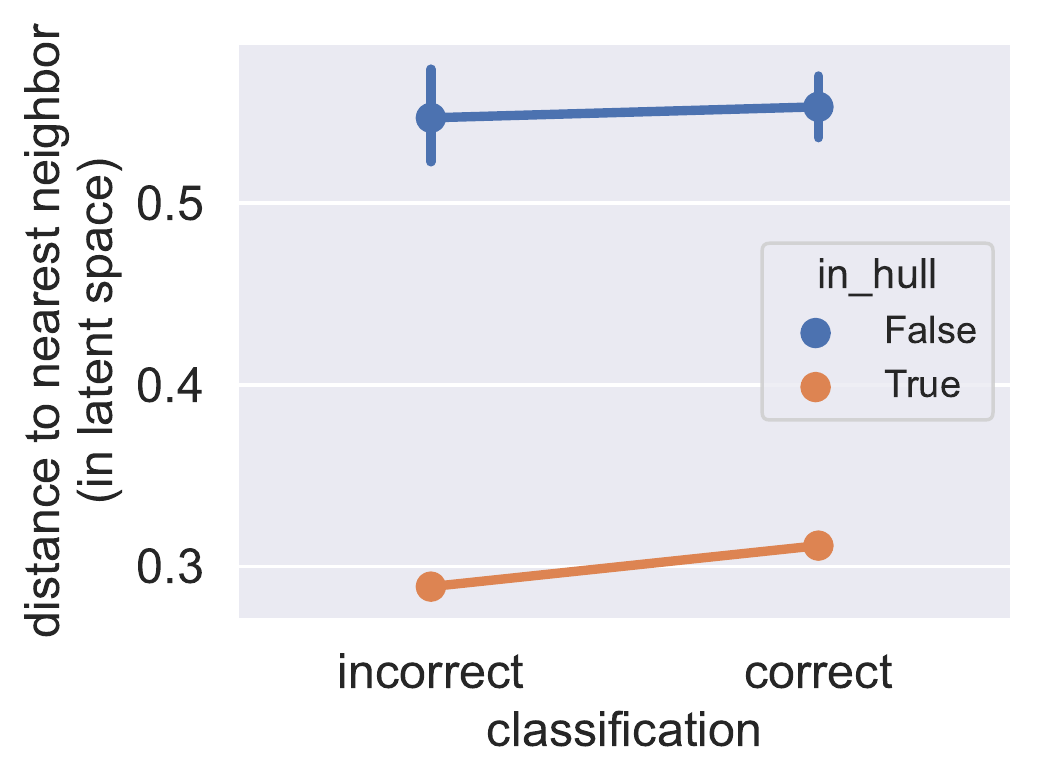}\\
		\textbf{D} 
		\hspace{-0.5cm}
		\includegraphics[width=0.9\linewidth,valign=T]{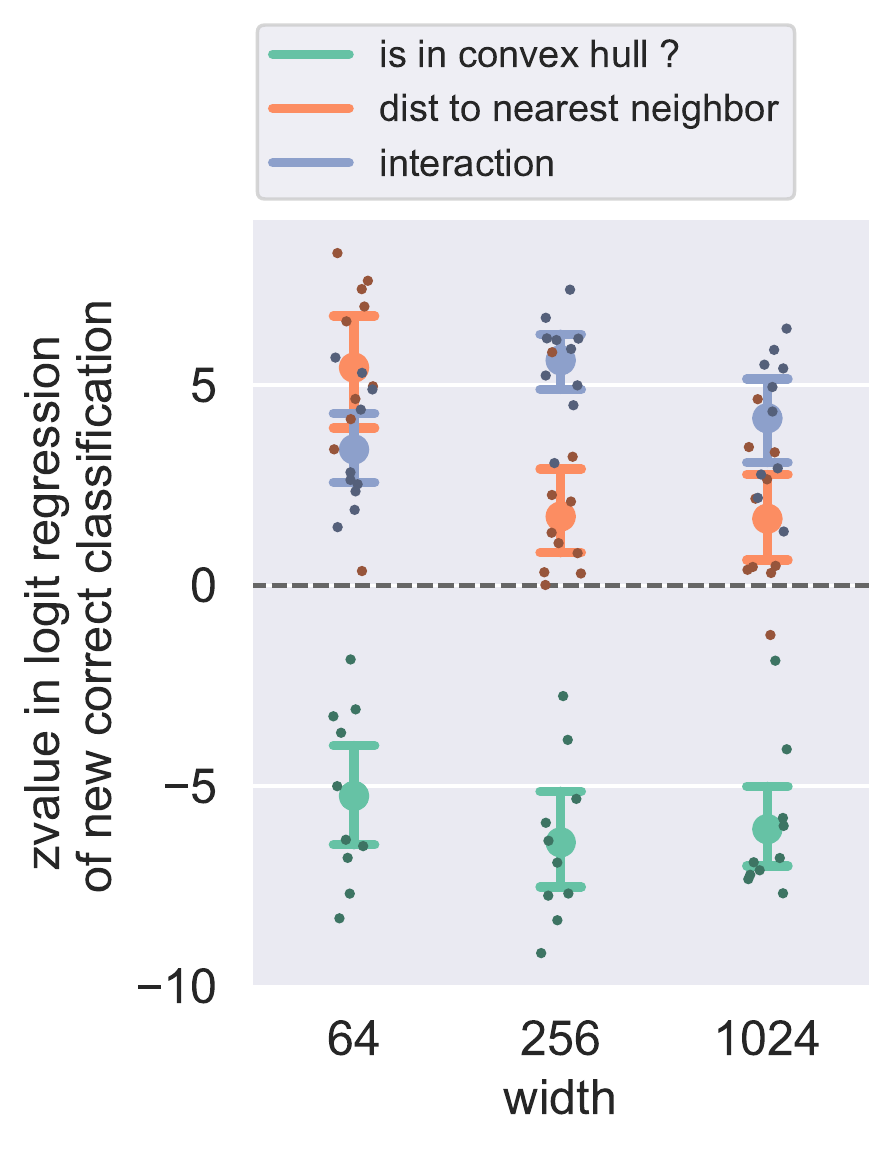} 
	\end{minipage}
	\hfill	
	\textbf{E}
	\hspace{-0.3cm}
	\includegraphics[width=0.32\linewidth,valign=T]{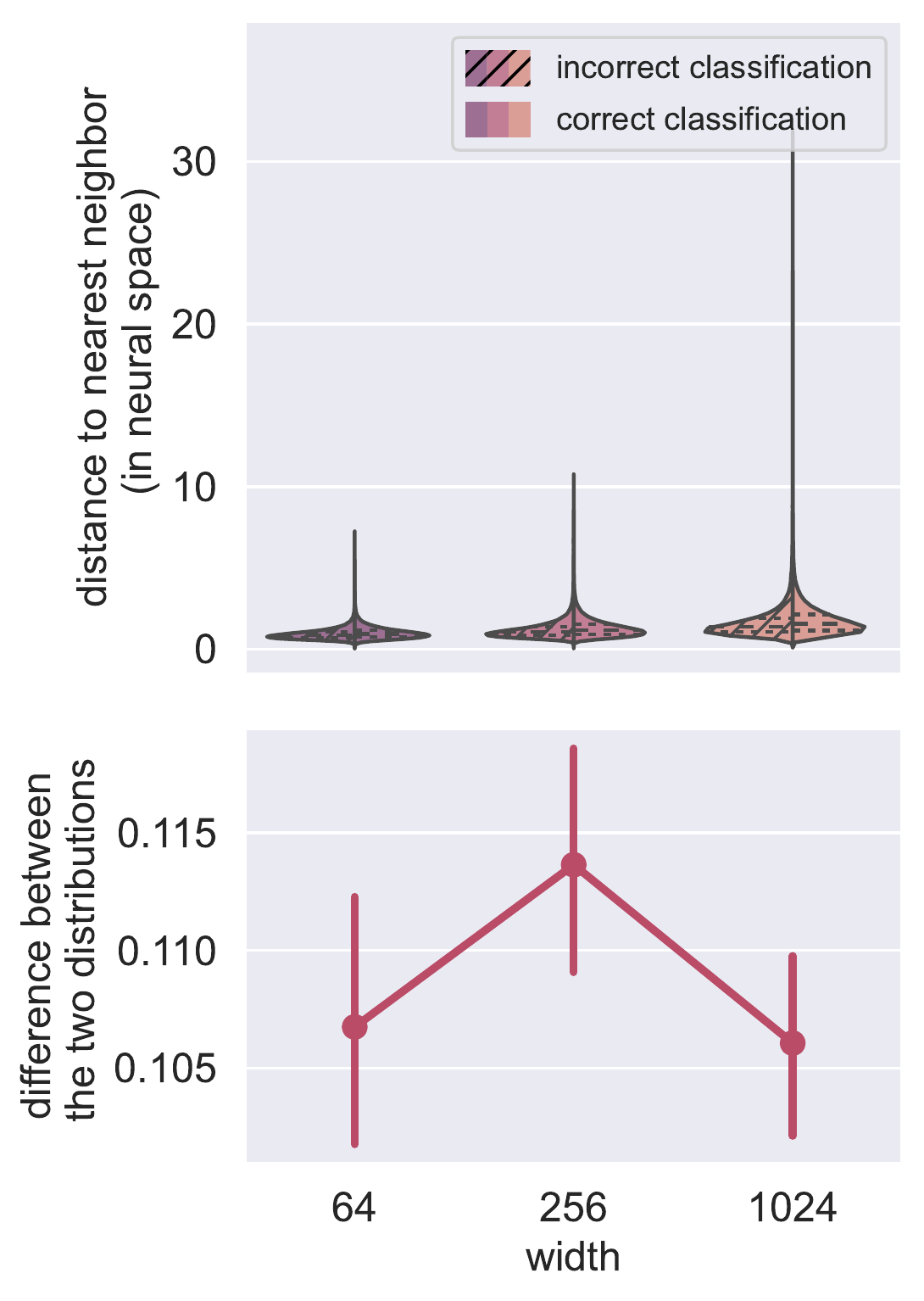} 
	
	\caption{\textbf{Distance to training set is indicative of generalization performance.} Results for the CIFAR-10 dataset and the three multilayer perceptrons considered (see Fig.~3A and Fig.~4C). The distance used here corresponds to the Euclidean distance of a new test sample to the nearest neighbor in the training set. Otherwise, same legend as in Figure~5.}
\end{figure}

\begin{figure}
	\centering
	\begin{minipage}[t]{0.28\linewidth}	
		\textbf{A}
		\hspace{-0.3cm}
		\includegraphics[width=.95\linewidth,valign=T]{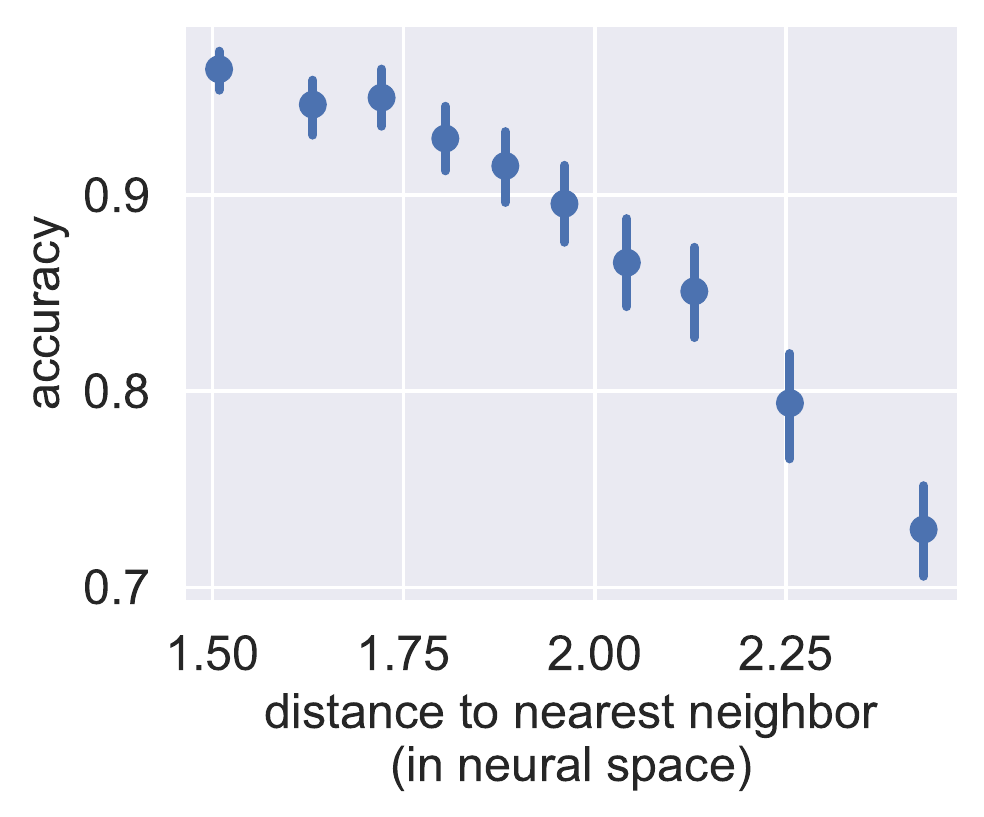}\\[20pt]
		\textbf{B}
		\hspace{-0.3cm}
		\includegraphics[width=.95\linewidth,valign=T]{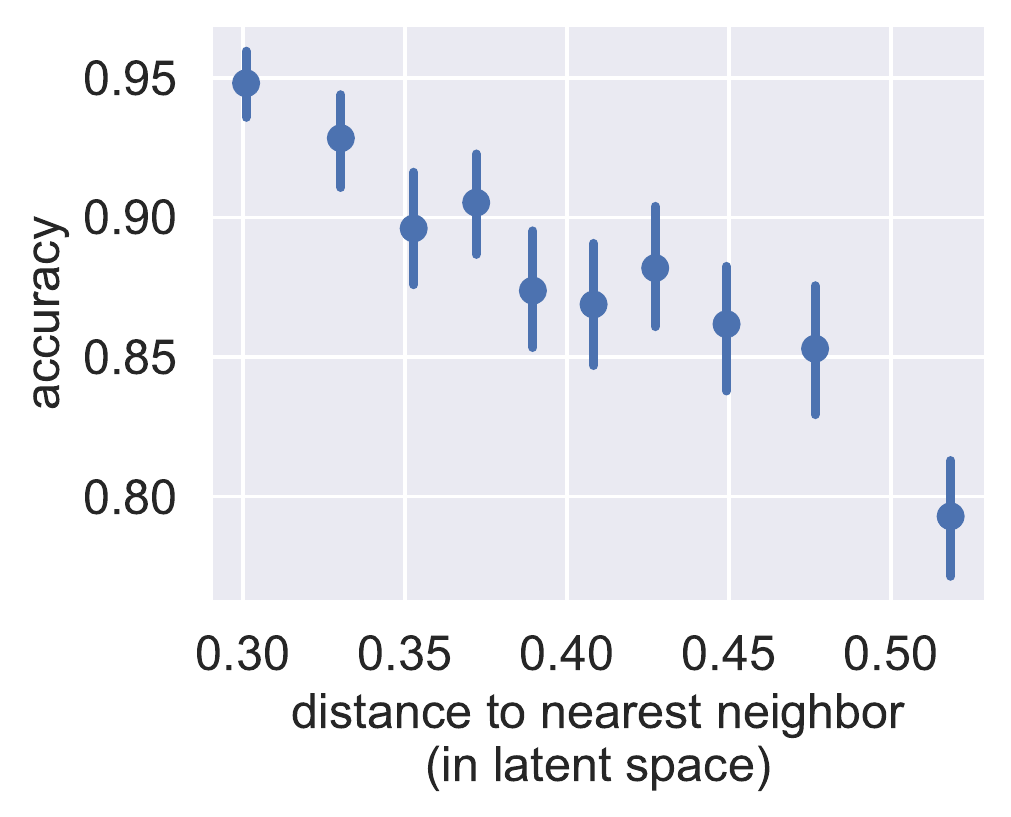} 
	\end{minipage}
	\hfill
	\begin{minipage}[t]{0.28\linewidth}	
		\textbf{C}
		\hspace{-0.5cm}\\	
		\includegraphics[width=0.9\linewidth,valign=T]{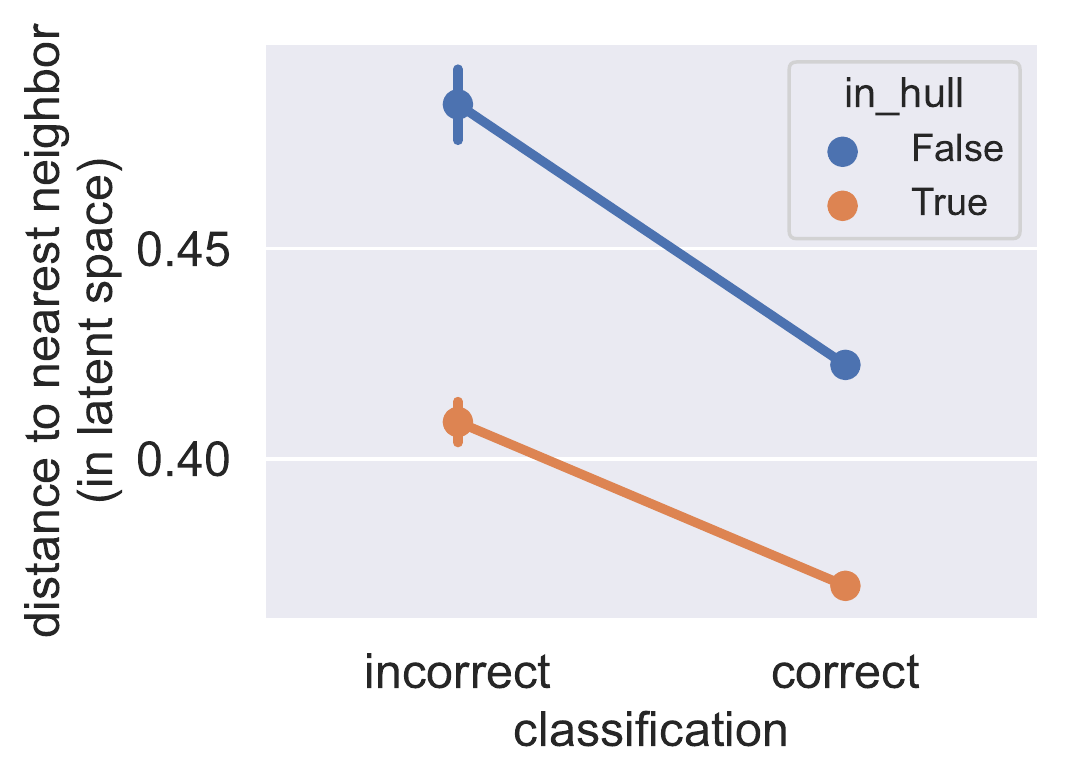}\\
		\textbf{D} 
		\hspace{-0.5cm}
		\includegraphics[width=0.9\linewidth,valign=T]{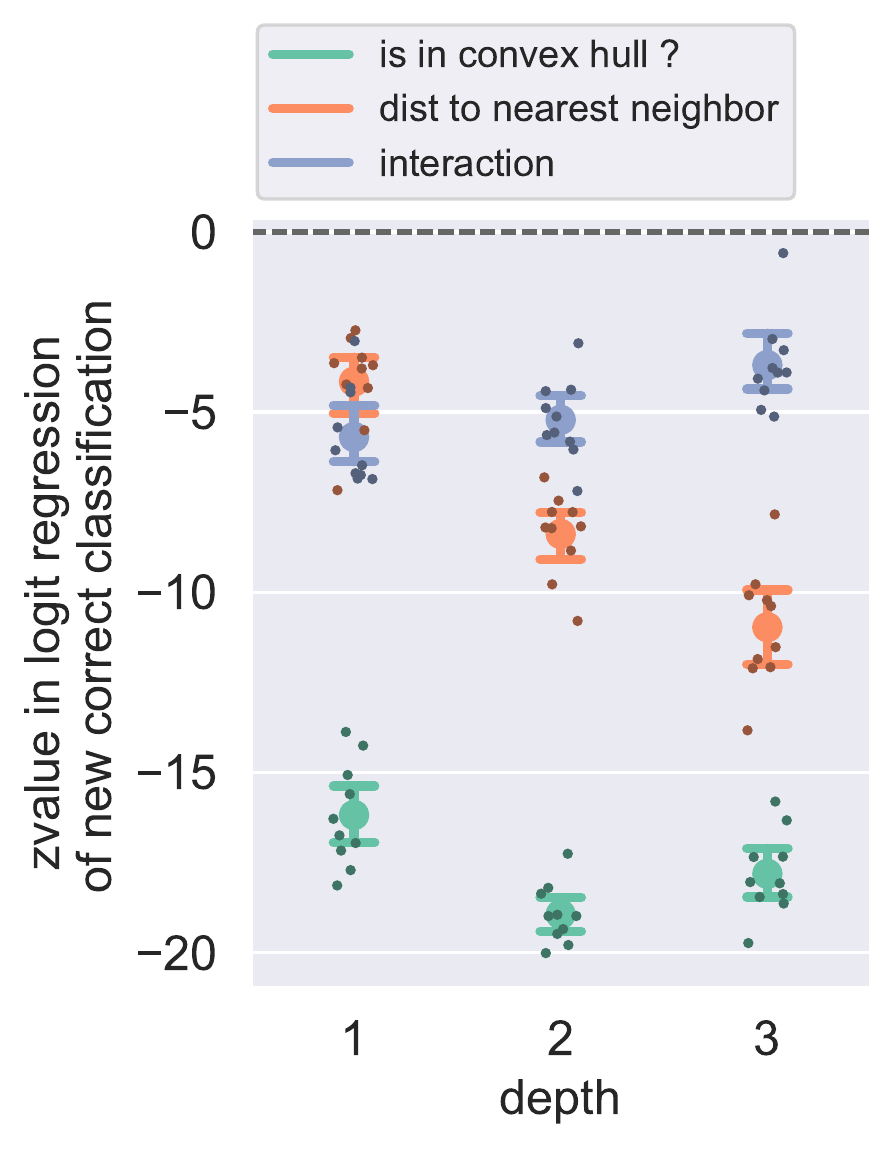} 
	\end{minipage}
	\hfill	
	\textbf{E}
	\hspace{-0.3cm}
	\includegraphics[width=0.32\linewidth,valign=T]{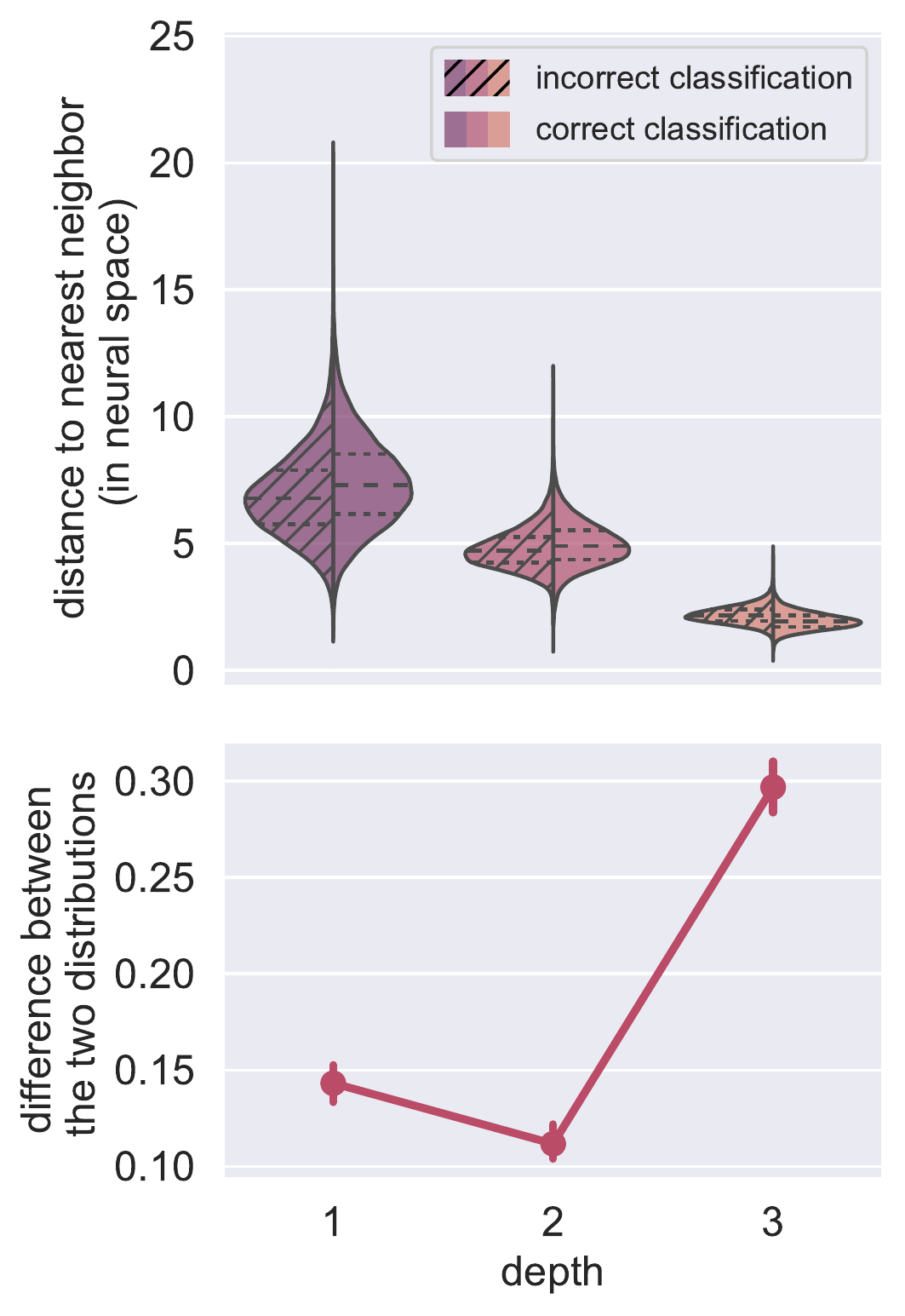} 
	
	\caption{\textbf{Distance to training set is indicative of generalization performance.} Results for the CIFAR-10 dataset and the three convolutional neural networks considered (see Fig.~3B and Fig.~4D). The distance used here corresponds to the Euclidean distance of a new test sample to the nearest neighbor in the training set. Otherwise, same legend as in Figure~5.}
\end{figure}

\begin{figure}
	\centering
	\begin{minipage}[t]{0.28\linewidth}	
		\textbf{A}
		\hspace{-0.3cm}
		\includegraphics[width=.95\linewidth,valign=T]{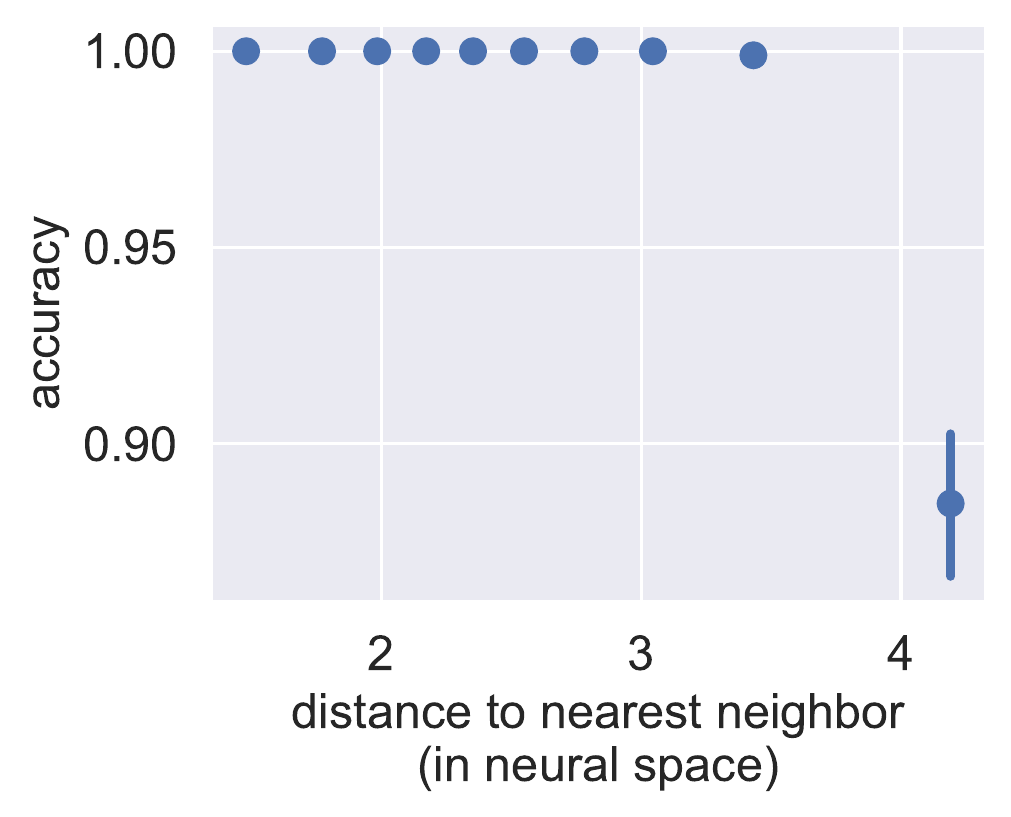}\\[20pt]
		\textbf{B}
		\hspace{-0.3cm}
		\includegraphics[width=.95\linewidth,valign=T]{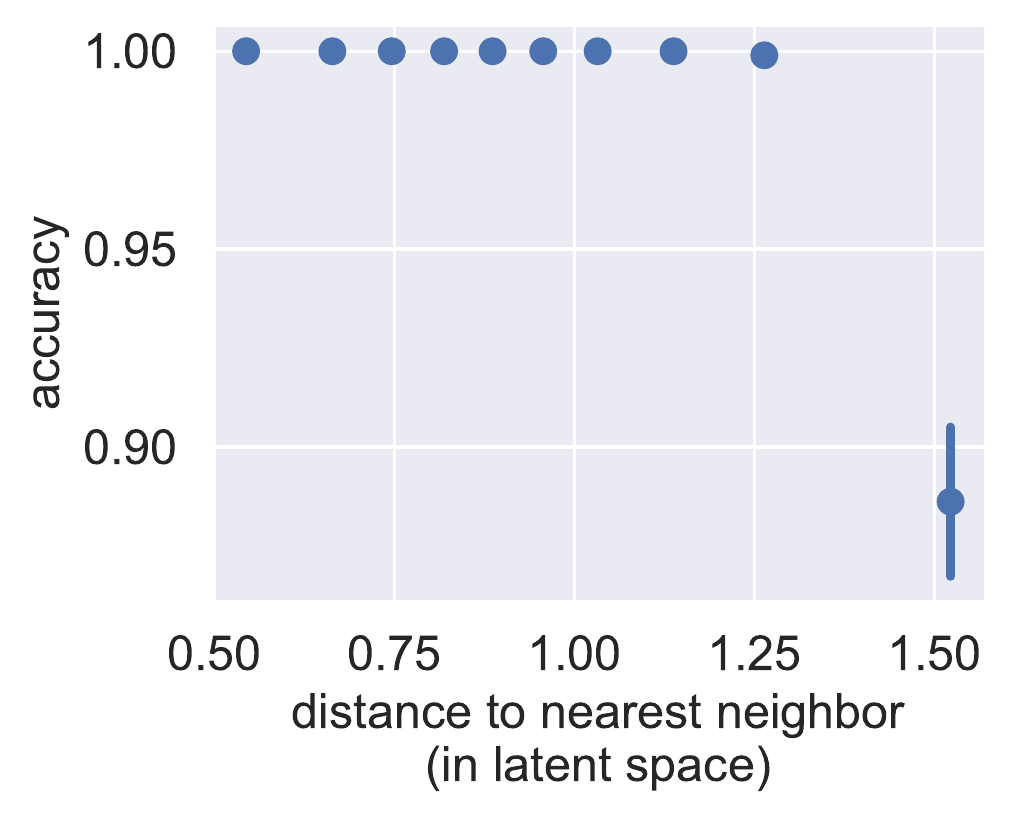} 
	\end{minipage}
	\hfill
	\begin{minipage}[t]{0.28\linewidth}	
		\textbf{C}
		\hspace{-0.5cm}\\	
		\includegraphics[width=0.9\linewidth,valign=T]{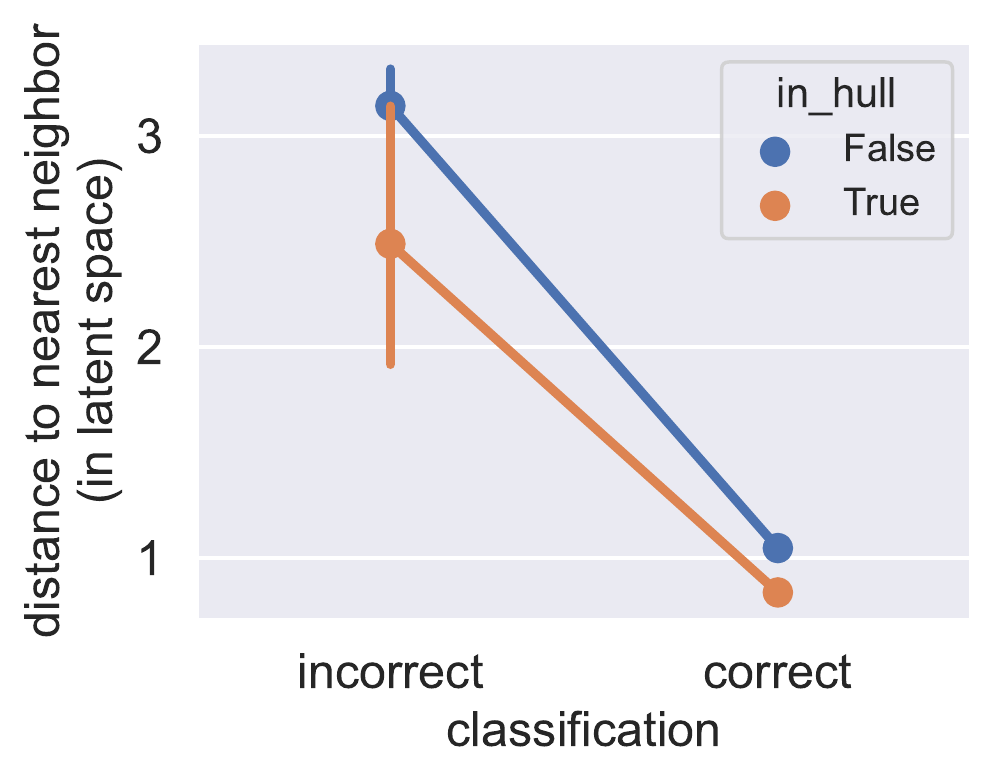}\\
		\textbf{D} 
		\hspace{-0.5cm}
		\includegraphics[width=0.9\linewidth,valign=T]{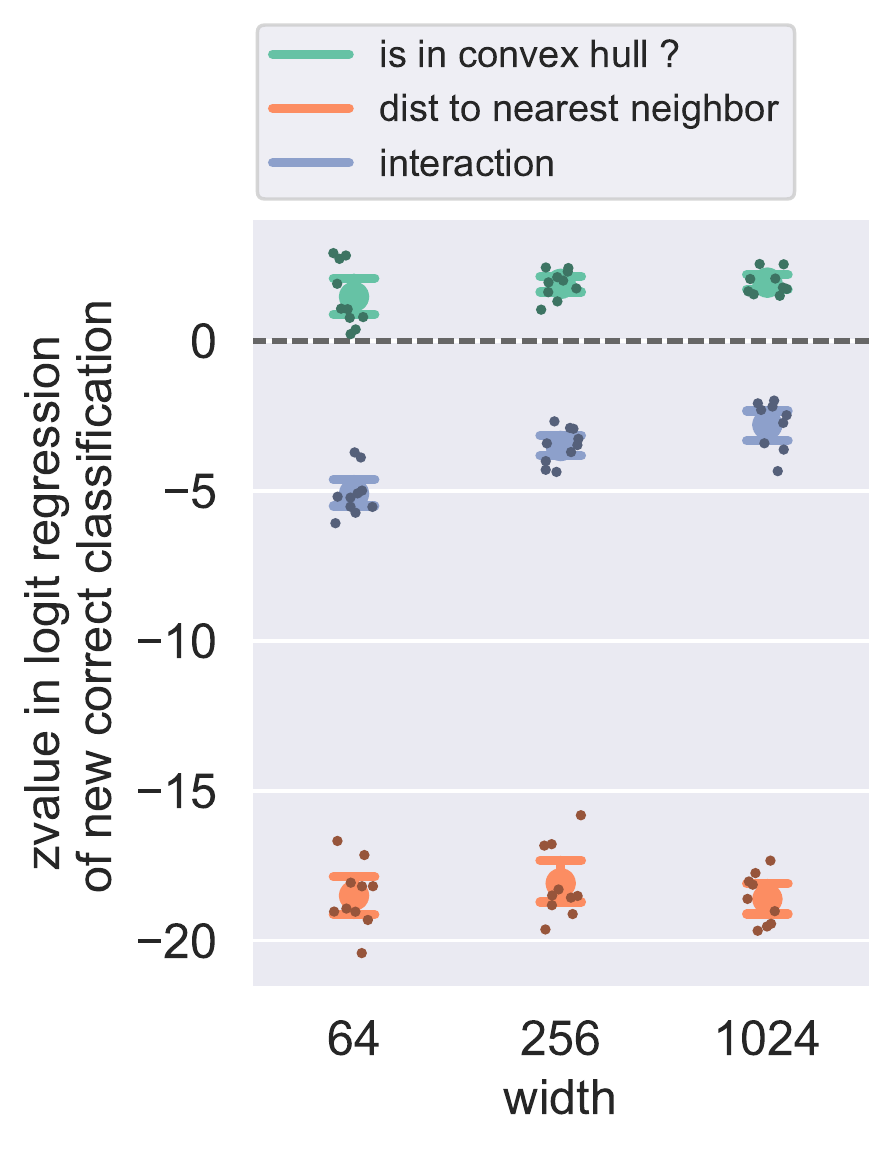} 
	\end{minipage}
	\hfill	
	\textbf{E}
	\hspace{-0.3cm}
	\includegraphics[width=0.32\linewidth,valign=T]{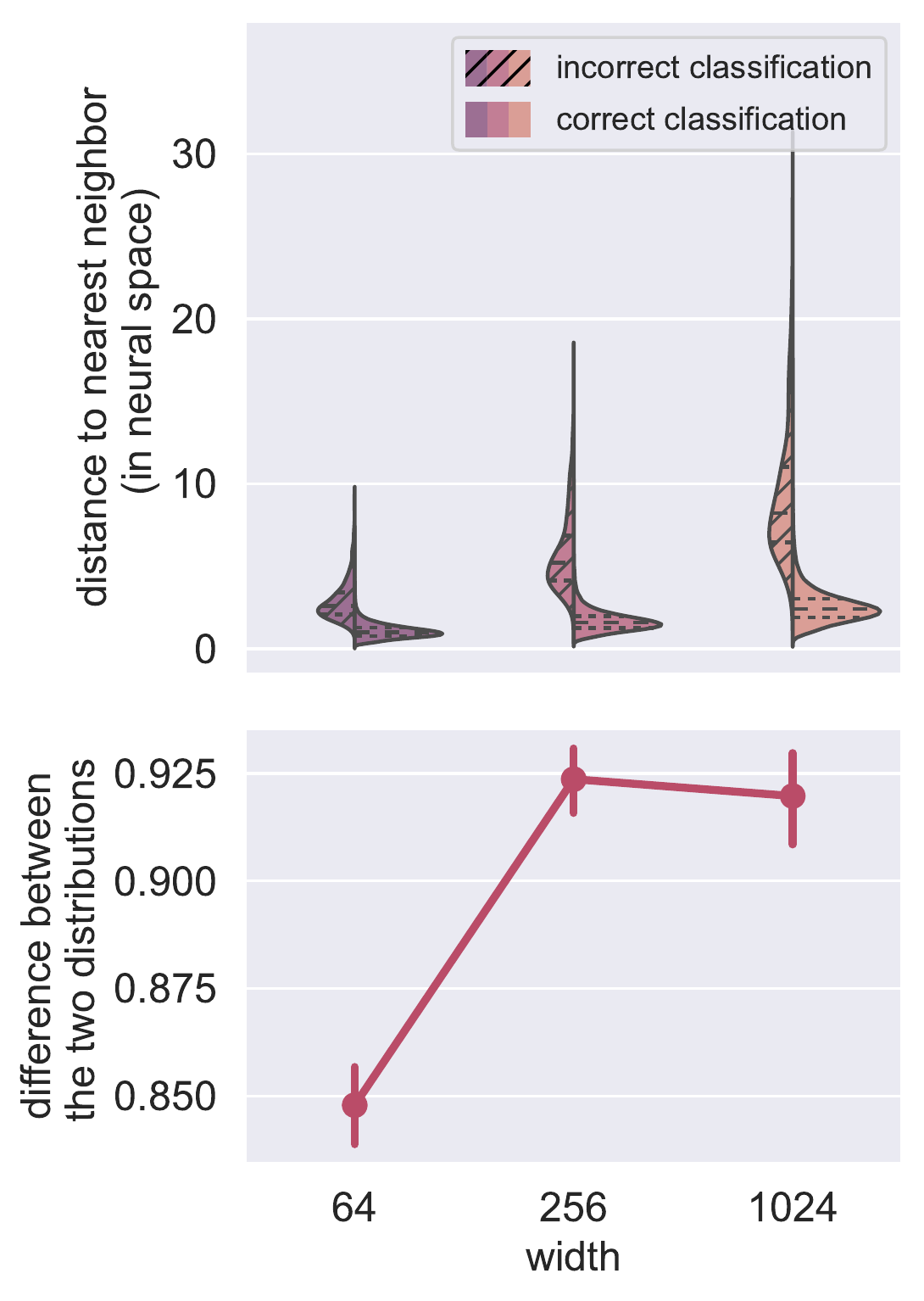} 
	
	\caption{\textbf{Distance to training set is indicative of generalization performance.} Results for the MNIST dataset and the three multilayer perceptrons considered (see Fig.~3A and Fig.~4A). The distance used here corresponds to the Euclidean distance of a new test sample to the nearest neighbor of the correct class in the training set. Otherwise, same legend as in Figure~5.}
\end{figure}

\begin{figure}
	\centering
	\begin{minipage}[t]{0.28\linewidth}	
		\textbf{A}
		\hspace{-0.3cm}
		\includegraphics[width=.95\linewidth,valign=T]{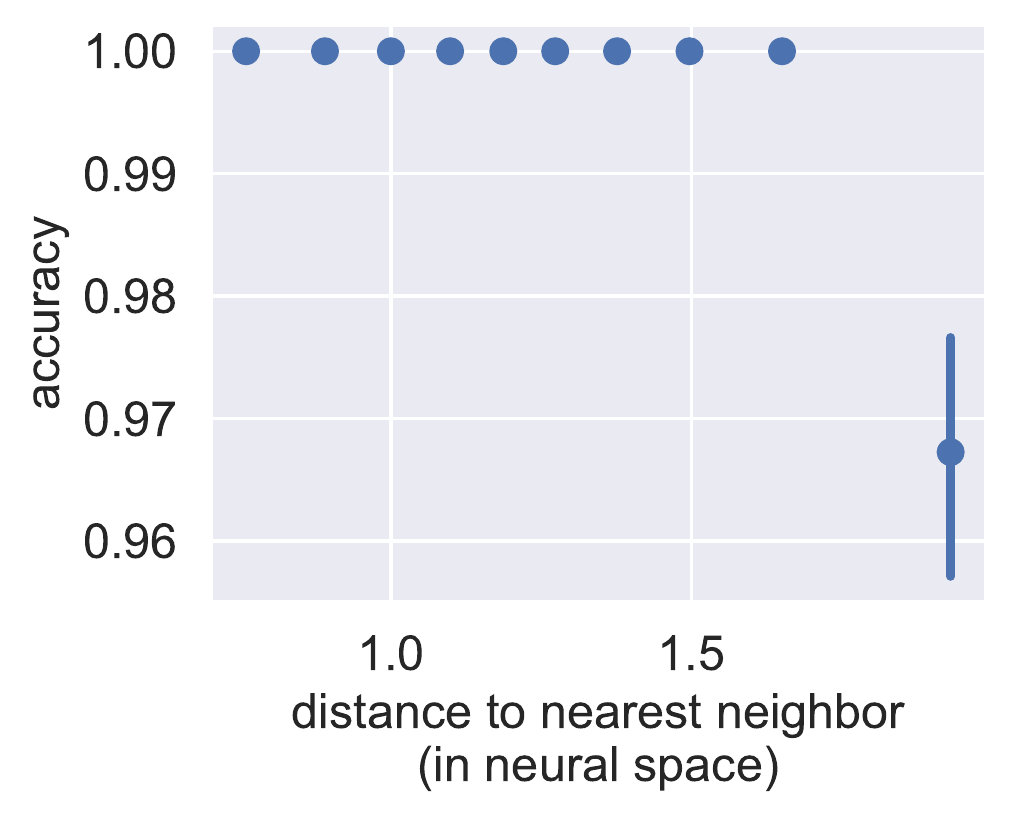}\\[20pt]
		\textbf{B}
		\hspace{-0.3cm}
		\includegraphics[width=.95\linewidth,valign=T]{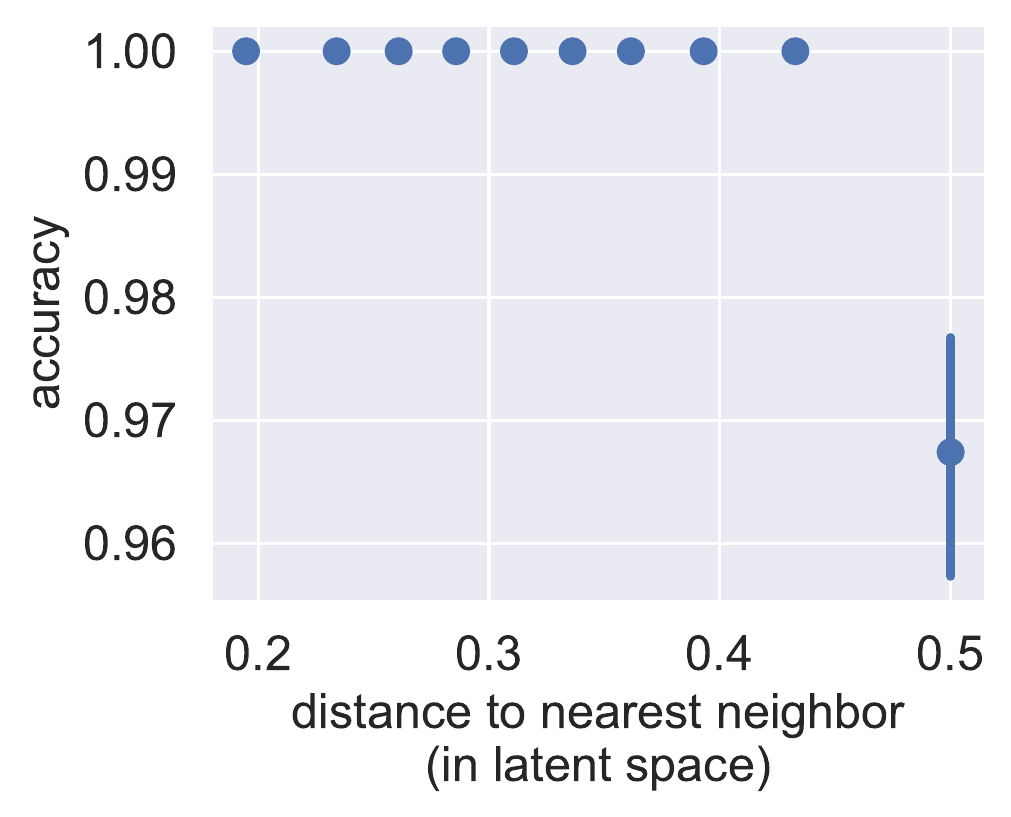} 
	\end{minipage}
	\hfill
	\begin{minipage}[t]{0.28\linewidth}	
		\textbf{C}
		\hspace{-0.5cm}\\	
		\includegraphics[width=0.9\linewidth,valign=T]{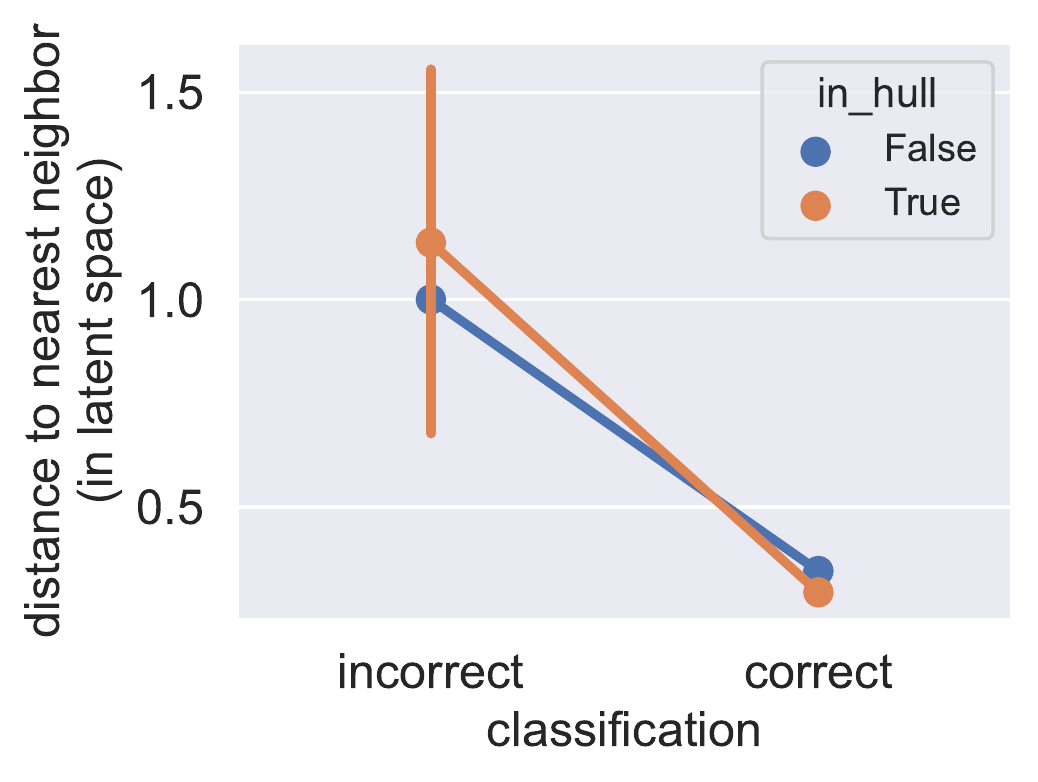}\\
		\textbf{D} 
		\hspace{-0.5cm}
		\includegraphics[width=0.9\linewidth,valign=T]{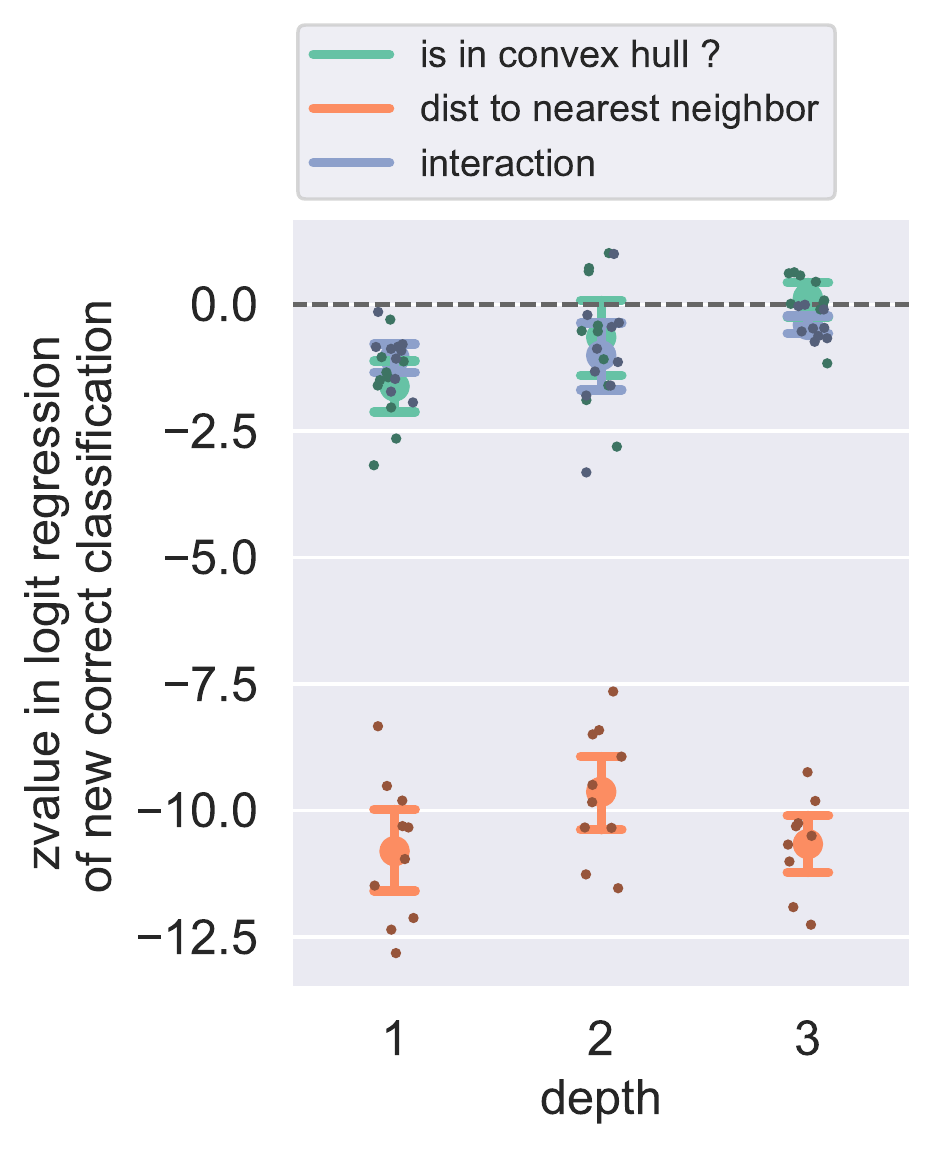} 
	\end{minipage}
	\hfill	
	\textbf{E}
	\hspace{-0.3cm}
	\includegraphics[width=0.32\linewidth,valign=T]{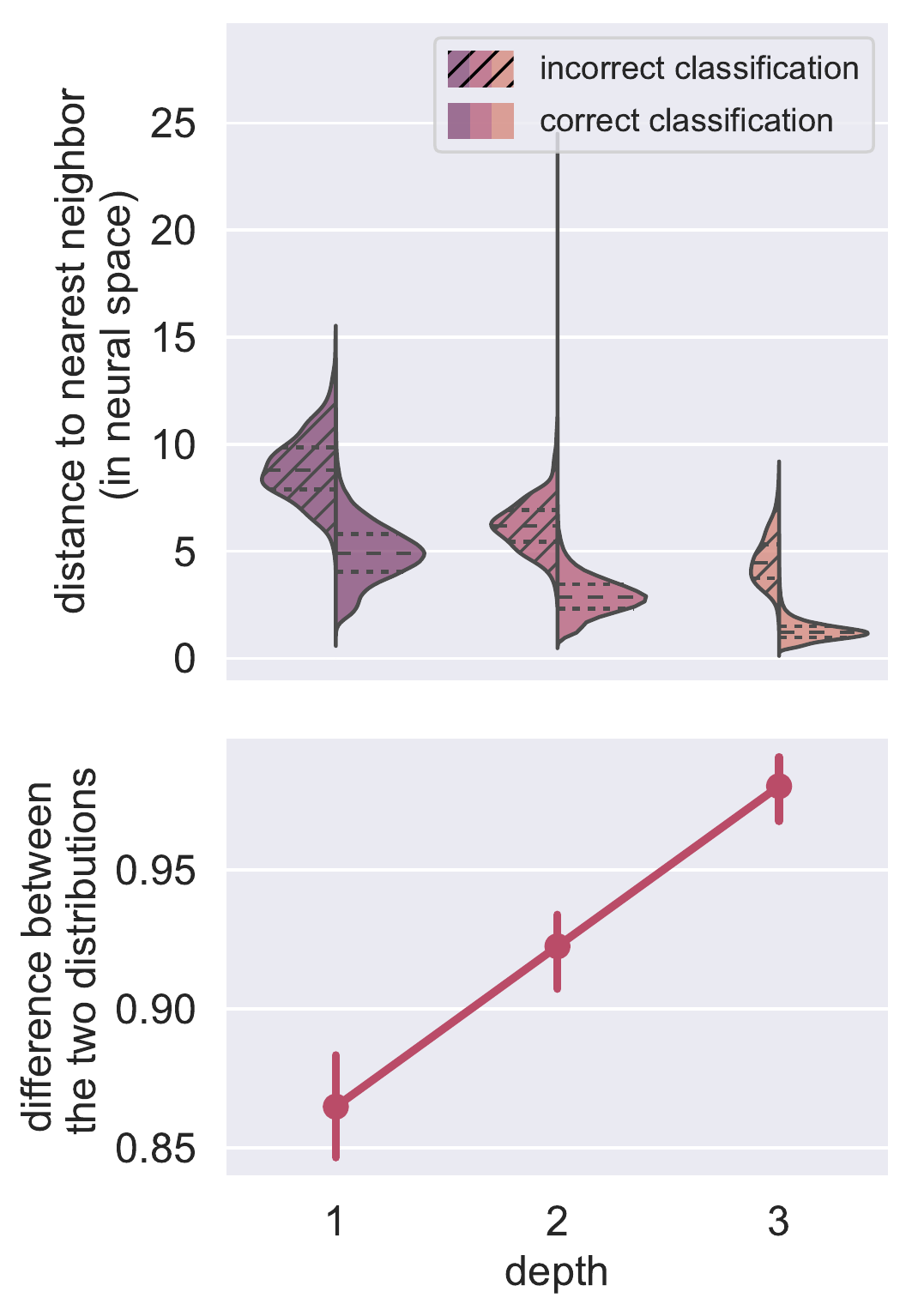} 
	
	\caption{\textbf{Distance to training set is indicative of generalization performance.} Results for the MNIST dataset and the three convolutional neural networks considered (see Fig.~3B and Fig.~4B). The distance used here corresponds to the Euclidean distance of a new test sample to the nearest neighbor of the correct class in the training set. Otherwise, same legend as in Figure~5.}
\end{figure}

\begin{figure}
	\centering
	\begin{minipage}[t]{0.28\linewidth}	
		\textbf{A}
		\hspace{-0.3cm}
		\includegraphics[width=.95\linewidth,valign=T]{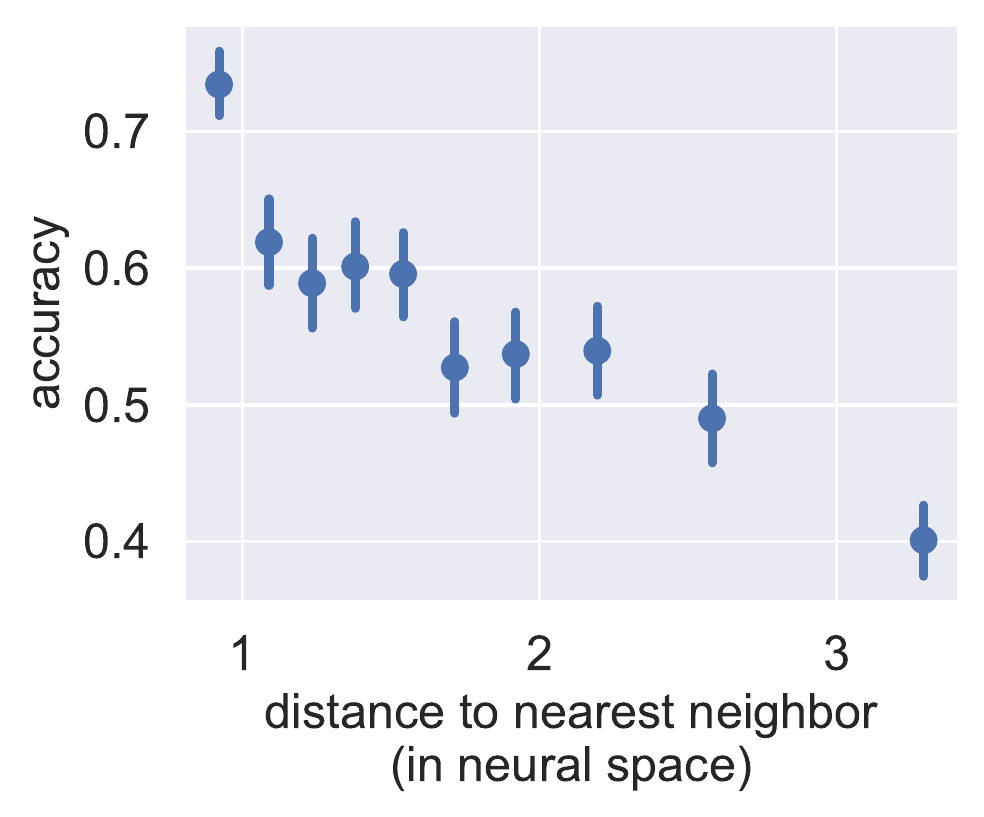}\\[20pt]
		\textbf{B}
		\hspace{-0.3cm}
		\includegraphics[width=.95\linewidth,valign=T]{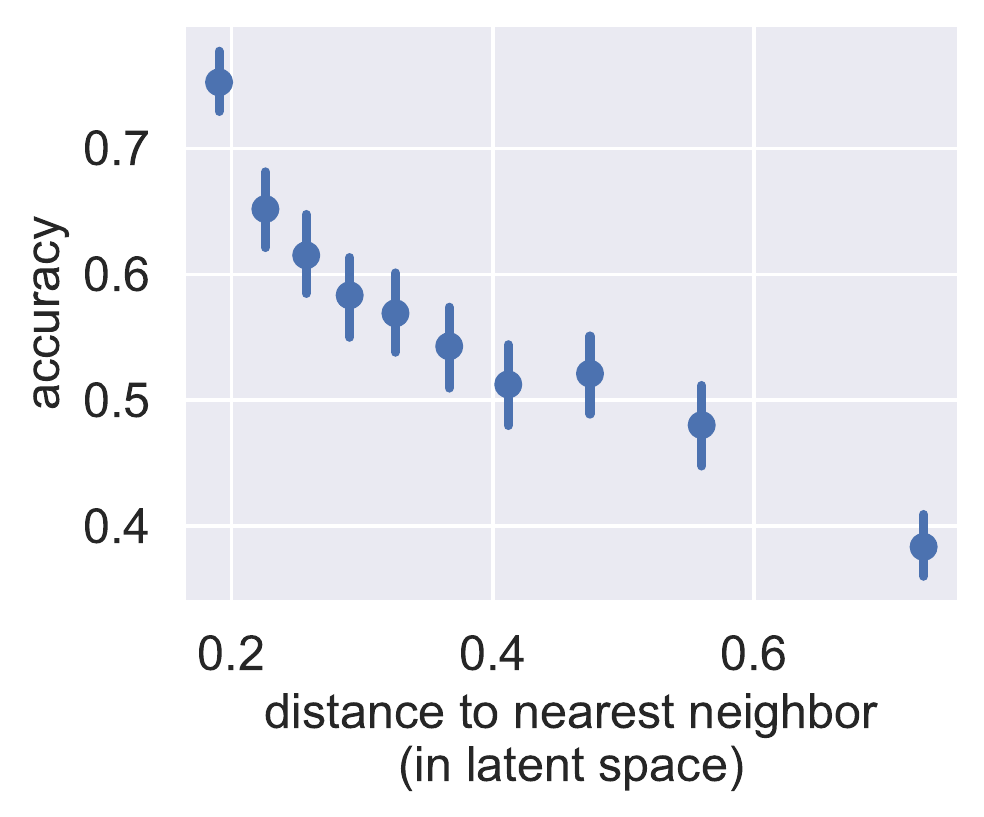} 
	\end{minipage}
	\hfill
	\begin{minipage}[t]{0.28\linewidth}	
		\textbf{C}
		\hspace{-0.5cm}\\	
		\includegraphics[width=0.9\linewidth,valign=T]{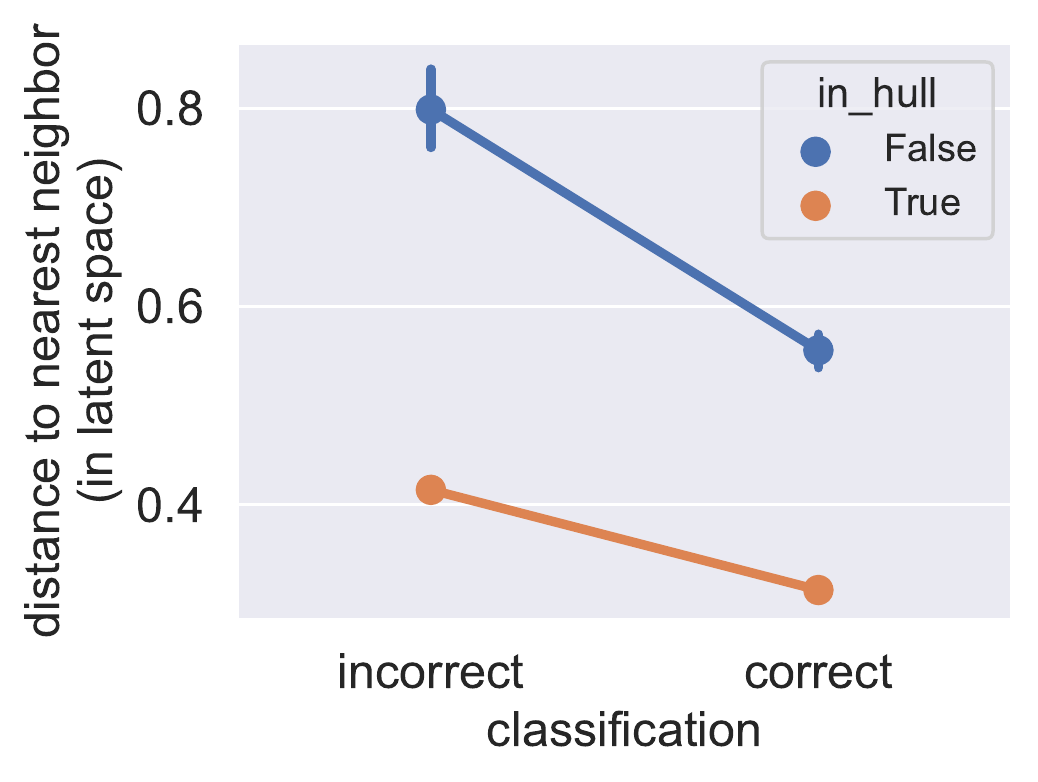}\\
		\textbf{D} 
		\hspace{-0.5cm}
		\includegraphics[width=0.9\linewidth,valign=T]{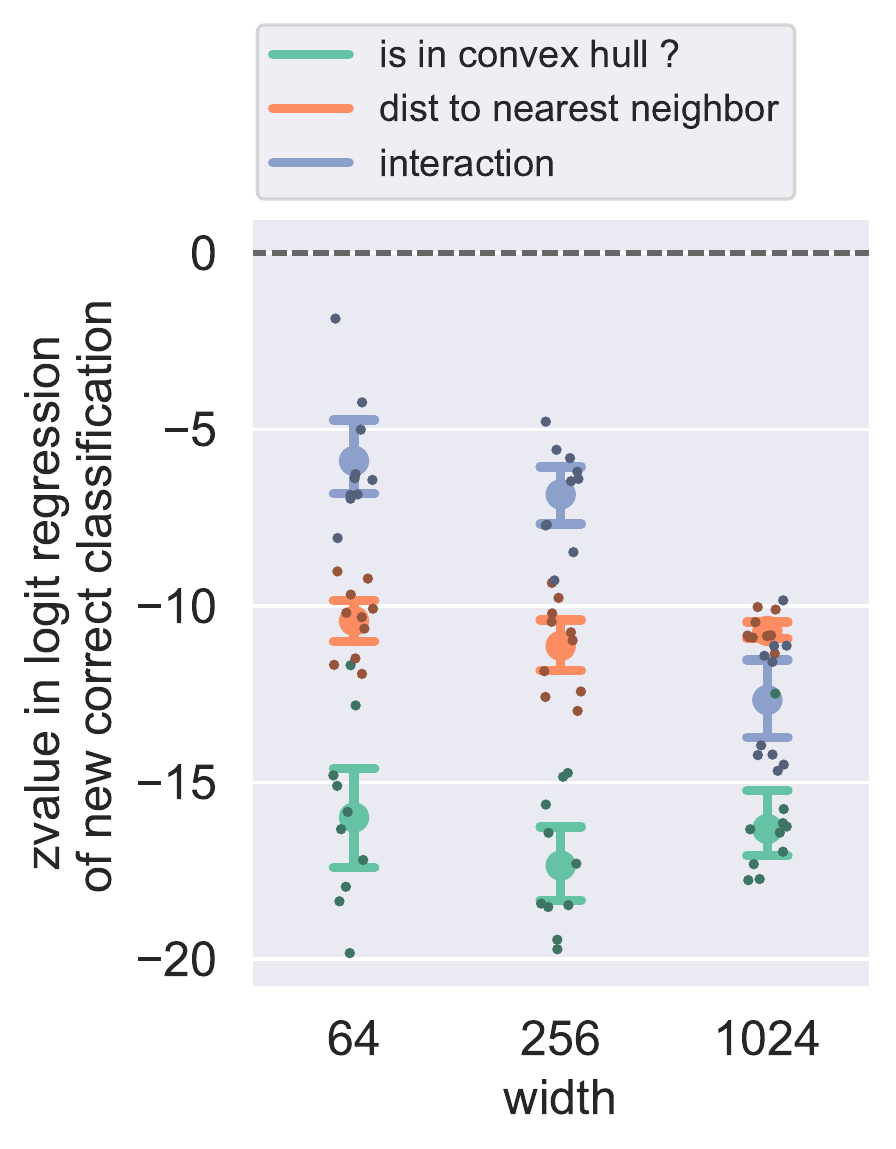} 
	\end{minipage}
	\hfill	
	\textbf{E}
	\hspace{-0.3cm}
	\includegraphics[width=0.32\linewidth,valign=T]{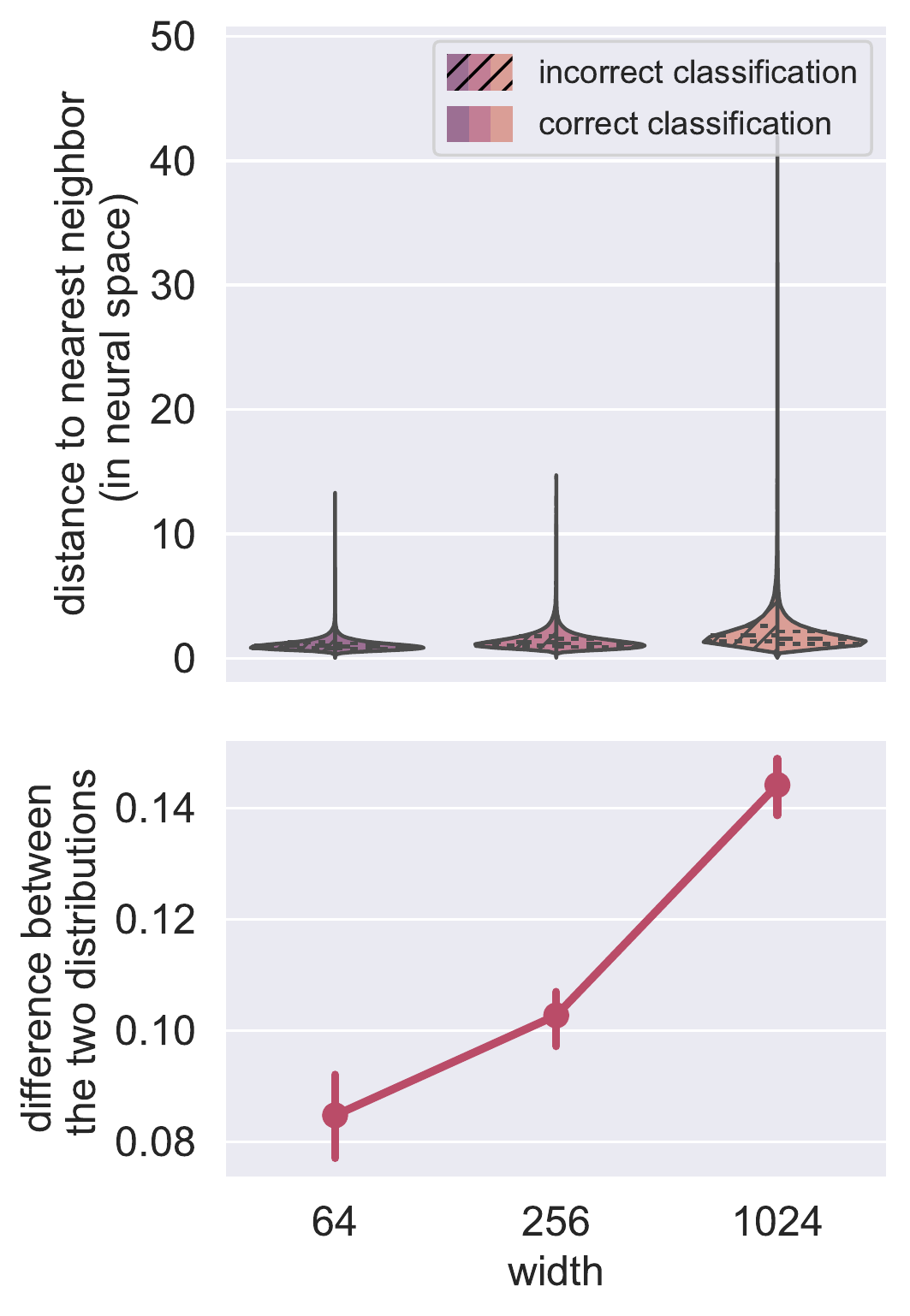} 
	
	\caption{\textbf{Distance to training set is indicative of generalization performance.} Results for the CIFAR-10 dataset and the three multilayer perceptrons considered (see Fig.~3A and Fig.~4C). The distance used here corresponds to the Euclidean distance of a new test sample to the nearest neighbor of the correct class in the training set. Otherwise, same legend as in Figure~5.}
\end{figure}

\begin{figure}
	\centering
	\begin{minipage}[t]{0.28\linewidth}	
		\textbf{A}
		\hspace{-0.3cm}
		\includegraphics[width=.95\linewidth,valign=T]{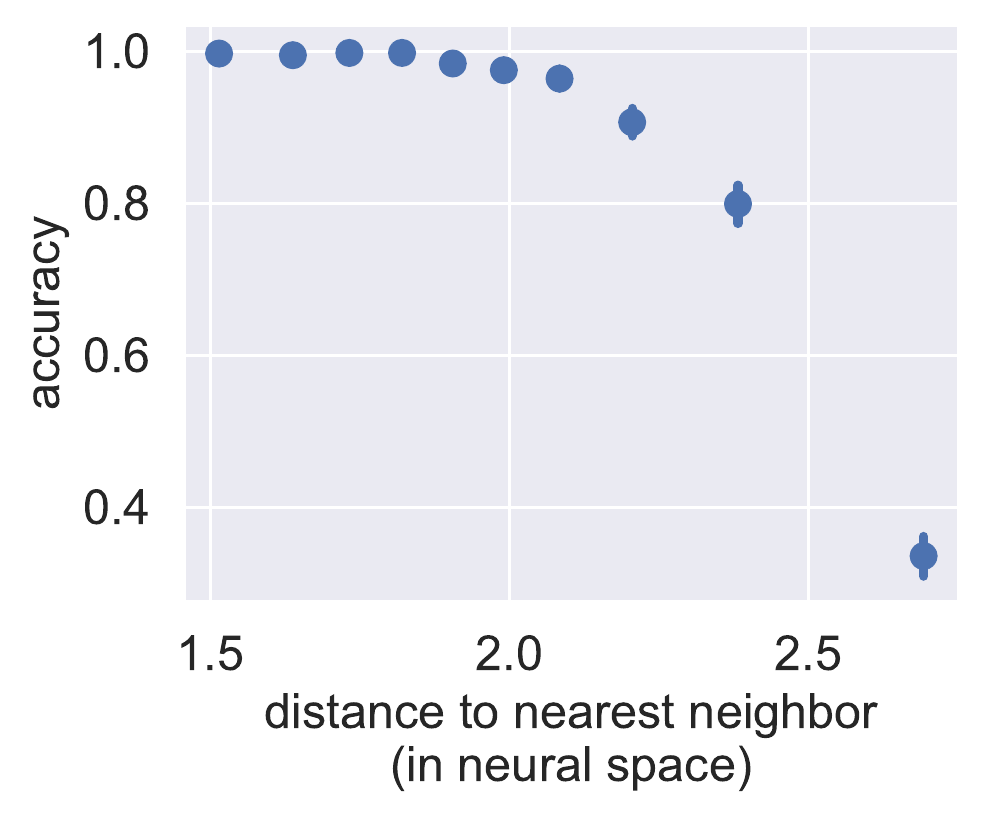}\\[20pt]
		\textbf{B}
		\hspace{-0.3cm}
		\includegraphics[width=.95\linewidth,valign=T]{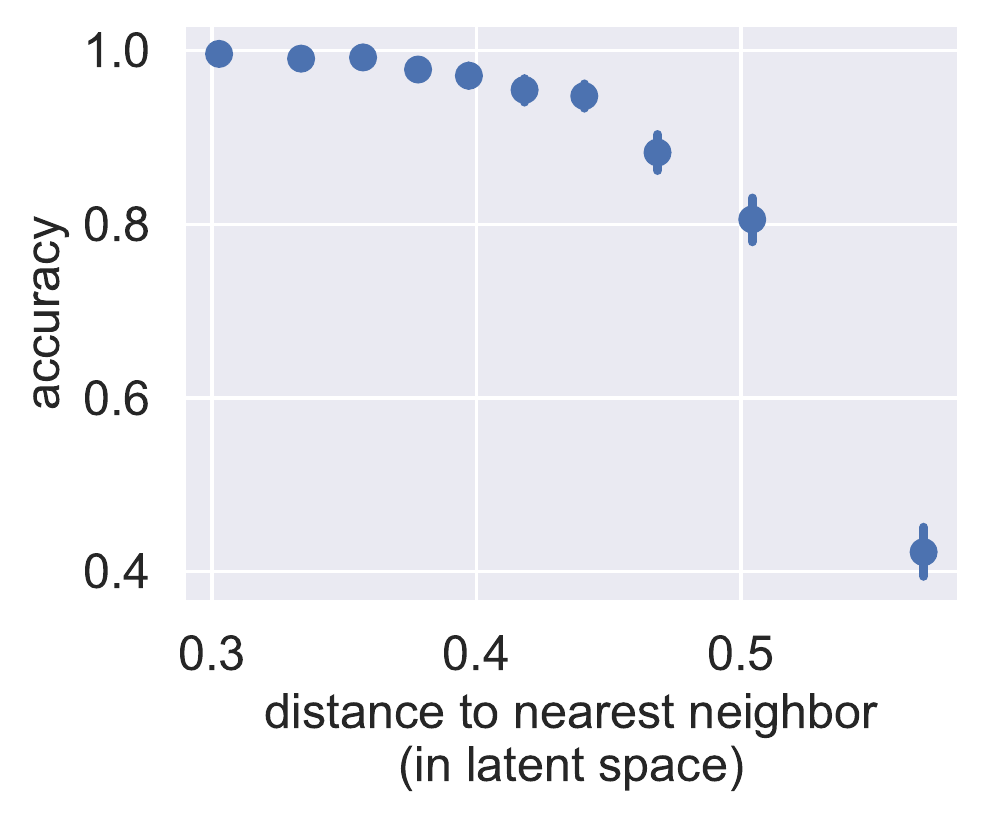} 
	\end{minipage}
	\hfill
	\begin{minipage}[t]{0.28\linewidth}	
		\textbf{C}
		\hspace{-0.5cm}\\	
		\includegraphics[width=0.9\linewidth,valign=T]{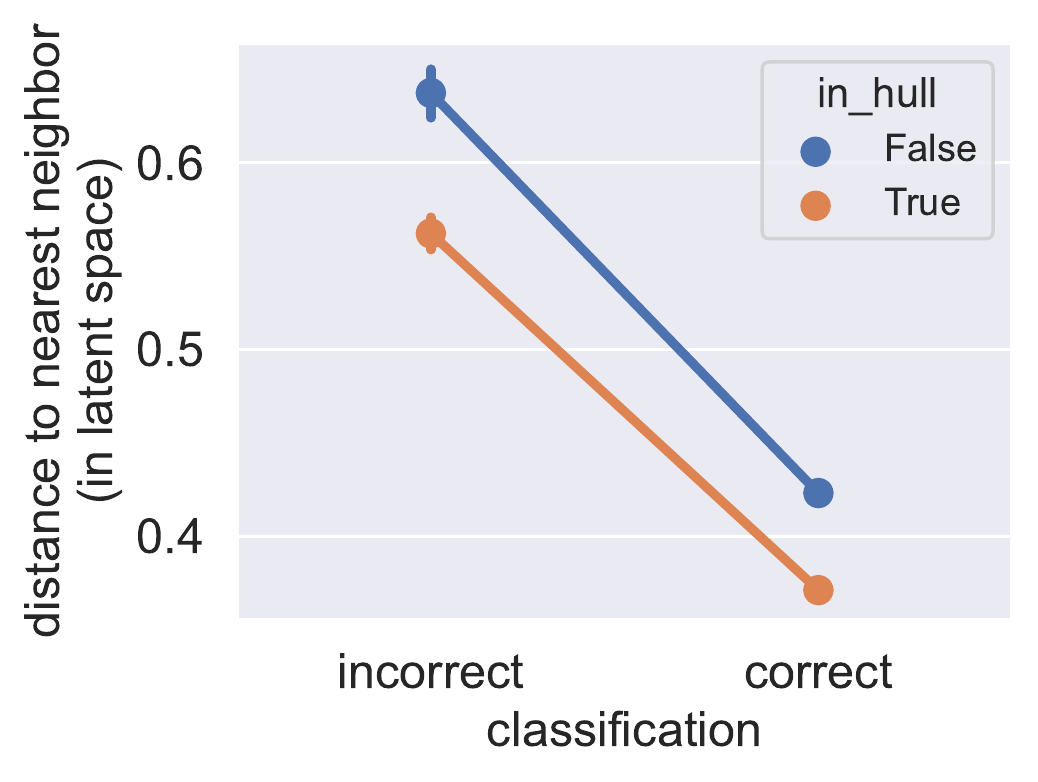}\\
		\textbf{D} 
		\hspace{-0.5cm}
		\includegraphics[width=0.9\linewidth,valign=T]{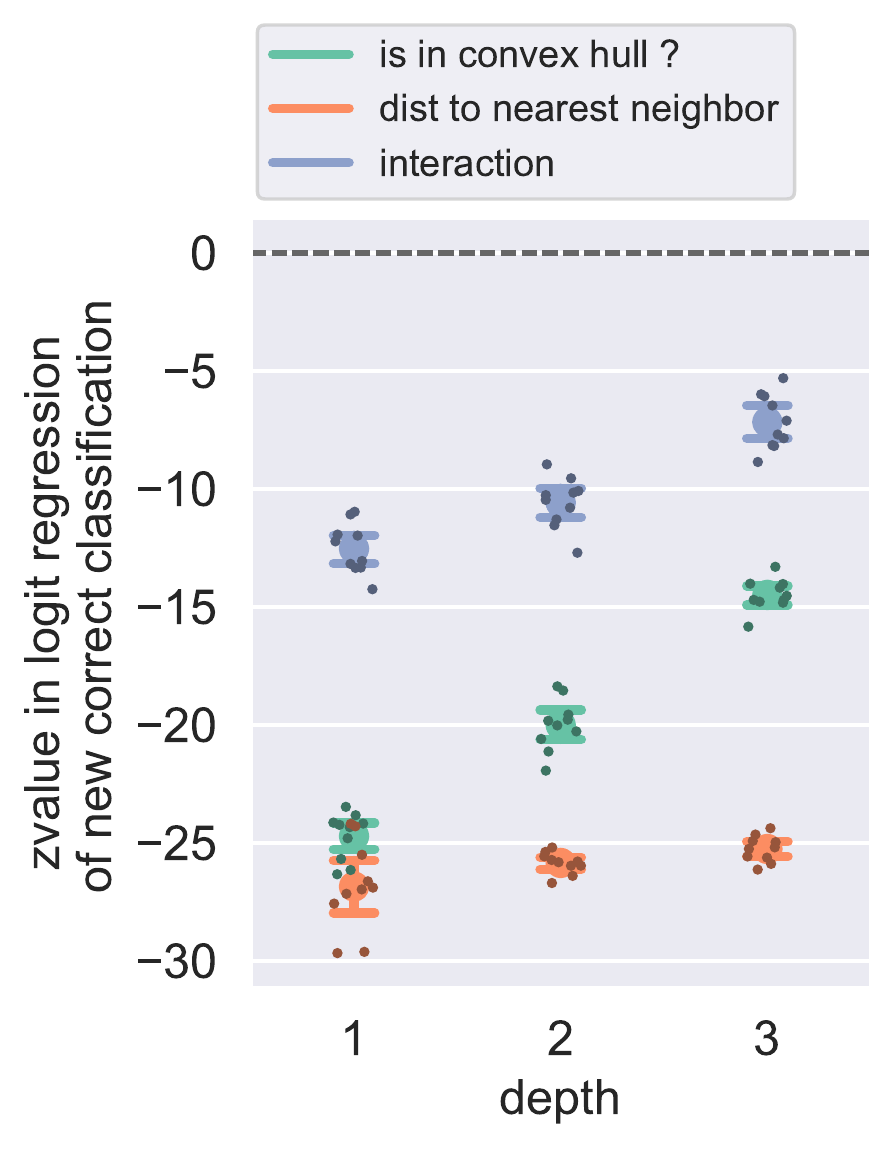} 
	\end{minipage}
	\hfill	
	\textbf{E}
	\hspace{-0.3cm}
	\includegraphics[width=0.32\linewidth,valign=T]{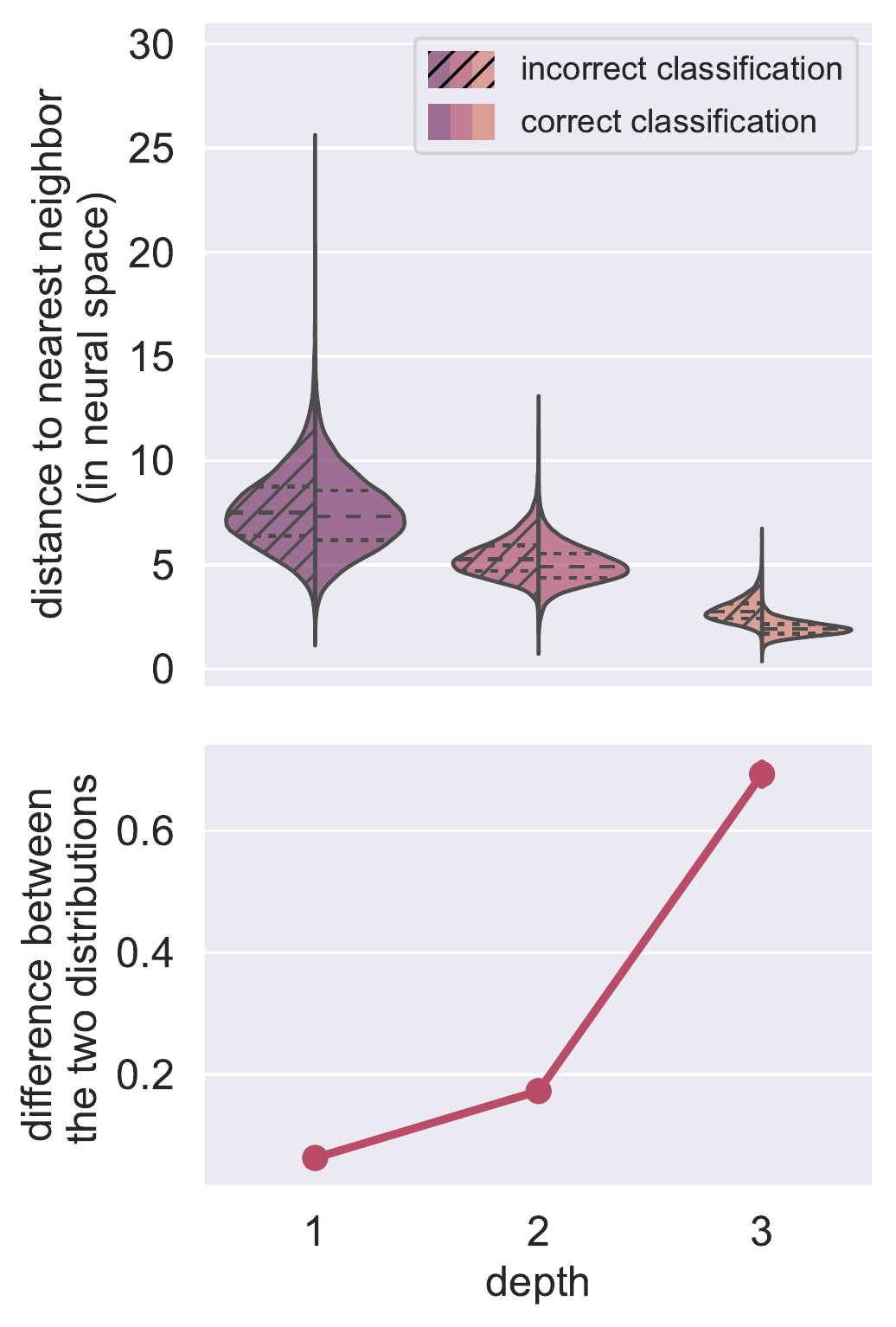} 
	
	\caption{\textbf{Distance to training set is indicative of generalization performance.} Results for the CIFAR-10 dataset and the three convolutional neural networks considered (see Fig.~3B and Fig.~4D). The distance used here corresponds to the Euclidean distance of a new test sample to the nearest neighbor of the correct class in the training set. Otherwise, same legend as in Figure~5.}
\end{figure}

\begin{figure}[h]
	\centering
	\begin{minipage}[t]{0.28\linewidth}	
		\textbf{A}
		\hspace{-0.3cm}
		\includegraphics[width=.95\linewidth,valign=T]{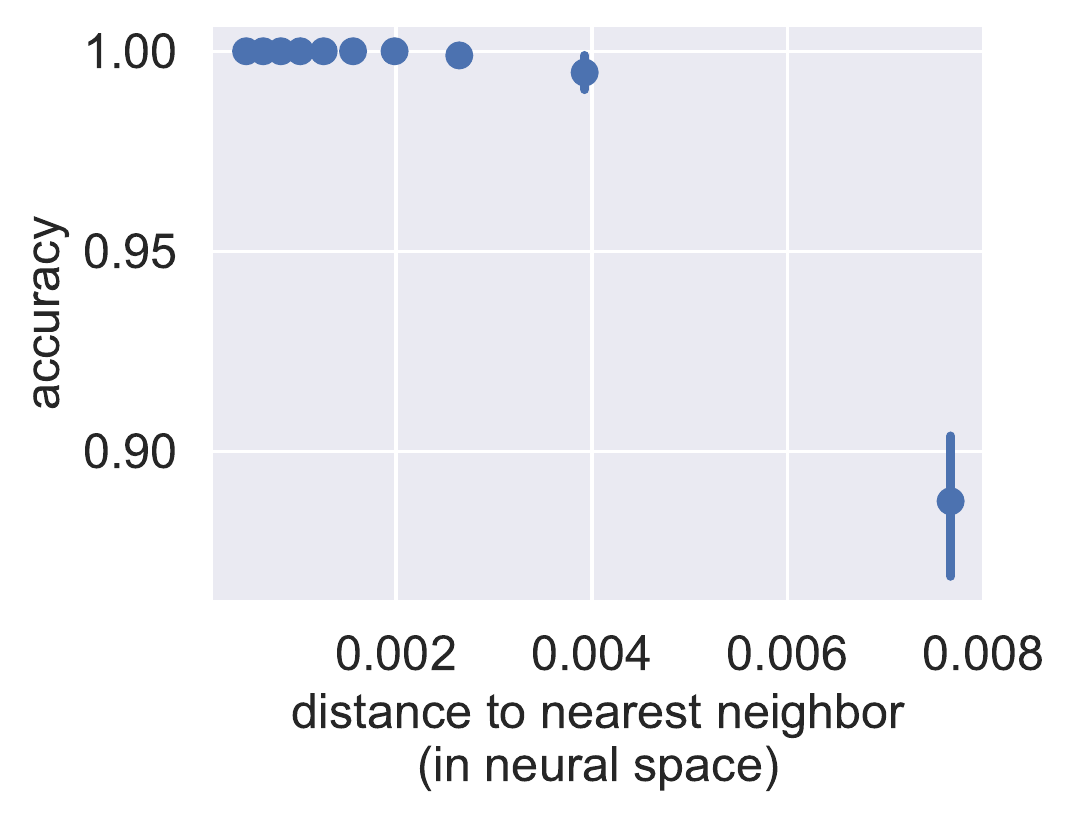}\\[20pt]
		\textbf{B}
		\hspace{-0.3cm}
		\includegraphics[width=.95\linewidth,valign=T]{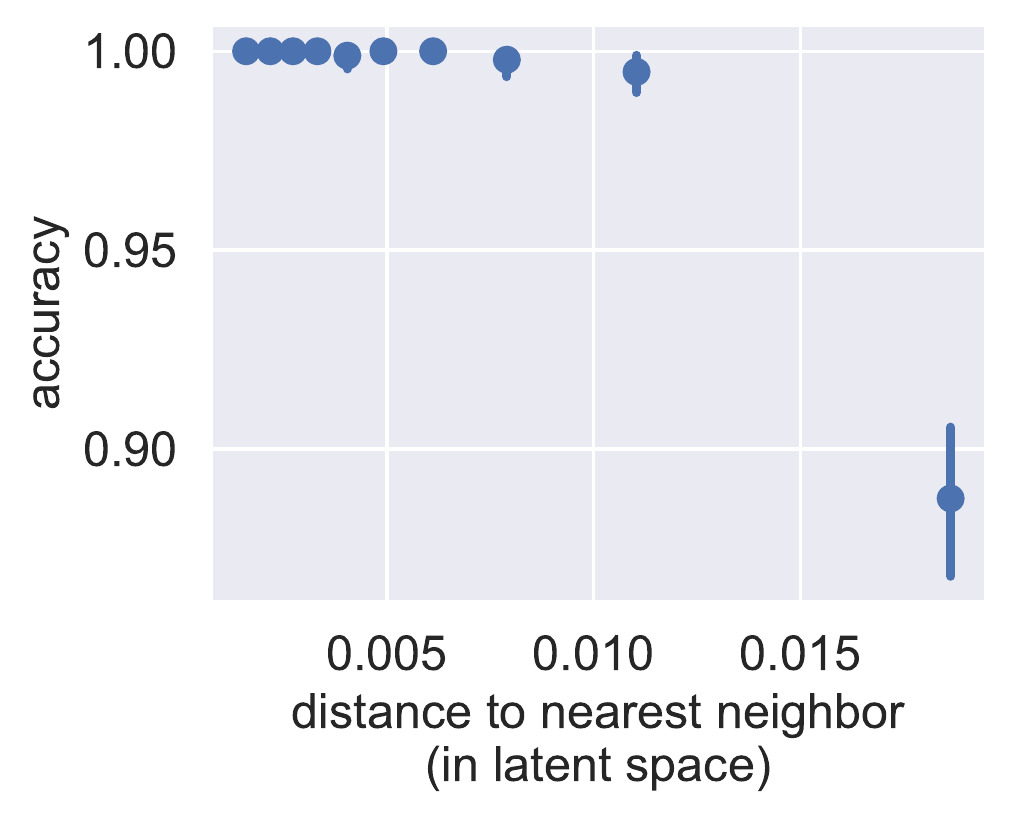} 
	\end{minipage}
	\hfill
	\begin{minipage}[t]{0.28\linewidth}	
		\textbf{C}
		\hspace{-0.5cm}\\	
		\includegraphics[width=0.9\linewidth,valign=T]{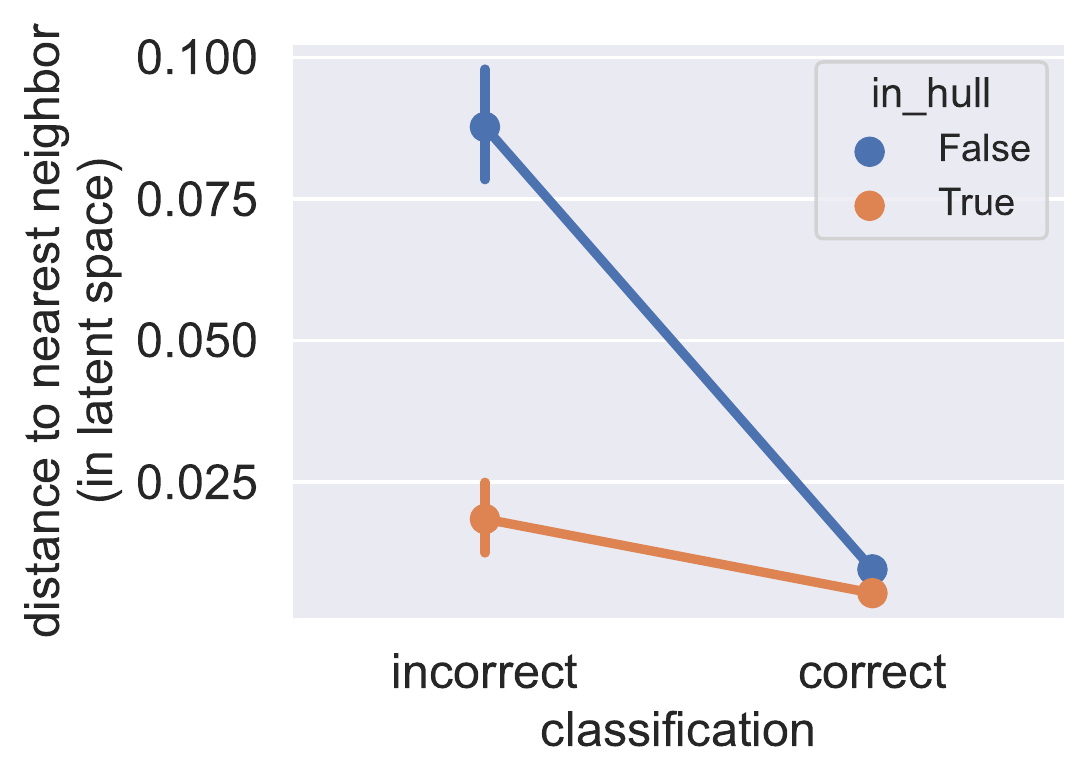}\\
		\textbf{D} 
		\hspace{-0.5cm}
		\includegraphics[width=0.9\linewidth,valign=T]{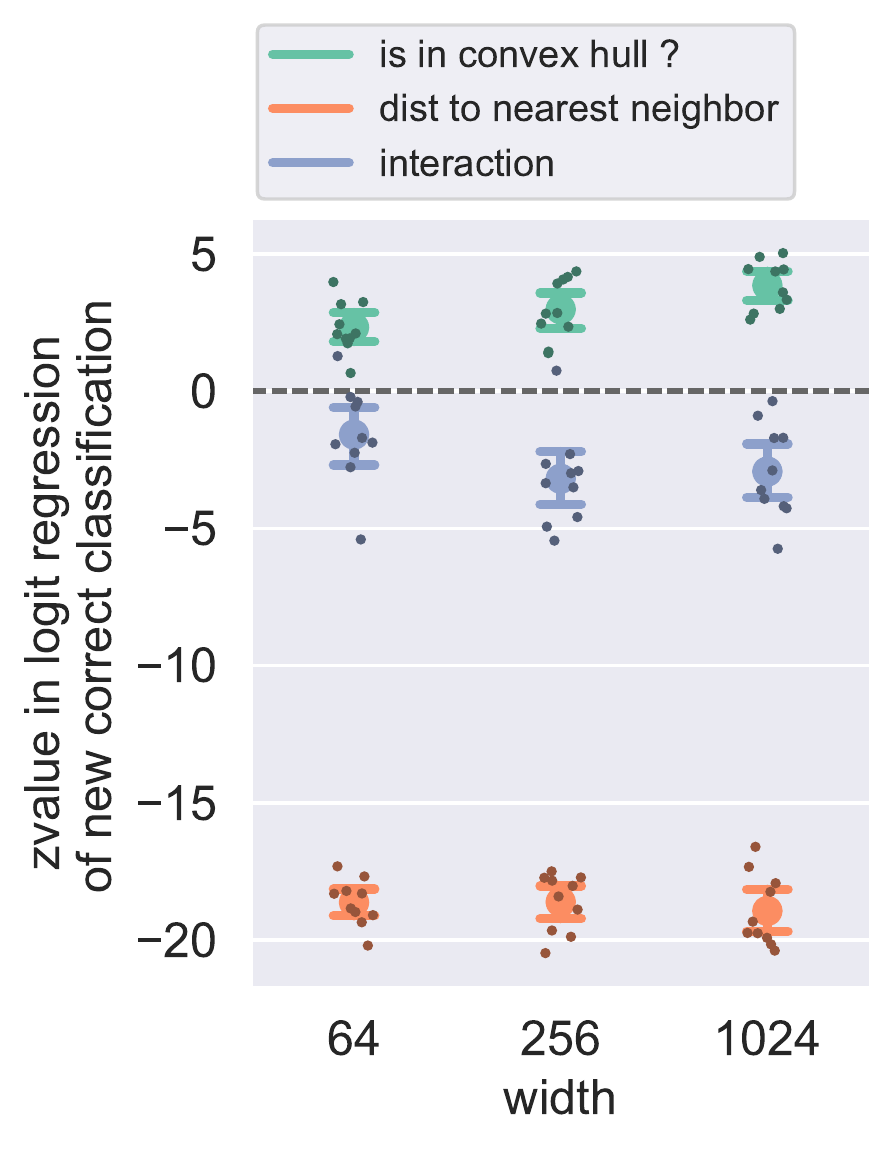} 
	\end{minipage}
	\hfill	
	\textbf{E}
	\hspace{-0.3cm}
	\includegraphics[width=0.32\linewidth,valign=T]{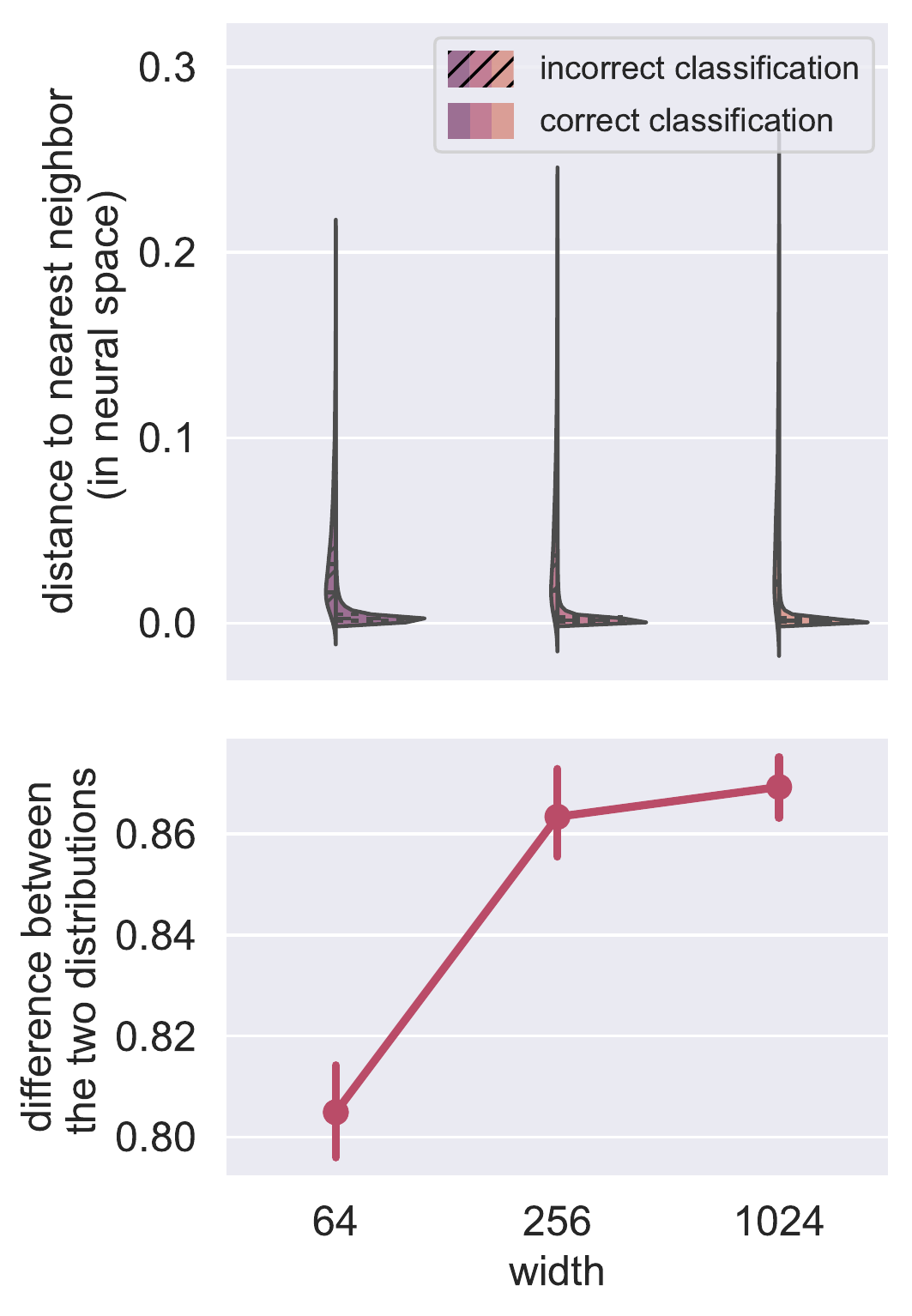} 
	
	\caption{\textbf{Distance to training set is indicative of generalization performance.} Results for the MNIST dataset and the three multilayer perceptrons considered (see Fig.~3A and Fig.~4A). The distance used here corresponds to the cosine distance of a new test sample to the nearest neighbor in the training set. Otherwise, same legend as in Figure~5.}
\end{figure}

\begin{figure}[!h]
	\centering
	\begin{minipage}[t]{0.28\linewidth}	
		\textbf{A}
		\hspace{-0.3cm}
		\includegraphics[width=.95\linewidth,valign=T]{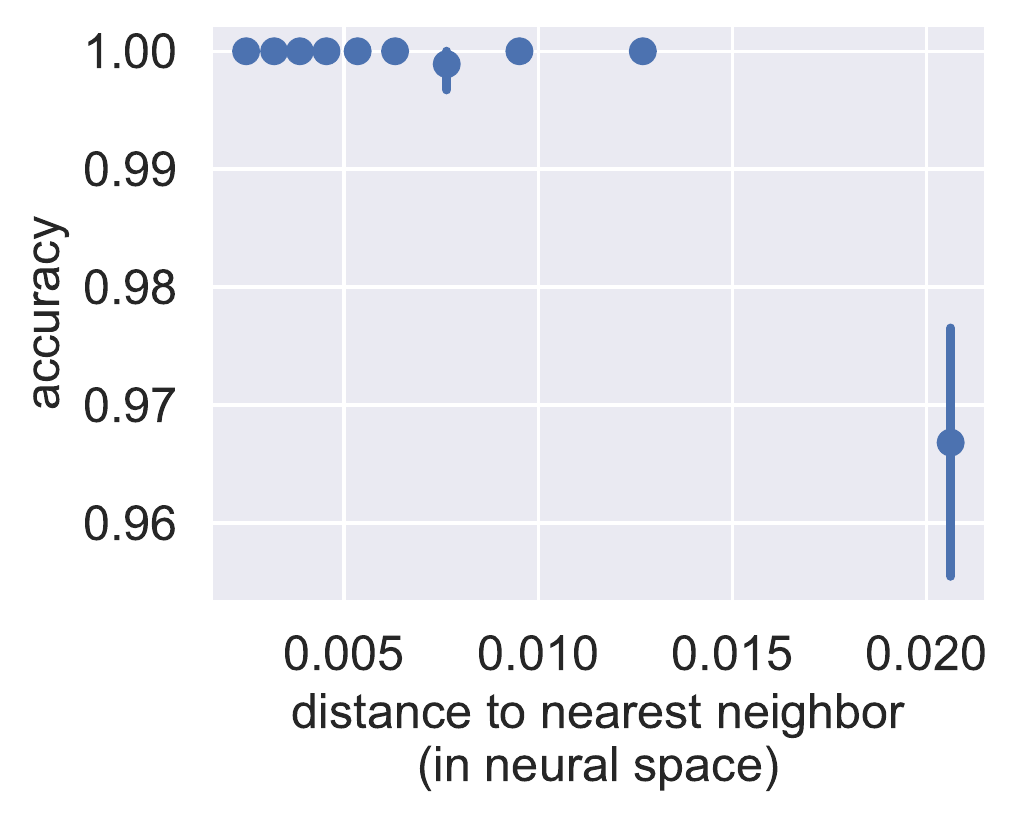}\\[20pt]
		\textbf{B}
		\hspace{-0.3cm}
		\includegraphics[width=.95\linewidth,valign=T]{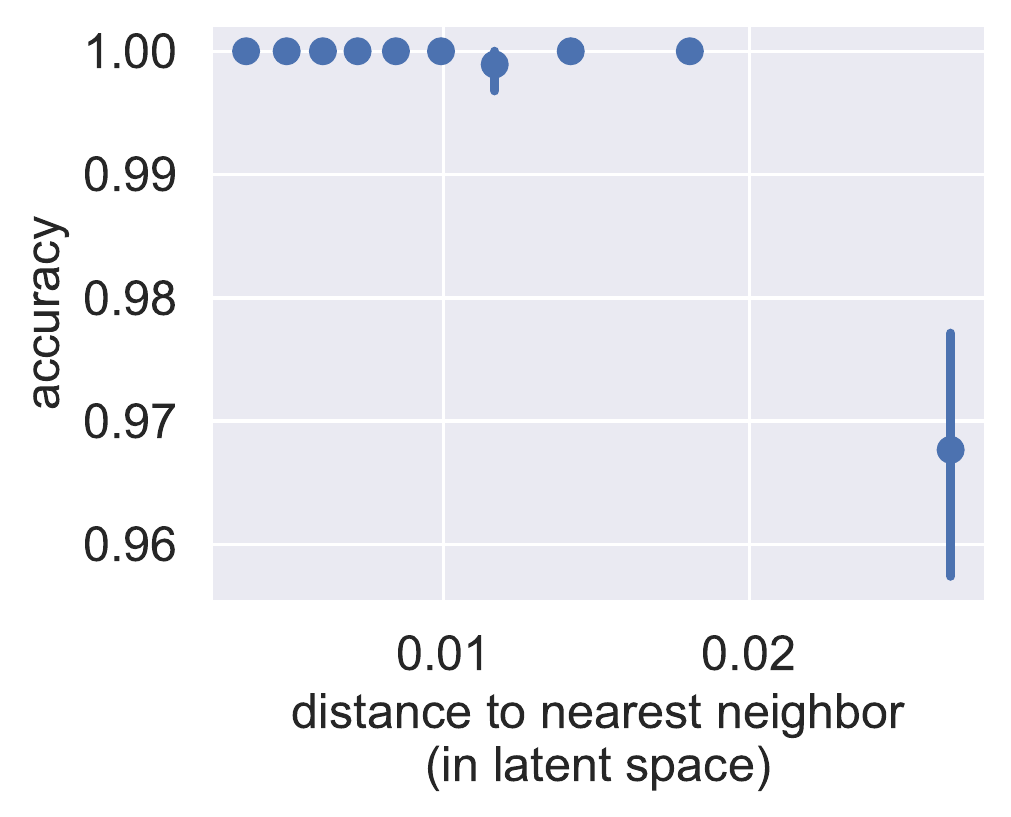} 
	\end{minipage}
	\hfill
	\begin{minipage}[t]{0.28\linewidth}	
		\textbf{C}
		\hspace{-0.5cm}\\	
		\includegraphics[width=0.9\linewidth,valign=T]{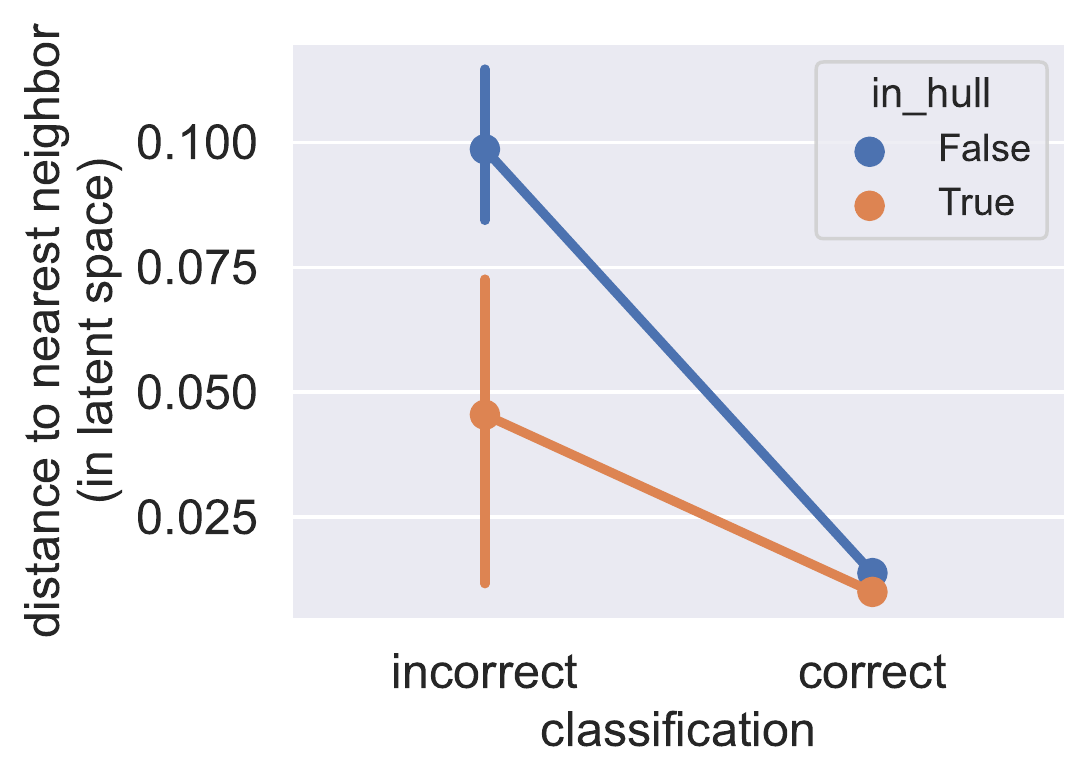}\\
		\textbf{D} 
		\hspace{-0.5cm}
		\includegraphics[width=0.9\linewidth,valign=T]{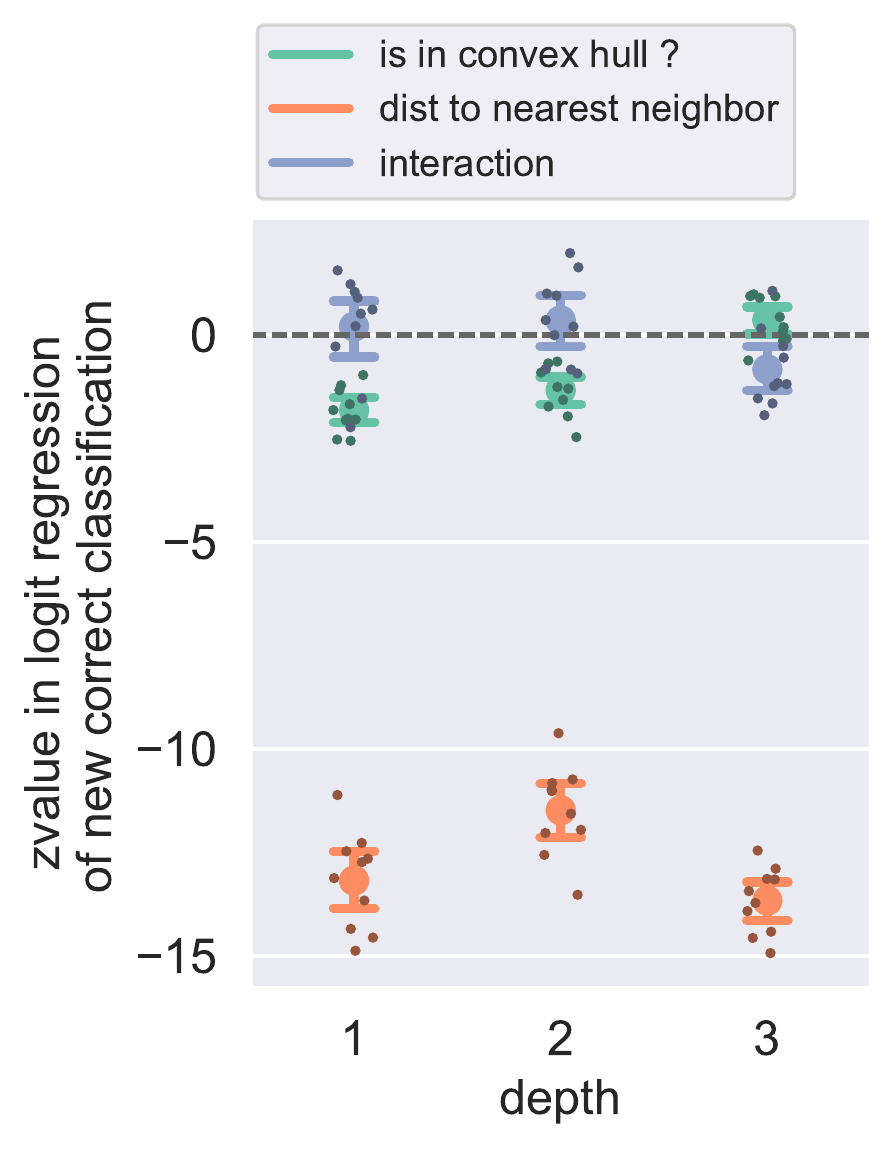} 
	\end{minipage}
	\hfill	
	\textbf{E}
	\hspace{-0.3cm}
	\includegraphics[width=0.32\linewidth,valign=T]{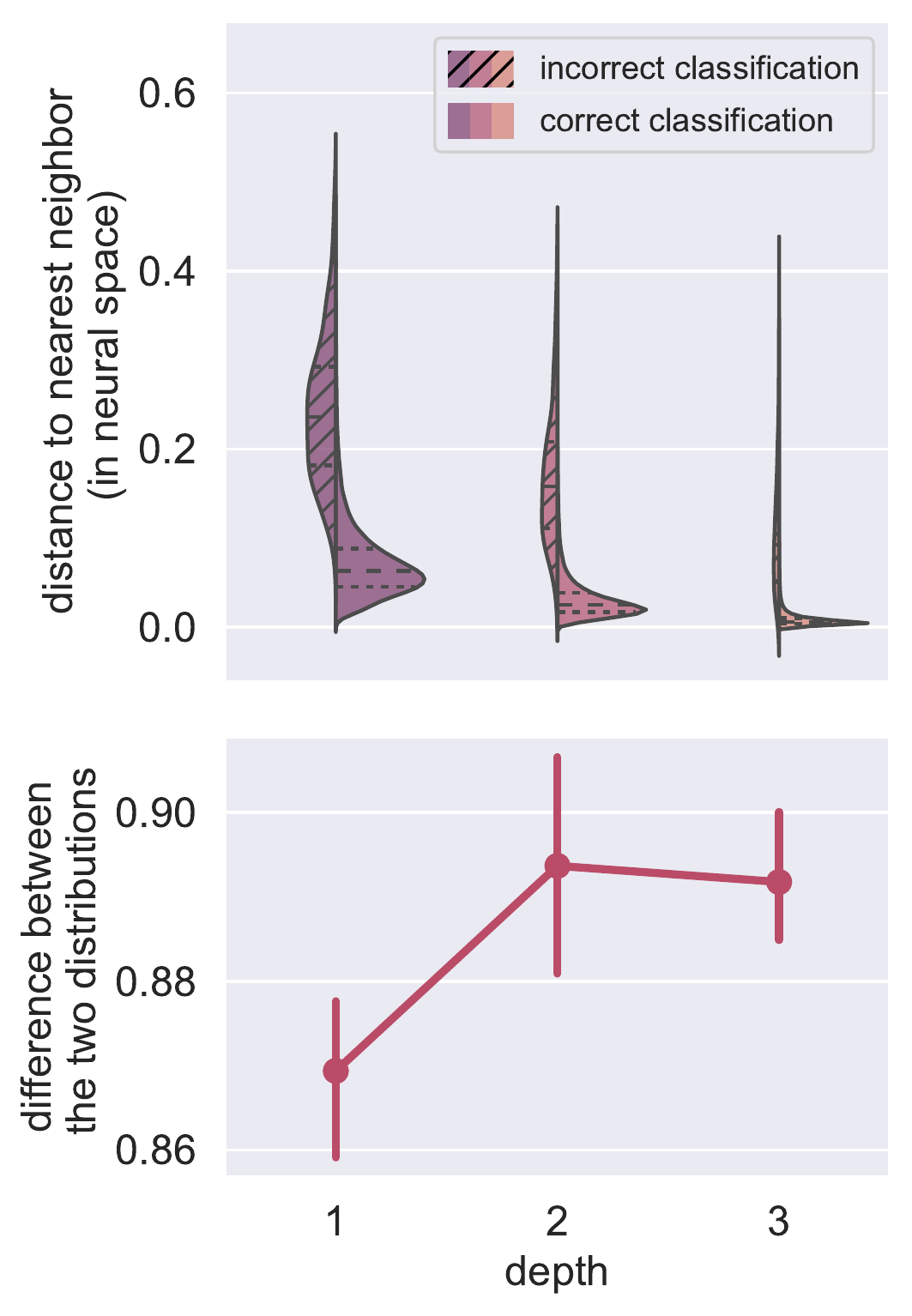} 
	
	\caption{\textbf{Distance to training set is indicative of generalization performance.} Results for the MNIST dataset and the three convolutional neural networks considered (see Fig.~3B and Fig.~4B). The distance used here corresponds to the cosine distance of a new test sample to the nearest neighbor in the training set. Otherwise, same legend as in Figure~5.}
\end{figure}

\begin{figure}
	\centering
	\begin{minipage}[t]{0.28\linewidth}	
		\textbf{A}
		\hspace{-0.3cm}
		\includegraphics[width=.95\linewidth,valign=T]{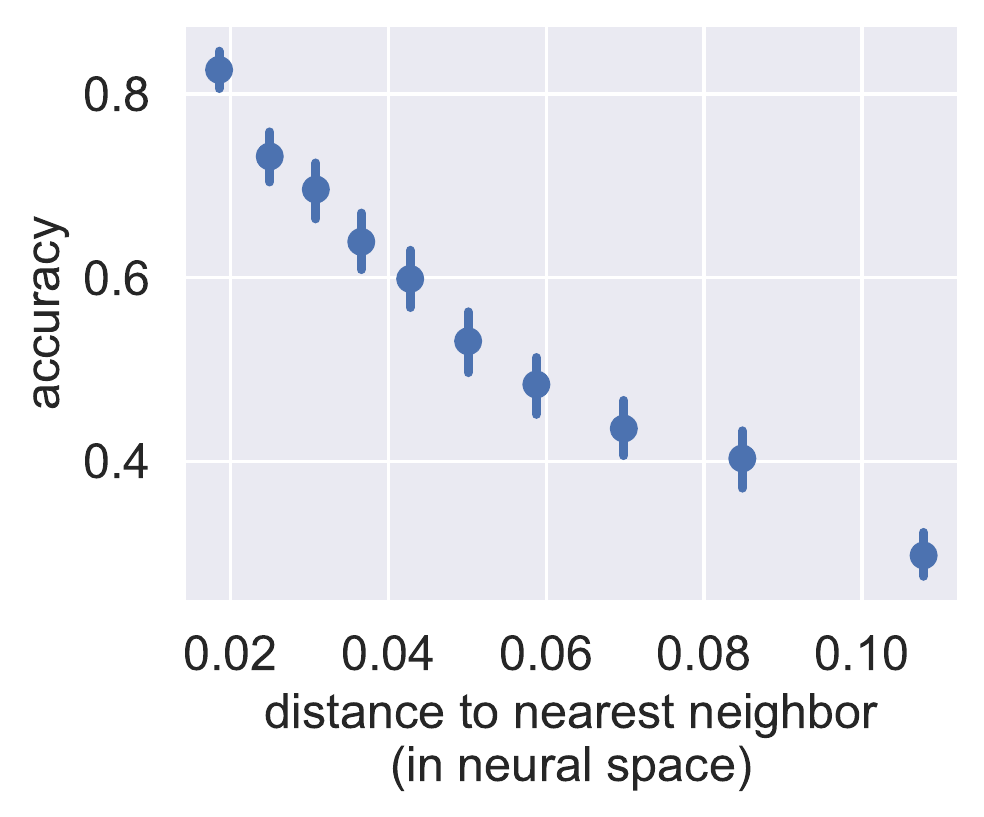}\\[20pt]
		\textbf{B}
		\hspace{-0.3cm}
		\includegraphics[width=.95\linewidth,valign=T]{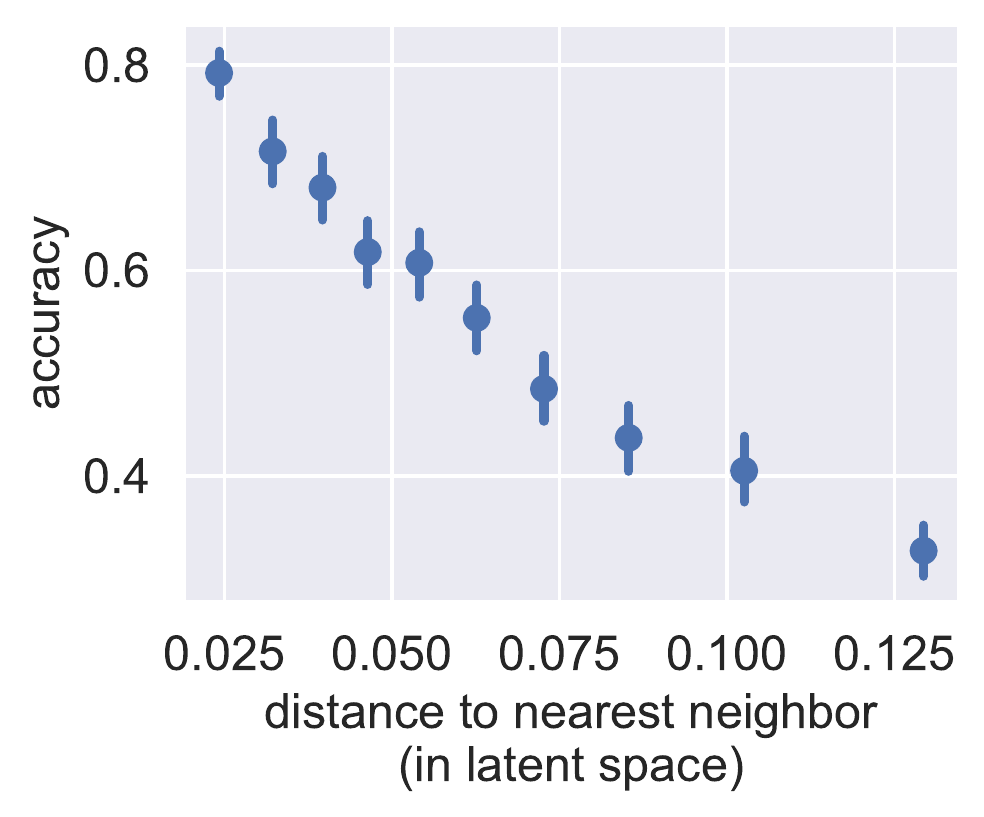} 
	\end{minipage}
	\hfill
	\begin{minipage}[t]{0.28\linewidth}	
		\textbf{C}
		\hspace{-0.5cm}\\	
		\includegraphics[width=0.9\linewidth,valign=T]{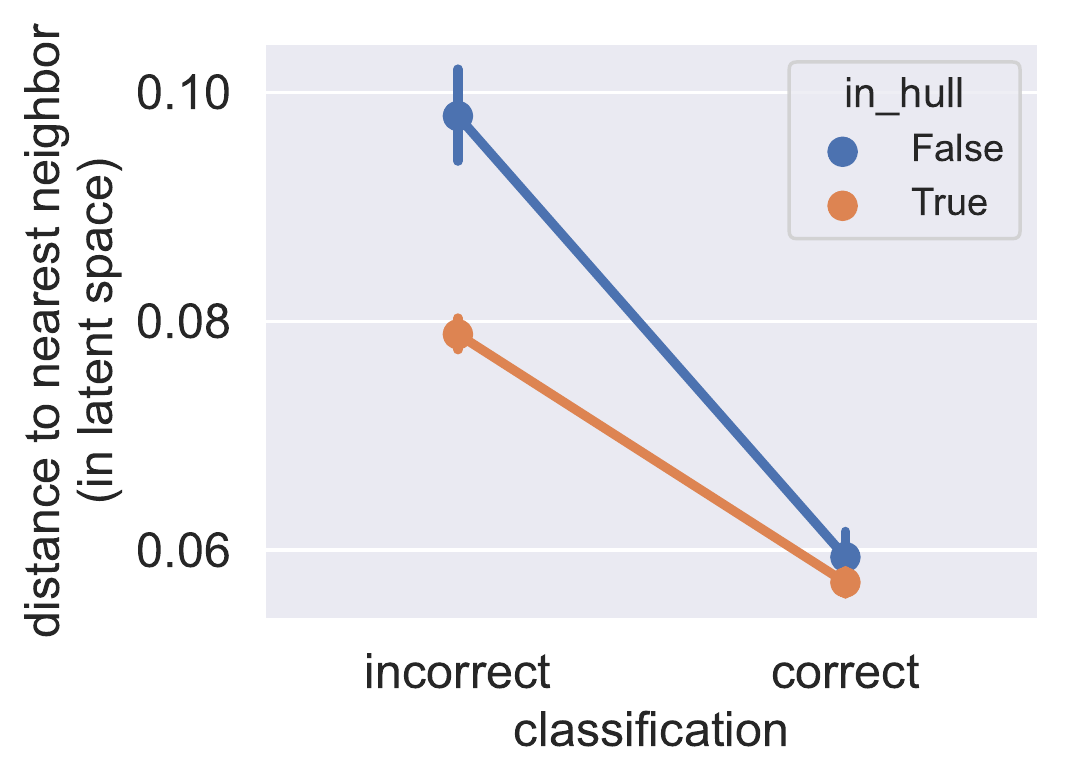}\\
		\textbf{D} 
		\hspace{-0.5cm}
		\includegraphics[width=0.9\linewidth,valign=T]{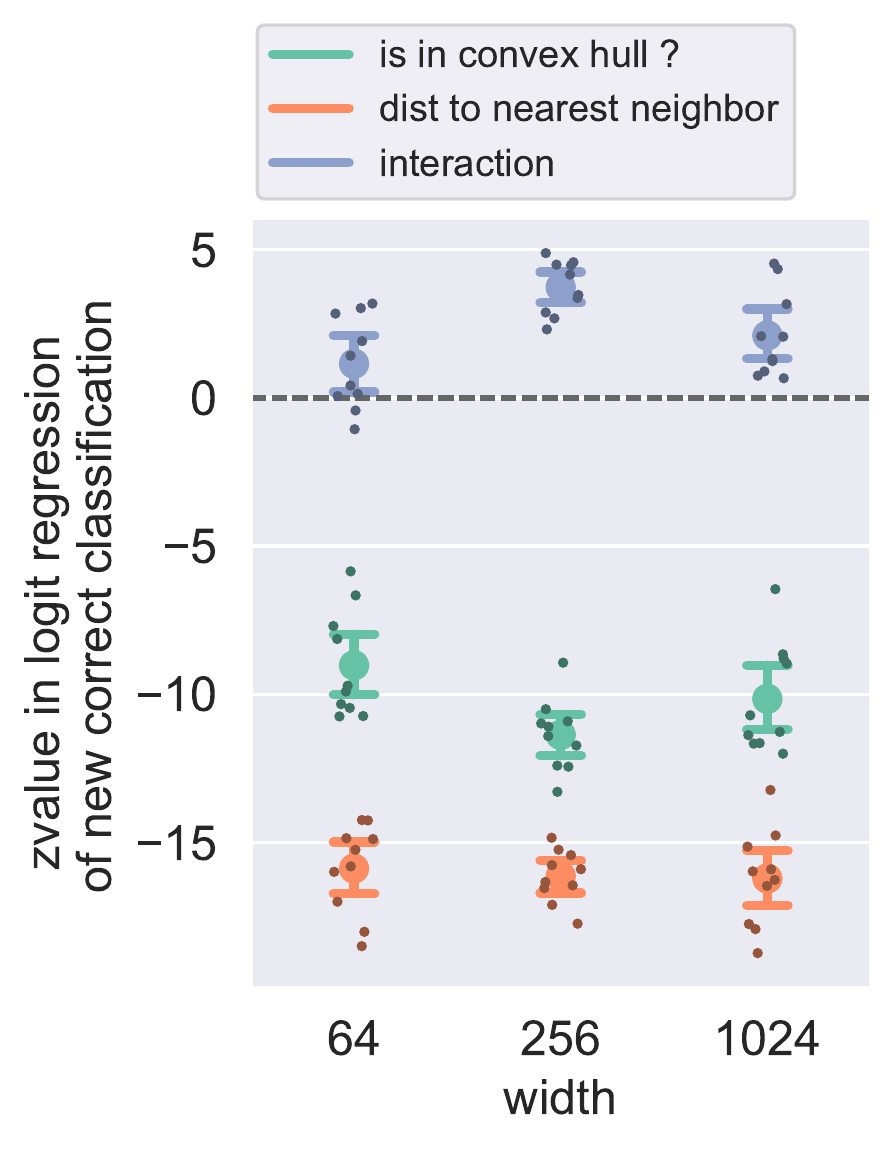} 
	\end{minipage}
	\hfill	
	\textbf{E}
	\hspace{-0.3cm}
	\includegraphics[width=0.32\linewidth,valign=T]{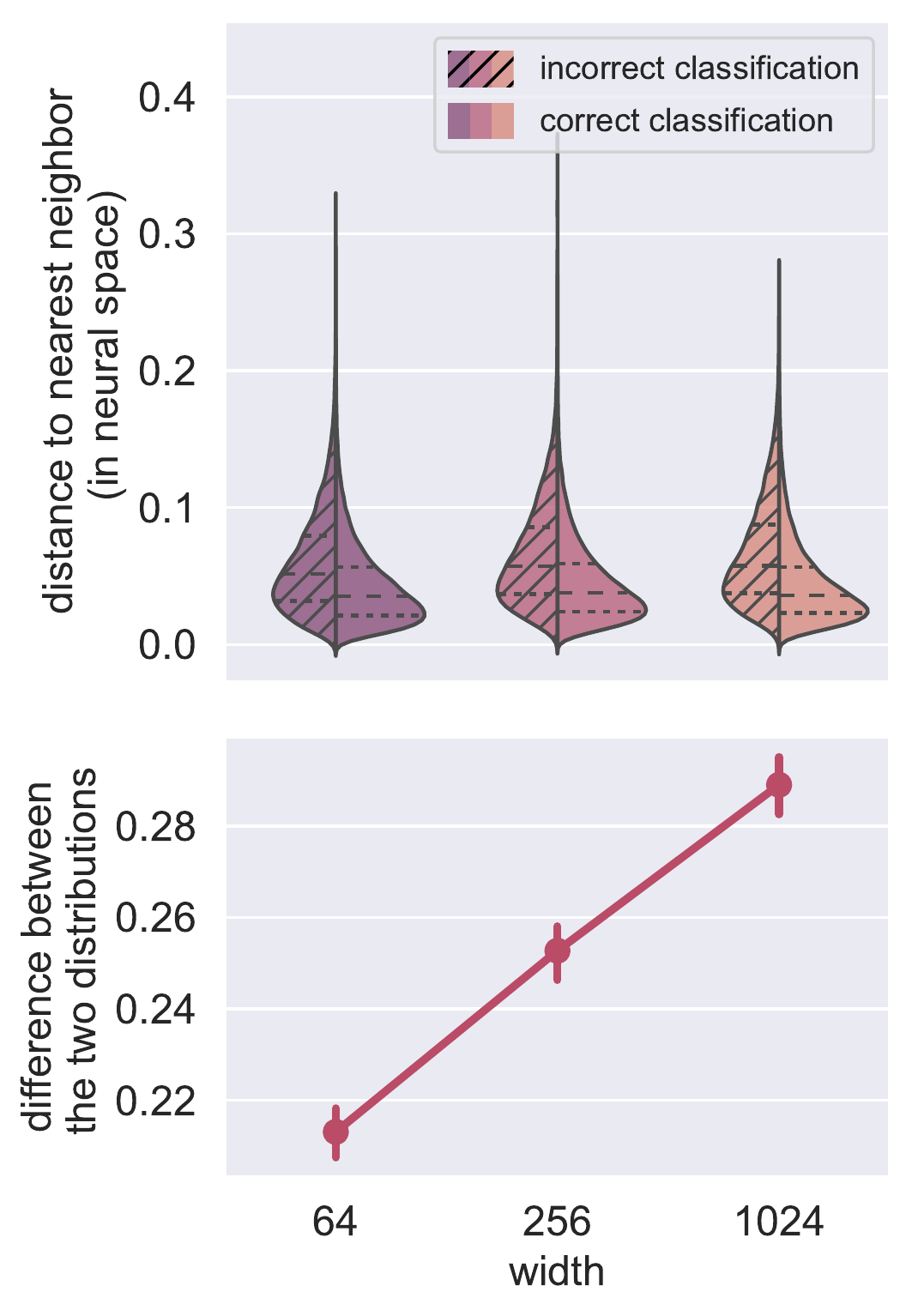} 
	
	\caption{\textbf{Distance to training set is indicative of generalization performance.} Results for the CIFAR-10 dataset and the three multilayer perceptrons considered (see Fig.~3A and Fig.~4C). The distance used here corresponds to the cosine distance of a new test sample to the nearest neighbor in the training set. Otherwise, same legend as in Figure~5.}
\end{figure}

\begin{figure}
	\centering
	\begin{minipage}[t]{0.28\linewidth}	
		\textbf{A}
		\hspace{-0.3cm}
		\includegraphics[width=.95\linewidth,valign=T]{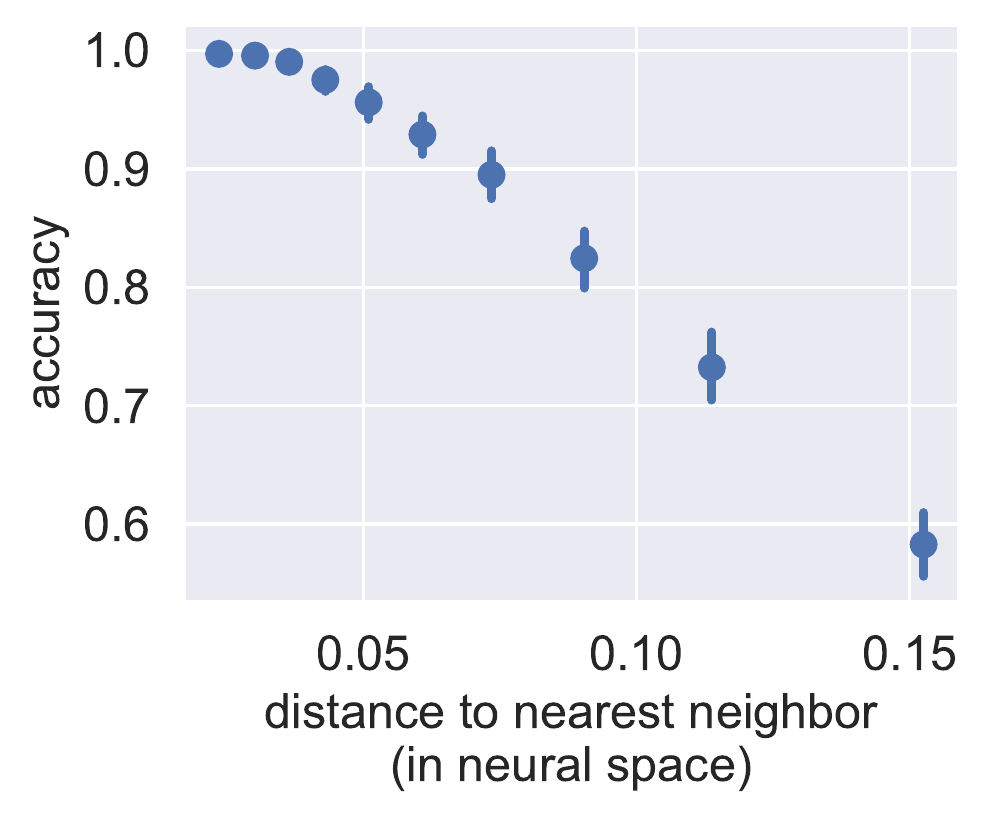}\\[20pt]
		\textbf{B}
		\hspace{-0.3cm}
		\includegraphics[width=.95\linewidth,valign=T]{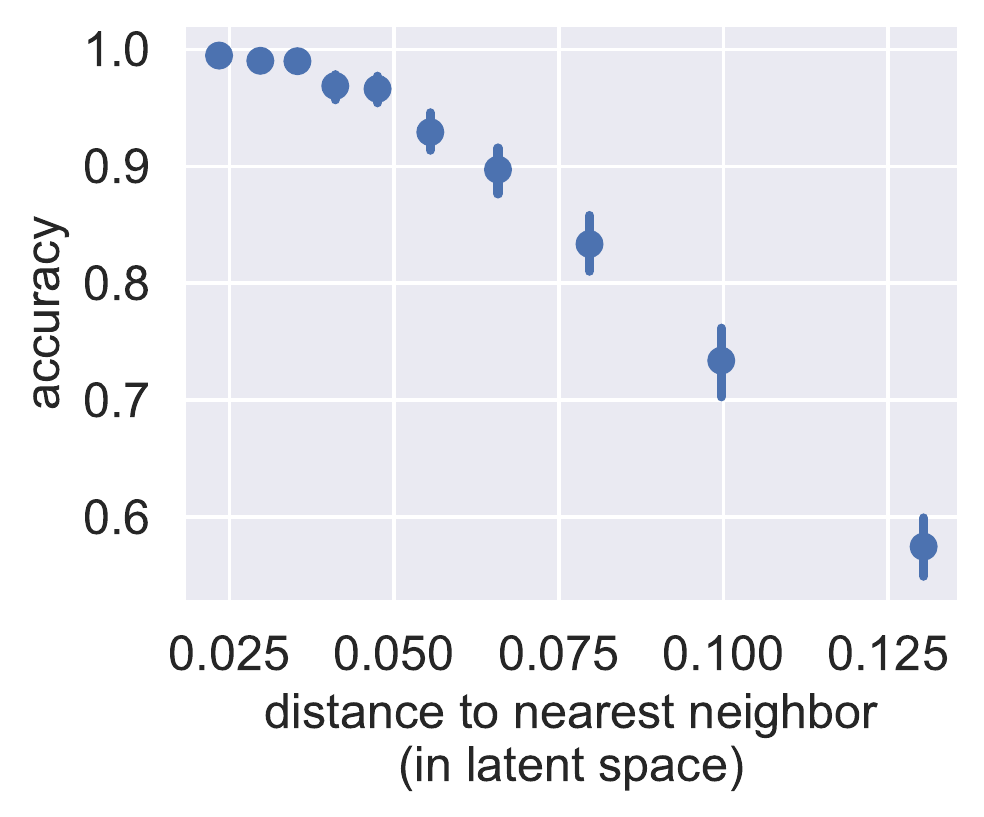} 
	\end{minipage}
	\hfill
	\begin{minipage}[t]{0.28\linewidth}	
		\textbf{C}
		\hspace{-0.5cm}\\	
		\includegraphics[width=0.9\linewidth,valign=T]{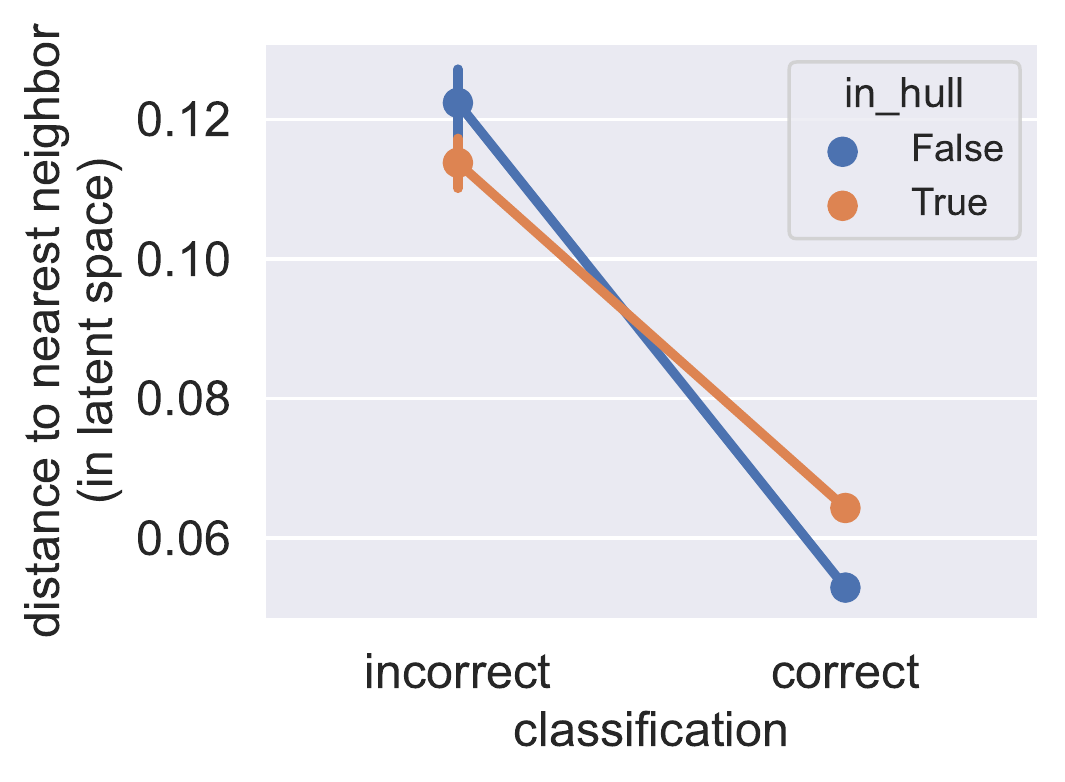}\\
		\textbf{D} 
		\hspace{-0.5cm}
		\includegraphics[width=0.9\linewidth,valign=T]{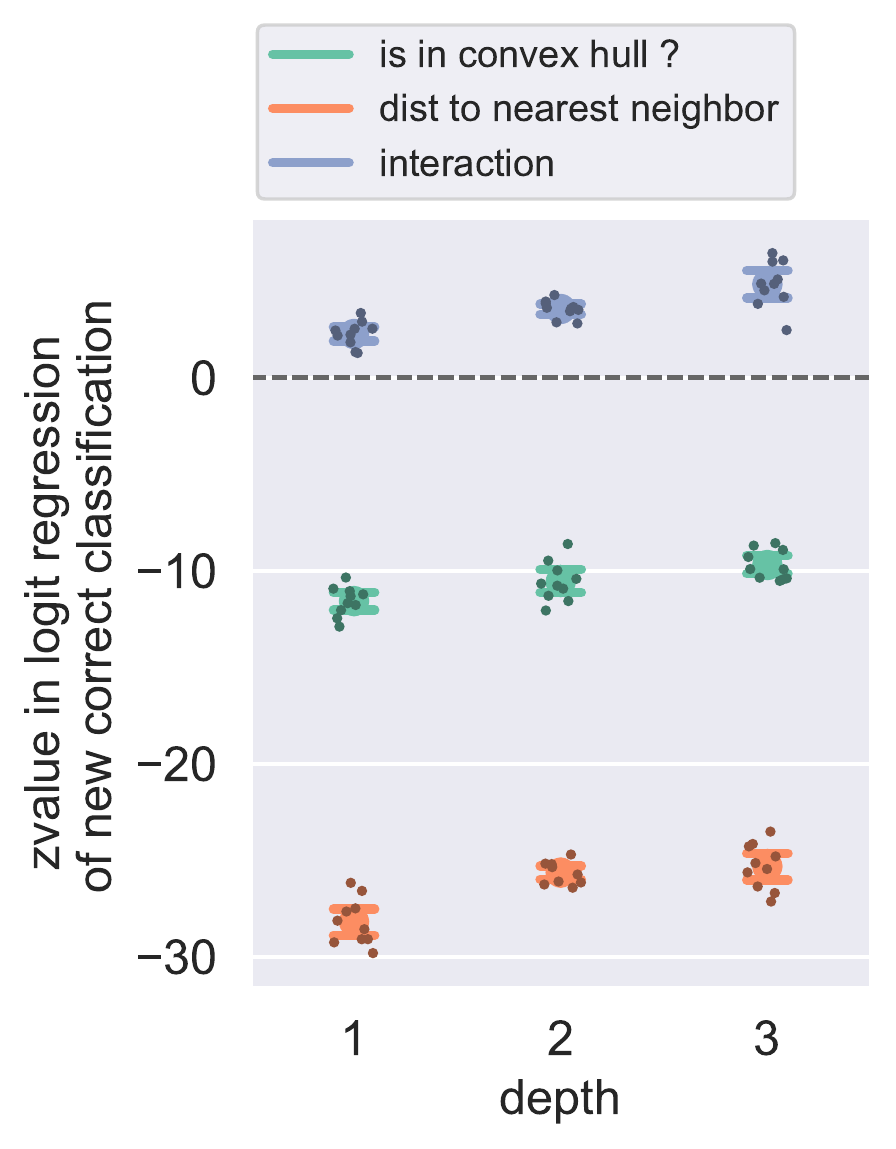} 
	\end{minipage}
	\hfill	
	\textbf{E}
	\hspace{-0.3cm}
	\includegraphics[width=0.32\linewidth,valign=T]{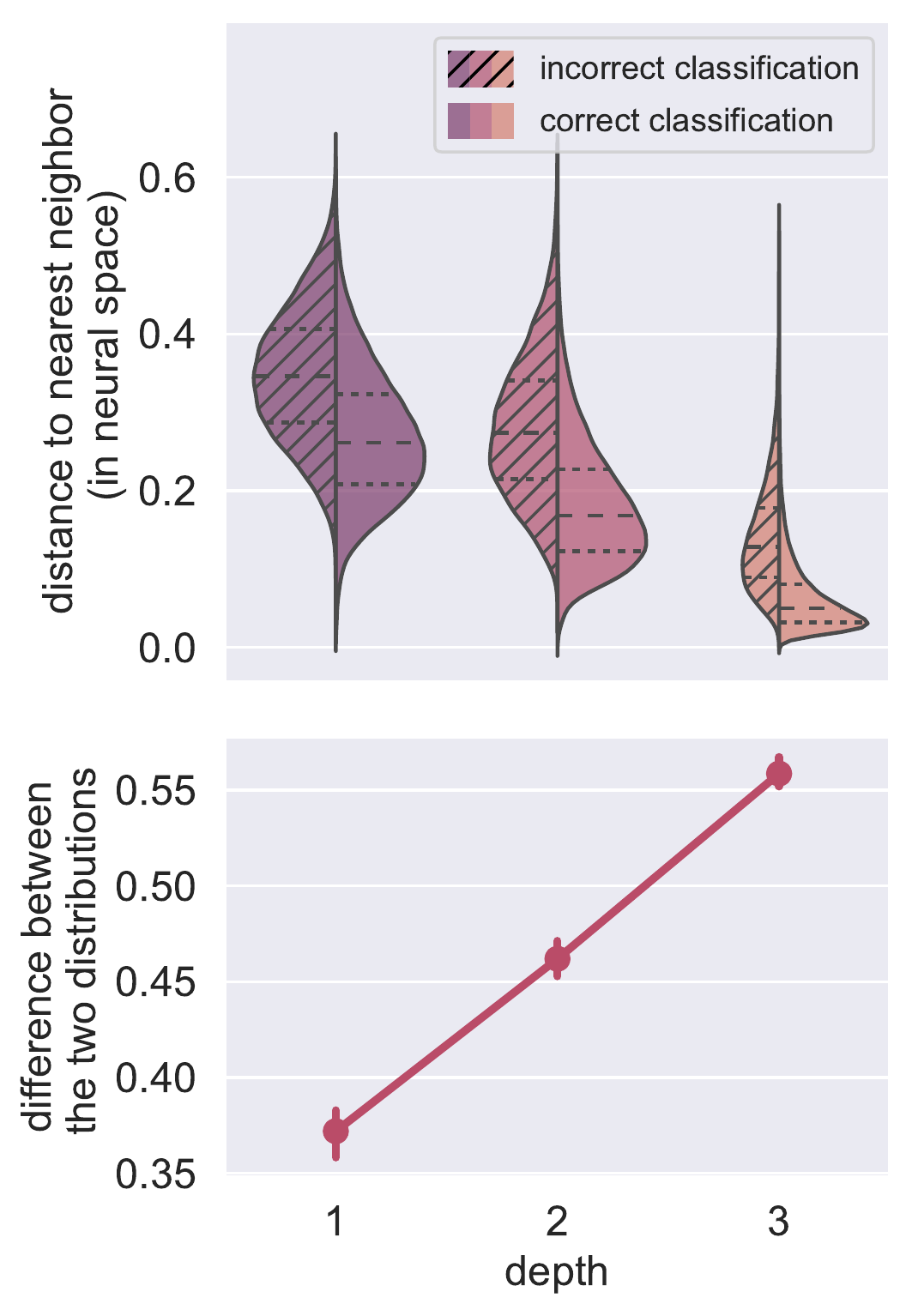} 
	
	\caption{\textbf{Distance to training set is indicative of generalization performance.} Results for the CIFAR-10 dataset and the three convolutional neural networks considered (see Fig.~3B and Fig.~4D). The distance used here corresponds to the cosine distance of a new test sample to the nearest neighbor in the training set. Otherwise, same legend as in Figure~5.}
\end{figure}